\documentclass[runningheads]{llncs}
\usepackage[T1]{fontenc}
\usepackage{lscape}
\usepackage{afterpage}
\usepackage{cite}
\usepackage{microtype}
\usepackage{amsmath}
\usepackage{amssymb}
\usepackage{subfigure}
\usepackage{booktabs} 
\usepackage{mathtools}
\usepackage{subfigure}
\usepackage{multirow}
\usepackage{xcolor}
\usepackage{enumerate}
\usepackage{float}
\usepackage{graphicx}
\usepackage{hyperref}
\usepackage[capitalize,noabbrev]{cleveref}

\usepackage{color}

\begin{document}
\title{Learning to solve Minimum Cost Multicuts efficiently using Edge-Weighted Graph Convolutional Neural Networks}
\titlerunning{Learning to solve Minimum Cost Multicuts using Edge-Weighted GNNs}
\author{
Steffen Jung\inst{1}\orcidID{0000-0001-8021-791X} \and
Margret Keuper\inst{1,2}\orcidID{0000-0002-8437-7993}
}
\authorrunning{
S. Jung, M. Keuper
}
\institute{
Max Planck Institute for Informatics, Saarland Informatics Campus \\
\email{\{steffen.jung,keuper\}@mpi-inf.mpg.de}
\and
University of Siegen
}
\maketitle              
\begin{abstract}
The minimum cost multicut problem is the NP-hard/APX-hard combinatorial optimization problem of partitioning a real-valued edge-weighted graph such as to minimize the total cost of the partition. 
While graph convolutional neural networks (GNN) have proven to be promising in the context of combinatorial optimization, most of them are only tailored to or tested on positive-valued edge weights, i.e.~they do not comply to the nature of the multicut problem.
We therefore adapt various GNN architectures including Graph Convolutional Networks, Signed Graph Convolutional Networks and Graph Isomorphic Networks to facilitate the efficient encoding of real-valued edge costs.
Moreover, we employ a reformulation of the multicut ILP constraints to a polynomial program as loss function that allows to learn feasible multicut solutions in a scalable way.
Thus, we provide the first approach towards end-to-end trainable multicuts.
Our findings support that GNN approaches can produce good solutions in practice while providing lower computation times and largely improved scalability compared to LP solvers and optimized heuristics, especially when considering large instances.
\end{abstract}
\section{Introduction}
Recent years have shown great advances of neural network-based approaches in various application domains from image classification \cite{2012AlexNet} and natural language processing \cite{2017MultiheadAttention} up to very recent advances in decision logics \cite{2021NeuralLink}.
While these successes indicate the importance and potential benefit of learning from data distributions, other domains such as symbolic reasoning or combinatorial optimization are still dominated by classical approaches. 
Recently, first attempts have been made to address specific NP-hard combinatorial problems in a learning-based setup \cite{selsam2019learning,2017CombOptGraphs, 2018CombOptGCN, 2019NPComplete}. Specifically, such papers employ (variants of) message passing neural networks ({MPNN})~\cite{2017MPNN}, defined on graphs \cite{scarselli,Micheli2009,2017GCN} in order to model, for example, the boolean satisfiability of conjunctive normal form formulas (SAT) \cite{selsam2019learning} or address the travelling salesman problem \cite{2019NPComplete} - both highly important NP-complete combinatorial problems. These first advances employ the ability of graph convolutional networks to efficiently learn representations of entities in graphs and prove the potential to solve hard combinatorial problems.
\begin{figure}[t]
    \centering
    \setlength\tabcolsep{0pt} 
    \begin{tabular}{c@{}c}
        \includegraphics[height=4.2cm]{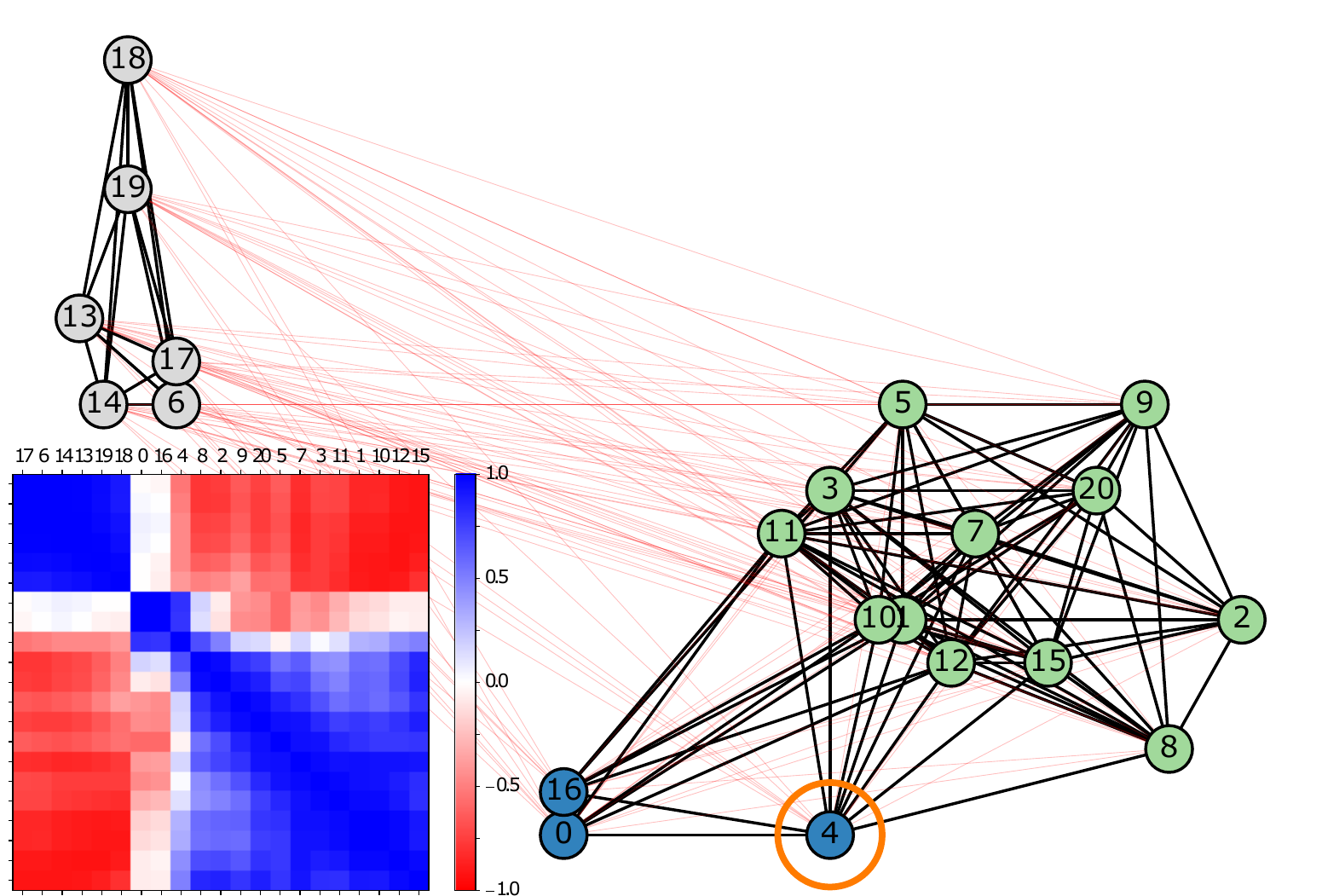} &
        \includegraphics[height=4cm]{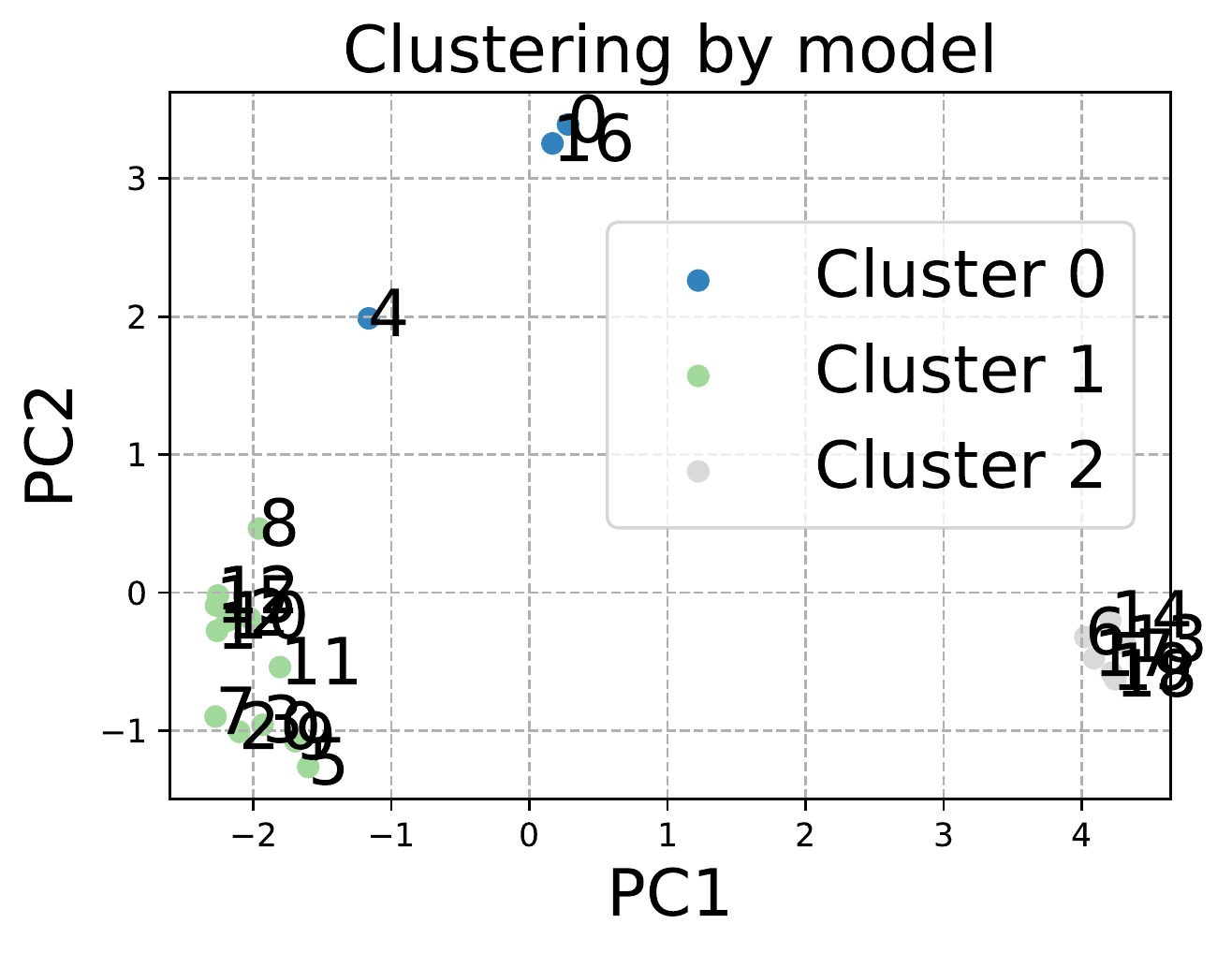} \\
        (a) Clustered nodes. &
        (b) Node emb. PCA.
    \end{tabular}
    \caption{
        (a) Node clustering of the proposed GCN\_W\_BN on a complete graph ($w=-220.6$) from IrisMP and the ordered cosine similarity between all learned node embeddings.
        (b) The first two principal components for each node embedding of (a).
        Node $4$ is part of the green cluster in the optimal solution ($w=-222.9$). 
        The closeness of both solutions is reflected in the embedding.
    }
    \label{fig:5-embeddings}
\end{figure}

In this paper, we address the \emph{minimum cost multicut} problem~\cite{demaine-2006,bansal-2004}, also known as the weighted correlation clustering problem
(see \autoref{fig:5-embeddings}).
This grouping problem is substantially different from the aforementioned examples as it aims to assign binary edge labels based on a signed edge cost function.
Such graph partitioning problems are ubiquitous in practical applications such as image segmentation \cite{2000NormalizedCut, 2011Contour, 2011Probabilistic, 2012ClosedSurface,2015Heuristics}, motion segmentation \cite{keuper17}, stereo matching \cite{2015InferenceTechniques}, inpainting \cite{2015InferenceTechniques}, object tracking \cite{2020ObjectTracking}, pose tracking \cite{insafutdinov2016deepercut}, or entity matching \cite{madoc49189}.
The minimum cost multicut problem is NP-hard, as well as APX-hard~\cite{bansal-2004}, which makes it a particularly challenging subject to explore on.
Its main difficulty lies in the exponentially growing number of constraints that define feasible solutions, especially whenever non-complete graphs are considered.
Established methods solve its binary linear program formulation or linear program relaxations \cite{2015InferenceTechniques}. However, deriving optimal solutions is oftentimes intractable for large problem instances.
In such cases, heuristic, iterative solvers are used as a remedy \cite{2015Heuristics}. A significant disadvantage of such methods is that they can not provide gradients that would allow to train downstream tasks in an end-to-end way. 

To address this issue, we propose a formulation of the minimum cost multicut problem as an {MPNN}.
 While the formulation of the multicut problem as a graph neural network seems natural, most existing GNN approaches are designed to aggregate \emph{node} features potentially under edge constraints~\cite{ecgcn}. In contrast, instances of the multicut problem are purely defined through their \emph{edge weights}. Graph Convolutional Networks (GCN)~\cite{2017GCN} rely on diverse node embeddings normalized by the graph Laplacian and an isotropic aggregation function.
 Yet, edge weights in general and signed edge weights in particular are not modeled in standard GCNs. In this paper, we propose a simple extension of GCNs and show that the \emph{signed} graph Laplacian can provide sufficiently strong initial node embeddings from signed edge information. This, in conjunction with an anisotropic update function which takes into account signed edge weights, facilitates GCNs to outperform more recent models such as
 Signed Graph Convolutional Networks (SGCN) \cite{2018Signed}, Graph Isomorphic Networks (GIN) \cite{2019GIN} as well as models that inherently handle real-valued edge weights such as Residual Gated Graph Convolutional Networks (RGGCN) \cite{2019GatedGCNTSP} and Graph Transformer Networks (GTN) \cite{2020GraphTransformer} on the multicut problem. 

To facilitate effective training, we consider a polynomial programming formulation of the minimum cost multicut problem to derive a loss function that encourages the network to issue valid solutions. Since currently available benchmarks for the minimum cost multicut problem are notoriously small, we propose two synthetic datasets with different statistics, for example w.r.t.~the graph connectivity, which we use for training and analysis. We further evaluate our models on the public benchmarks BSDS300~\cite{BSDS300}, CREMI~\cite{2017CREMI}, and Knott3D~\cite{2012ClosedSurface}.

In the following, we first briefly review the minimum cost multicut problem and commonly employed solvers.
Then, we provide an overview on GNN approaches and their application in combinatorial optimization.
In \autoref{sec:model}, we present the proposed approach for solving the minimum cost multicut problem with GNNs including model adaptations and the derivation of the proposed loss function.
\autoref{sec:results} provides an empirical evaluation of the proposed approach.


\section{The Minimum Cost Multicut Problem}
The minimum cost multicut problem \cite{1993PartProb,deza-1997} is a binary edge labeling problem defined on a graph 
$G=(V,E)$, where the connectivity is defined by edges 
$e\in E \subseteq 
{\binom{V}{2}}$, 
i.e.~$G$ is not necessarily complete. It allows for the definition of real-valued edge costs $w_e \forall e \in E$. Its solutions decompose $G$ such as to minimize the overall cost. Specifically, the MP can be defined by the following ILP \cite{1993PartProb}:
\begin{definition}
    \label{def:mp}
	For a simple, connected graph $G = (V, E)$ and an associated cost function $w\colon E \rightarrow \mathbb{R}$, written below is an instance of the multicut problem
	\begin{align}
		\min_{\mathbf{y}\in\{0,1\}^{|E|}} \;\;\;\; & c(\mathbf{y})
 	= \mathbf{y}^T \mathbf{w}
 	= \sum_{e \in E} w_e y_e
		\label{eq:problem}
	\end{align}
	with $y$ subject to the linear constraints
	\begin{align}
		\forall C \in \text{cycles}(G), \forall e \in C&\colon\,\,
		(1 - y_e) \leq \sum_{e' \in C \setminus \{e\}} (1 - y_{e'}), \label{eq:cycle_ineq}
	\end{align}
\end{definition}
where $\text{cycles}(\cdot)$ enumerates all cycles in graph $G$.
The resulting $\mathbf{y}$ is a vector of binary decision variables for each edge. Eq.~\eqref{eq:cycle_ineq} defines the cycle inequality constraints and ensures that if an edge is cut between two nodes, there can not be another path in the graph connecting them. 
Chopra and Rao \cite{1993PartProb} further showed that the facets of the MP can be sufficiently described by cycle inequalities on all chordless cycles of $G$.
The problem in Eq.~\eqref{eq:problem}-\eqref{eq:cycle_ineq} can be reformulated in a more compact way as a polynomial program (PP):
\begin{equation}
    \label{eq:qp}
    \begin{split}
    	\min_{\mathbf{y}\in\{0,1\}^{|E|}}	 	& \sum_{e \in E} w_e y_e + K \sum_{C \in \text{cc(}G\text{)}} \sum_{e \in C} y_{e}  \prod_{e' \in C \backslash \{e\}} (1-y_{e'}),
    \end{split}
\end{equation}
 for a sufficiently large
penalty $K$. 
The above problem is well behaved for complete graphs where it suffices to consider all cycles of length three and Eq.~\eqref{eq:qp} becomes a quadratic program. For sparse graphs, sufficient constraints may have arbitrary length $\leq |V|$ and their enumeration might be practically infeasible. 
Finding an optimal solution is NP-hard and APX-hard~\cite{bansal-2004}.
Therefore, exact solvers are intractable for large problem instances.
Linear program relaxations as well as primal feasible heuristics have been proposed to overcome this issue, which we will briefly review in the following.

\paragraph{Related work on Multicut Solvers.}
To solve the ILP from Definition \eqref{def:mp}, one can use general purpose {LP} solvers, like Gurobi~\cite{2020Gurobi} or CPLEX, such that optimal solutions might be in reach for small instances if the enumeration of constraints is tractable. 
However, no guarantees on the runtime can be provided.
To mitigate the exponentially growing number of constraints, various cutting-plane~\cite{kappes-2011,2011HigherOrder,kim-2014} or branch-and-bound~\cite{2012ClosedSurface,2015InferenceTechniques} algorithms exist.
For example, \cite{kappes-2011} employ a relaxed version of the {ILP} in Eq.~\eqref{eq:problem} without cycle constraints.
In each iteration, violated constraints are searched and added to the {ILP}.
This approach provides optimal solutions to formerly intractable instances - yet without any runtime guarantees.
Linear program relaxations \cite{2011HigherOrder,2015InferenceTechniques,2017MessagePassing} increase the tightness of the relaxation, for example using additional constraints,
and provide optimality bounds.
While such approaches can yield solutions within optimality bounds, their computation time can be very high and the proposed solution can be arbitrarily poor in practice.
In contrast, heuristic solvers can provide runtime guarantees and have shown good results in many practical applications.
The primal feasible heuristic KLj \cite{2015Heuristics} iterates over pairs of partitions and computes local moves which allow to escape local optima.
Competing approaches have been proposed, for example by \cite{CGC} or \cite{Beier2016}.
The highly efficient Greedy Additive Edge Contraction (GAEC) \cite{2015Heuristics} approach aggregates nodes in a greedy procedure with an $O(|E|\mathrm{log}|E|)$ worst case complexity. While such primal feasible heuristics are highly efficient in practice, they share one important draw-back with ILP solvers that becomes relevant in the learning era: they can not provide gradients that would allow for backpropagation for example to learn edge weights.

In contrast, \cite{2019EndToEnd} have proposed a third order conditional random field based on the PP in Eq.~\eqref{eq:qp}, which can be optimized in an end-to-end fashion using mean field iterations. This approach is strictly limited to the optimization on complete graphs. Our approach employs graph neural networks to overcome this limitation and provides a general purpose end-to-end trainable multicut approach.

\section{Message Passing Neural Networks for Multicuts}

\cite{2017MPNN} provide a general framework to describe convolutions for graph data spatially as message-passing scheme.
In each convolutional layer, each node is propagating its current node features via edges to all of its neighboring nodes and updates its own features based on the messages it receives.
The update is commonly described by an update function
\begin{small}
\begin{equation}
    \mathbf{h}_\mathbf{u}^{(t)} = g^{(t)} \left( \mathbf{h}_\mathbf{u}^{(t-1)}, \mathop{\hat{\sum}}_{v\in\mathcal{N}(u)} f^{(t)} \left( \mathbf{h}_\mathbf{u}^{(t-1)}, \mathbf{h}_\mathbf{v}^{(t-1)}, \mathbf{x_{v,u}} \right) \right),
    \label{eq:mpnn}
\end{equation}
\end{small}
\noindent
where $\mathbf{h}_\mathbf{u}^{(t)} \in \mathbb{R}^{F}$ is the feature representation of node $u$ in layer $t$ with dimensionality $F$, and $\mathbf{x_{v,u}}$ are edge features.
Here, $f$ and $g$ are differentiable functions, and $\hat{\Sigma}$ is a differentiable, permutation invariant aggregation function, mostly sum, max, or mean.
Commonly, the message function $f$ and the update function $g$ are parameterized, and apply the same parameters at each location in the graph.

Various formulations have been proposed to define $g$.
Graph Convolutional Network (GCN) \cite{2017GCN} normalizes messages with the graph Laplacian and linearly transforms their sum to update node representations.
Signed Graph Convolutional Network (SGCN) \cite{2018Signed} aggregates messages depending on the sign of the connectivity and keeps two representations per node, one for balanced paths and one for unbalanced paths.
Graph Isomorphic Network (GIN) \cite{2019GIN} learns an injective function by defining message aggregation as a sum and learning the update function as an MLP.
Residual Gated Graph Convolutional Network (RGGCN) \cite{2019GatedGCNTSP} computes edge gates to aggregate messages in an anisotropic manner and learns to compute the residuals to the previous representations. Edge conditioned GCNs~\cite{ecgcn} aggregate node features using dynamic weights computed from high-dimensional edge features.
Graph Transformer Network (GTN) \cite{2020GraphTransformer} also aggregate messages anisotropically by learning a self-attention model based on the transformer models in NLP \cite{2017MultiheadAttention}. While the latter three can directly handle real-valued edge weights, all are tailored towards aggregating meaningful node features. 
In the following, we review recent approaches to employ such models in the context of combinatorial optimization.

\paragraph{MPNNs and Combinatorial Optimization}
Recently, MPNNs have been applied to several hard combinatorial optimization problems, such as the minimum vertex cover~\cite{2018CombOptGCN}, maximal clique~\cite{2018CombOptGCN}, maximal independent set~\cite{2018CombOptGCN}, the satisfiability problem~\cite{2018CombOptGCN}, and the travelling salesman problem~\cite{2019GatedGCNTSP}. Their objective is either to learn heuristics such as branch-and-bound variable selection policies for exact or approximate inference  \cite{conf/nips/GasseCFCL19,Ding_Zhang_Shen_Li_Wang_Xu_Song_2020} or to use attention \cite{2015PointerNetworks}, reinforcement learning \cite{2017CombOptReinforcement, 2017CombOptGraphs}, or both \cite{2018VRP, 2019RoutingProblems, 2020CombOptGraphPointer} in an iterative, autoregressive procedure.
\cite{2019GatedGCNTSP} address the 2D Euclidean travelling salesman problem using the RGGCN model to learn edge representations. 
Other recent approaches address combinatorial problems by \emph{decoding}, using supervised training such as \cite{chen2020learning}.
The proposed approach is related to the work of \cite{2019GatedGCNTSP}, since we cast the minimum cost multicut problem as binary edge classification problem that we address using MPNN approaches, including RGGCN. We train our model in a supervised way, yet employing a dedicated loss function which encourages feasible solutions w.r.t.~Eq.~\eqref{eq:cycle_ineq}.

\begin{figure}[t]
    \centering
    \includegraphics[width=0.9\linewidth]{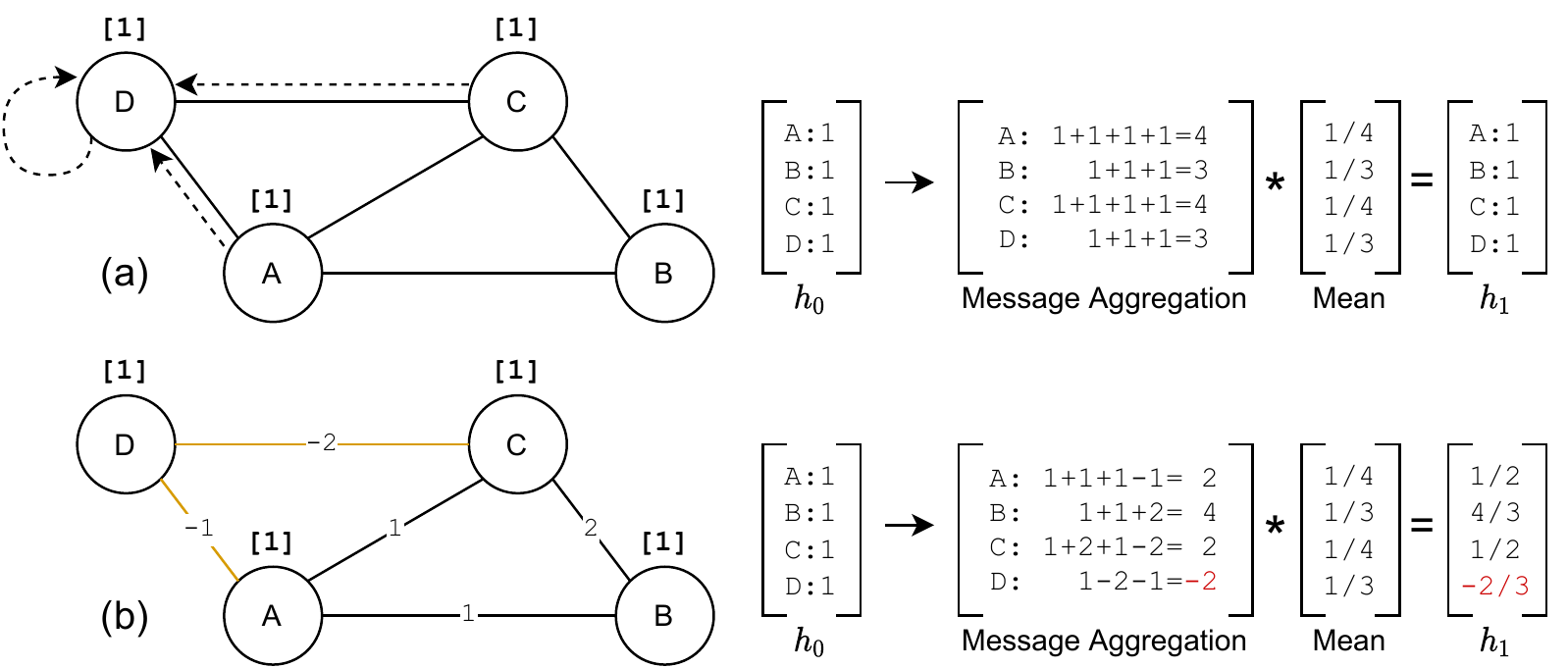} 
    \caption{
        Message aggregation in an undirected, weighted graph where node features ($h_0$) are initialized with 1.
        (a) Standard message aggregation in an isotropic fashion leads to no meaningful node embeddings ($h_1=h_0$).
        (b) Our proposed method takes edge weights into account leading to anisotropic message aggregation and meaningful node embeddings. A simple decision boundary at $h=0$ can now partition the graph.
    }
    \label{fig:4-aggregation}
\end{figure}

\subsection{Multicut Neural Network}
\label{sec:model}
We cast the multicut problem into a binary edge classification task, where label $y_{u,v}=1$ is assigned to an edge $(u,v)$ if it is cut, and $y_{u,v}=0$ otherwise.
The task of the model is to learn a probability distribution $\hat{y}_{u,v} = p(y_{u,v}=1 \;|\; G)$ over the edges of a given graph, inferring how likely it is that an edge is cut.
Based on these probabilities, we derive a configuration of edge labels, $\mathbf{y}=\{0,1\}^{|E|}$.
In contrast to existing autoregressive {MPNN}-based models in combinatorial optimization, we derive a solution after a \emph{single forward pass} over the graph to achieve an efficient bound on the runtime of the model.
In this scenario, our model can be defined by three functions, i.e., $\mathbf{y}_{u,v}=f_r(f_c(f_e(G,w)))$.
First, ${f_e}$ is the edge representation mapping, assigning meaningful embeddings to each edge in the graph given a multicut problem instance.
This function is learned by an {MPNN}.
Second, ${f_c}$ assigns to every edge its probability to be cut.
This function is learned by an {MLP}.
Last, function ${f_r}$ translates the resulting configuration of edge probabilities to a feasible configuration of edge labels, hence, computes a feasible solution.

\subsubsection{Edge Representation Mapping}
Given a multicut problem instance $(G,w)$, the edge representation mapping $f_e$ learns to assign meaningful edge embeddings via {MPNN}s.
One specific case of MPNN is GCN \cite{2017GCN}, where the node representation update function is defined as follows:
\begin{align}
    \mathbf{h}_\mathbf{u}^{(t)} &= g^{(t)}_\theta \left(
    \mathbf{h}_\mathbf{u}^{(t-1)} +
    \sum_{v\in\mathcal{N}(u)}  \mathbf{{L}}_{[v,u]} \mathbf{h}_\mathbf{v}^{(t-1)}
    \right),
    \label{eq:3-gcn}
\end{align}
where $\mathbf{h}_\mathbf{u}^{(t)} \in \mathbb{R}^{F}$ denotes the feature representation of node $u$ in layer $t$ with channel size $F$.
In each layer, node representations of all neighbors of $u$ are aggregated and normalized by $\mathbf{{L}}_{[v,u]}=1 / \sqrt{\text{deg}(u)\text{deg}(v)}$, where $\mathbf{{L}} = \mathbf{\tilde{D}}^{1/2} \mathbf{\tilde{A}} \mathbf{\tilde{D}}^{1/2}$ is the normalized graph Laplacian with additional self-loops in the adjacency matrix $\mathbf{\tilde{A}} = \mathbf{{A}} + \mathbf{I}$ and degree matrix $\mathbf{\tilde{D}}$.
Conventionally, $\mathbf{h}_\mathbf{u}^{(0)}$ is initialized with node features $\mathbf{x_u}$.
Intuitively, we expect normalization with the graph Laplacian to be beneficial in the MP setting, since i) its eigenvectors encode similarities of nodes within a graph \cite{2000NormalizedCut} and ii) even sparsely connected nodes can be assigned meaningful representations \cite{2011Contour}.
However, MP instances consist of real-valued edge-weighted graphs and the normalized graph Laplacian is not defined for negative node degrees.
To the best of our knowledge, there is no work enabling GCN to incorporate \emph{real-valued} edge weights so far.

\paragraph{Real-valued Edge Weights}
Hence, our first task is to enable negative-valued edge weights in GCN.
We can achieve this via the \emph{signed} normalized graph Laplacian \cite{2005SignedLaplacian, 2010SignedLaplacian}:
\begin{equation}\begin{split}\label{eq:signed-laplacian}
    \overline{\mathbf{L}}_{[v,u]} &= \left( \overline{\mathbf{D}}^{1/2}\tilde{\mathbf{W}}\overline{\mathbf{D}}^{1/2} \right)_{[v,u]} \\
    &= w_{v,u} / \sqrt{\overline{\text{deg}}(u)\overline{\text{deg}}(v)},
\end{split}\end{equation}
where $\tilde{\mathbf{W}}$ is the weighted adjacency matrix and $\overline{\mathbf{D}}$ is the signed node degree matrix with $\overline{\text{deg}}(u) = \sum_{v\in \mathcal{N}(u)} |w_{u,v}|$.
\cite{2016SignedLaplacian} shows that this formulation preserves the desired properties from the graph Laplacian w.r.t. encoding pairwise similarities as well as  representation learning on sparsely connected nodes (see i) and ii) above).

Incorporting Eq. \eqref{eq:signed-laplacian} into Eq. \eqref{eq:3-gcn}, we get
\begin{equation}\begin{split}\label{eq:3-gcn-signed}
    \mathbf{h}_\mathbf{u}^{(t)} &= g^{(t)}_\theta \Big(
    \mathbf{h}_\mathbf{u}^{(t-1)} + 
    \sum_{v\in\mathcal{N}(u)}
        w_{v,u} \cdot \left({\overline{\text{deg}}(u)\overline{\text{deg}}(v)}\right)^{-1/2}
    \mathbf{h}_\mathbf{v}^{(t-1)}
    \Big).
\end{split}\end{equation}
Here, we can observe two new terms.
First, each message is weighted by the edge weight $w_{v,u}$ between two nodes enabling an anisotropic message-passing scheme.
\autoref{fig:4-aggregation} motivates why this is necessary.
While \cite{2019GIN} show that GNNs with mean aggregation have theoretical limitations, they also note that these limitations vanish in scenarios where node features are diverse.
Additionally, \cite{2019GIN} only consider the case where neighboring nodes are aggregated in an isotropic fashion.
As we show here, diverse node features are not necessary when messages are aggregated in the anisotropic fashion we propose.
The resulting node representations enable to distinguish nodes in the graph despite the lack of meaningful node features. 
This is important in our case, since the multicut problem does not provide node features.
Second, we are now able to normalize messages via the Laplacian in real-valued graphs.
The normalization acts stronger on messages that are sent to or from nodes whose adjacent edges have weights with large magnitudes.
Large magnitudes on the edges usually indicate a confident decision towards joining (for positive weights) or cutting (negative weights).
Thus, the normalization will allow nodes with less confident edge cues to converge to a meaningful embedding while, without such normalization, the network would notoriously focus on embedding nodes with strong edge cues, i.e.~on easy decisions.

\paragraph{Node Features}
\label{sec:A-node-features}
Conventionally, node representations at timestep $0$, $\mathbf{h}_\mathbf{u}^{(0)}$, are initialized with node features $\mathbf{x_u}$.
However, multicut instances describe the magnitude of similarity or dissimilarity between two items via edge weights and provide no node features.
Therefore, we initialize node representations with a two-dimensional vector of node degrees as:
\begin{equation}
    \mathbf{x_u} = {\left(
        \sum_{v \in \mathcal{N}^+(u)} w_{u,v},
        \sum_{v \in \mathcal{N}^-(u)} w_{u,v}
    \right)},
\end{equation}
where $\mathcal{N}^+(u)$ is the set of neighboring nodes of $u$ connected via positive edges, and $\mathcal{N}^-(u)$ is the set of neighboring nodes of $u$ connected via negative edges.

\paragraph{Node-to-Edge Representation Mapping}
\label{sec:A-node-concat}
To map two node representations to an edge representation, we use concatenation
$
    \mathbf{h_{u,v}} = 
    {f_e}({\mathbf{h_u}},{\mathbf{h_v}}) = 
    \binom{\mathbf{h_u}}{\mathbf{h_v}}  \in \mathbb{R}^{2\cdot F}
$, where $\mathbf{h_{u,v}}$ is the representation of edge $(u,v)$ and $F$ the dimension of node embedding $\mathbf{h_u}$.
Since we consider undirected graphs, the order of the concatenation is ambiguous.
Therefore, we generate two representations for each edge, one for each direction.
This doubles the number of edges to be classified in the next step. The final classification result is the average computed from both representations.

\paragraph{Edge Classification}
We learn edge classification function ${f_c}$ via an {MLP} that computes likelihoods $\mathbf{\hat{y}} \in [0,1]^{|E|}$ for each edge in graph $G$, expressing the confidence whether an edge should be cut.
A binary solution $\mathbf{{y}} \in \{0,1\}^{|E|}$ is retrieved by thresholding the likelihoods at $0.5$. %
Since there is no strict guarantee that the edge label configuration $\mathbf{{y}}$ is feasible w.r.t.~Eq.~\eqref{eq:cycle_ineq}, we postprocess $\mathbf{{y}}$ to \emph{round} it to feasible solutions. Therefore, we compute a connected component labeling on $G$ after removing cut edges from $E$ and reinstate removed edges for which both corresponding nodes remain within the same component.
For efficiency, we implement the connected component labeling as a message-passing layer and can therefore assign cluster identifications to each node efficiently on the GPU.
\subsubsection{Training}

Since we cast the multicut problem to a binary edge labelling problem, we can formulate a supervised training process that minimizes the Binary Cross Entropy (BCE) loss w.r.t.~the optimal solution $\mathbf{\tilde{y}}$, which we denote $ \mathcal{L}_{BCE}$. 
\paragraph{Cycle Consistency Loss}

The {BCE} loss encodes feasibility only implicitly by comparison to the optimal solution. To explicitly learn feasible solutions,
we take recourse to the PP formulation of the multicut problem in Eq.~\eqref{eq:qp} and formulate a \emph{feasibility loss}, that we denote \emph{Cycle Consistency Loss} (CCL):
\begin{equation}
    \label{eq:ccloss}
    \mathcal{L}_{CCL} = 
	\alpha \cdot \sum_{C \in cc(G,l)} \sum_{e \in C} \hat{y}_e \prod_{e' \in C \backslash \{e\}} (1-\hat{y}_{e'}),
\end{equation}
where $\alpha$ is a hyperparameter, balancing {BCE} and {CCL}, and $cc(G,l)$ is a function that returns all chordless cycles in $G$ of length at most $l$.
Term effectively penalizes infeasible edge label configurations during training; it adds a penalty of at most $\alpha$ for each chordless cycle that is only cut once.
In practice, we only consider chordless cycles of maximum length $l$, and we only consider a cycle if $e$ is cut, hence $\hat{y}_e \geq 0.5$.
This is necessary to ensure practicable training runtimes.
The total training loss is given by $ \mathcal{L} = \mathcal{L}_{BCE} + \mathcal{L}_{CCL}$. 
For best results, we train all models using batch normalization.
\begin{figure}[t]
    \centering
   
        \setlength\tabcolsep{0pt}
        \begin{tabular}{@{}c@{}c@{}c@{}}
            \includegraphics[width=0.25\linewidth]{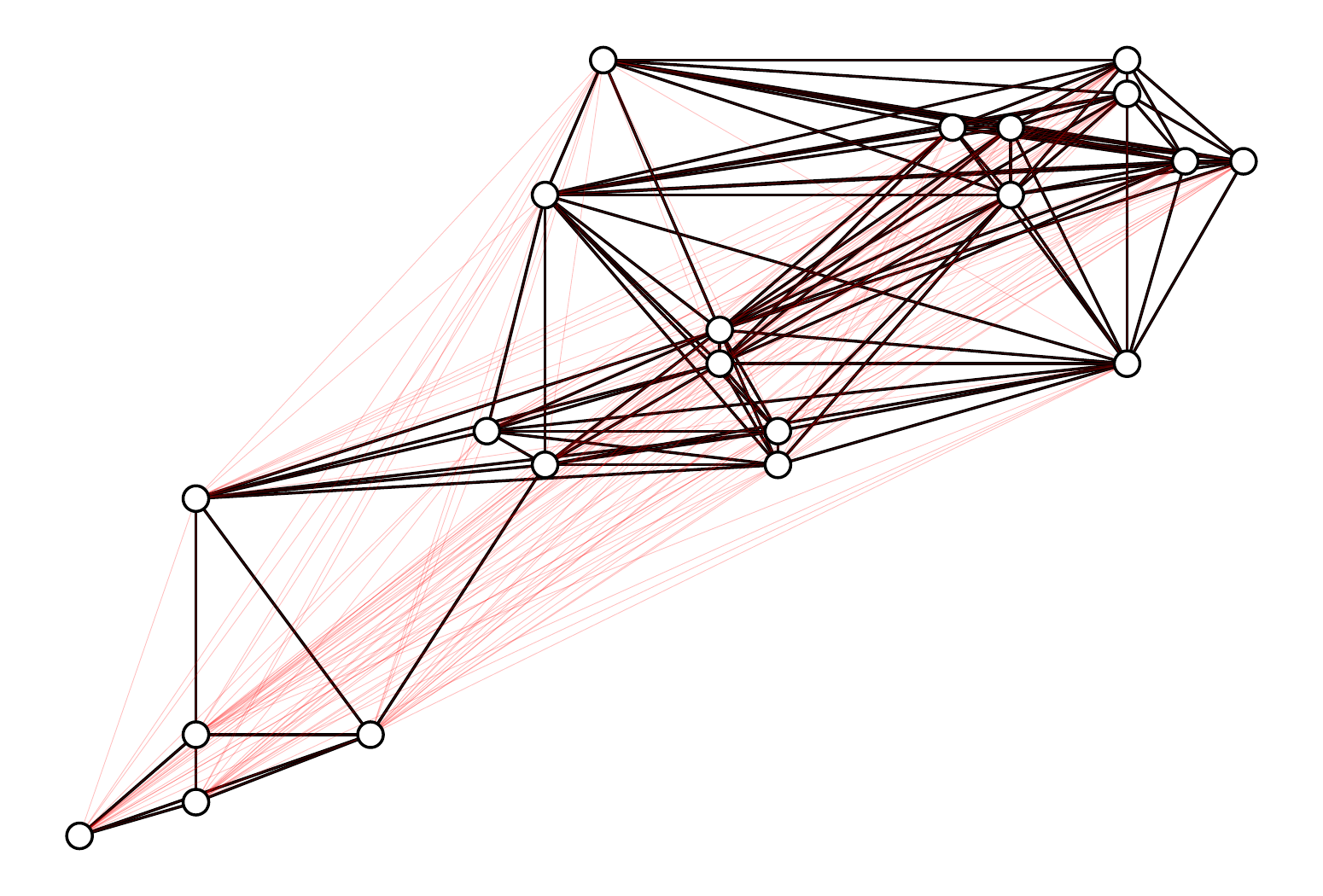} &
            \includegraphics[width=0.25\linewidth]{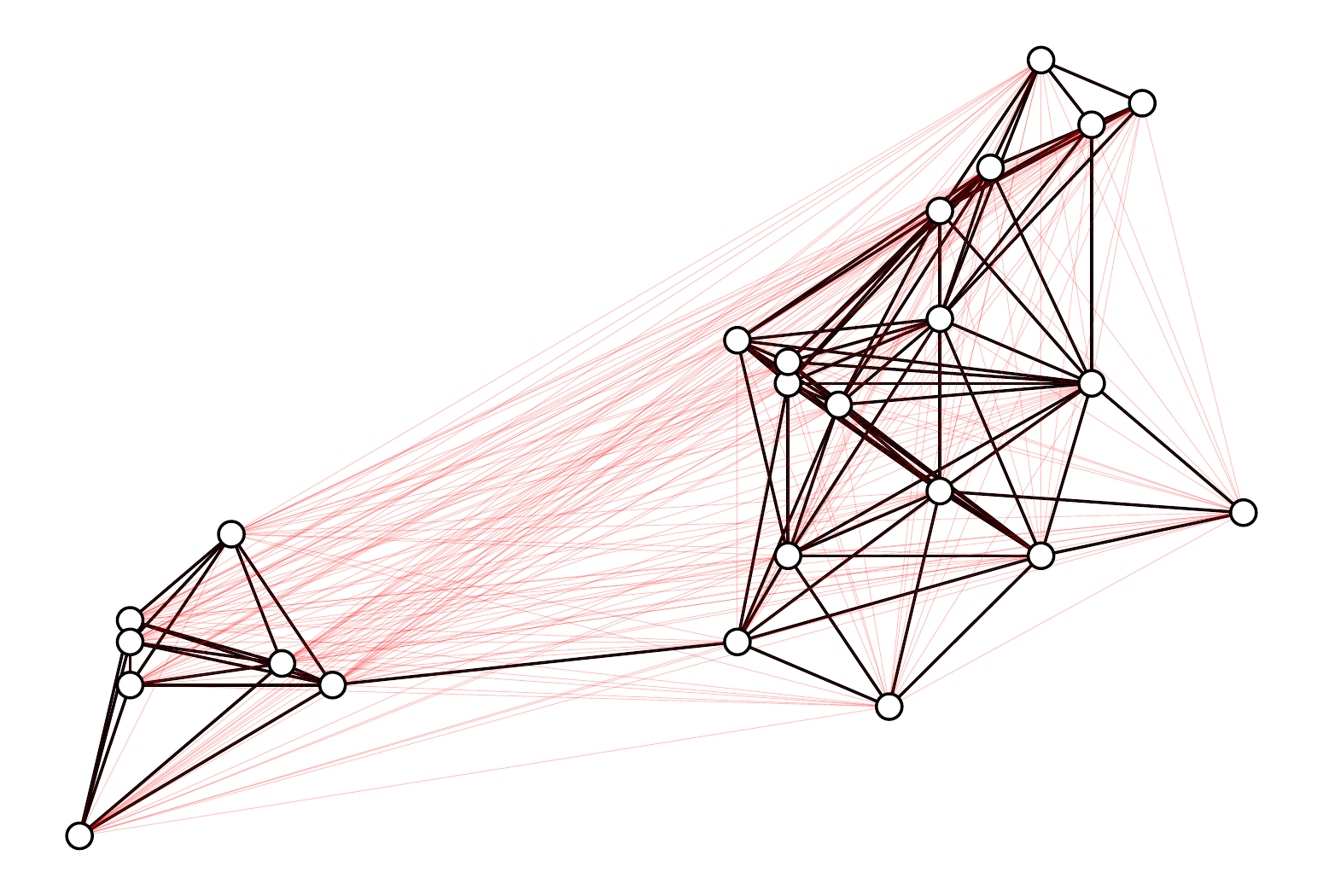} & 
            \includegraphics[width=0.25\linewidth]{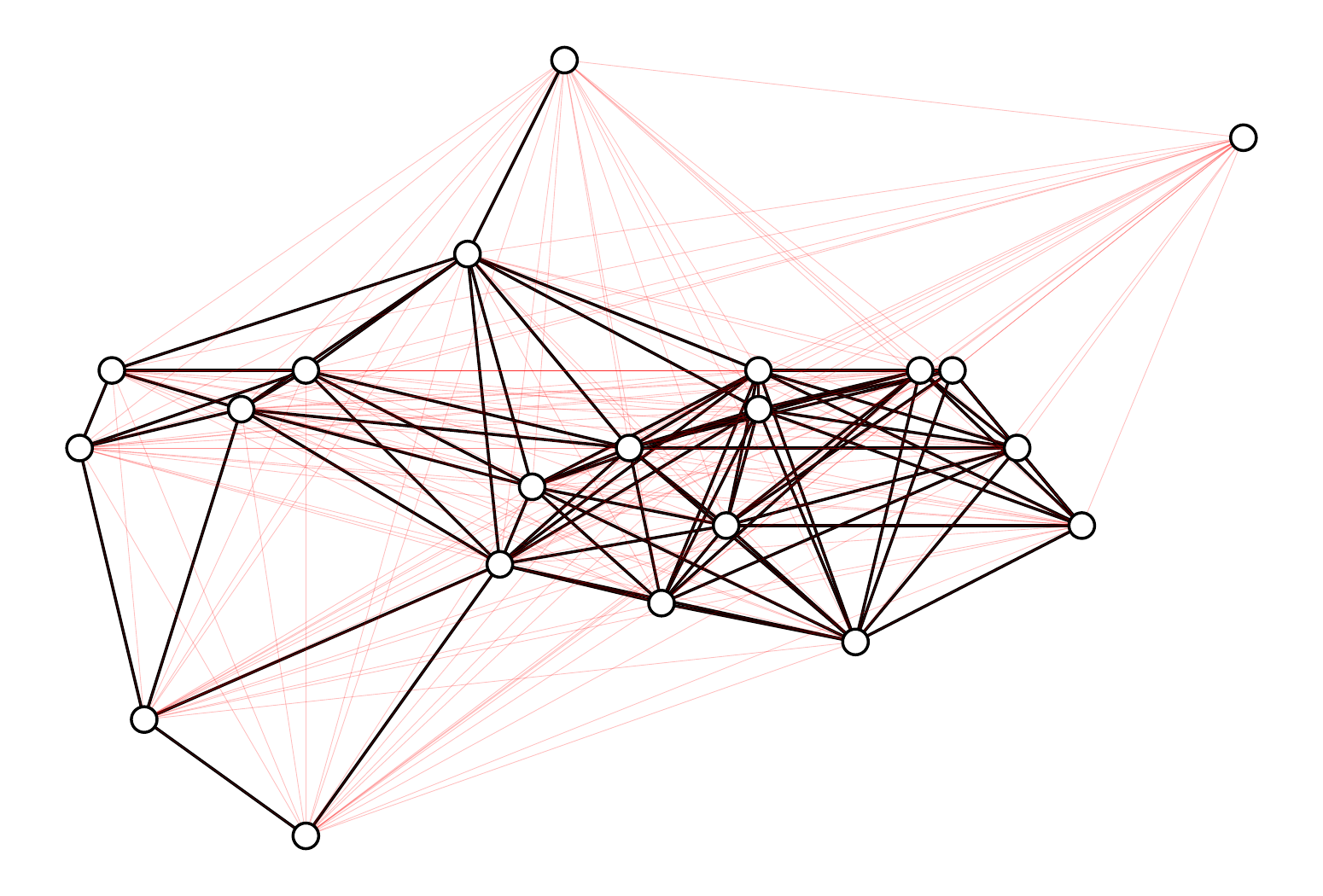} \\
            (a) Graph 1. &
            (b) Graph 2. & 
            (c) Graph 3. \\
            
            \multicolumn{3}{l}{} \\
        
            \includegraphics[width=0.25\linewidth]{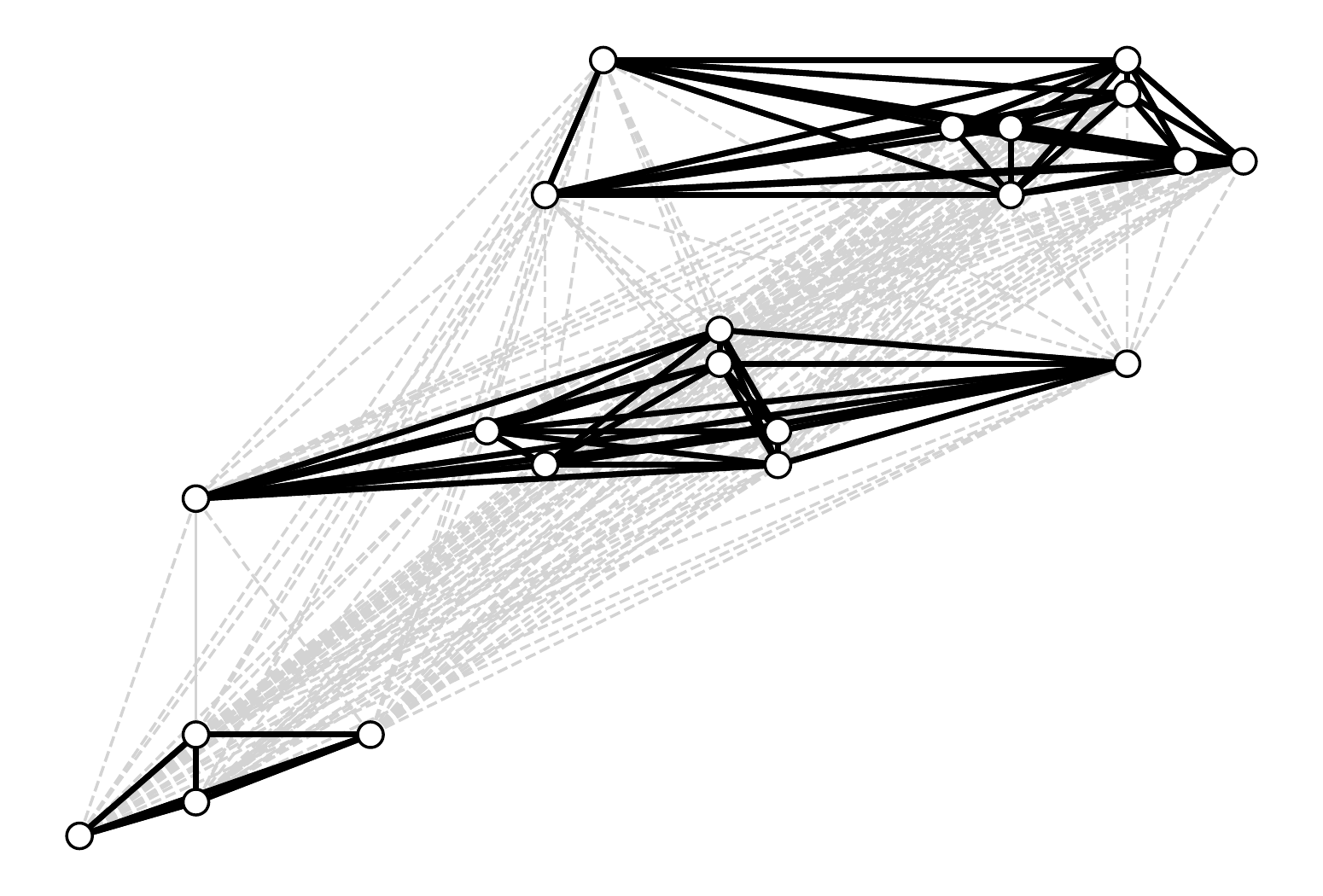} &
            \includegraphics[width=0.25\linewidth]{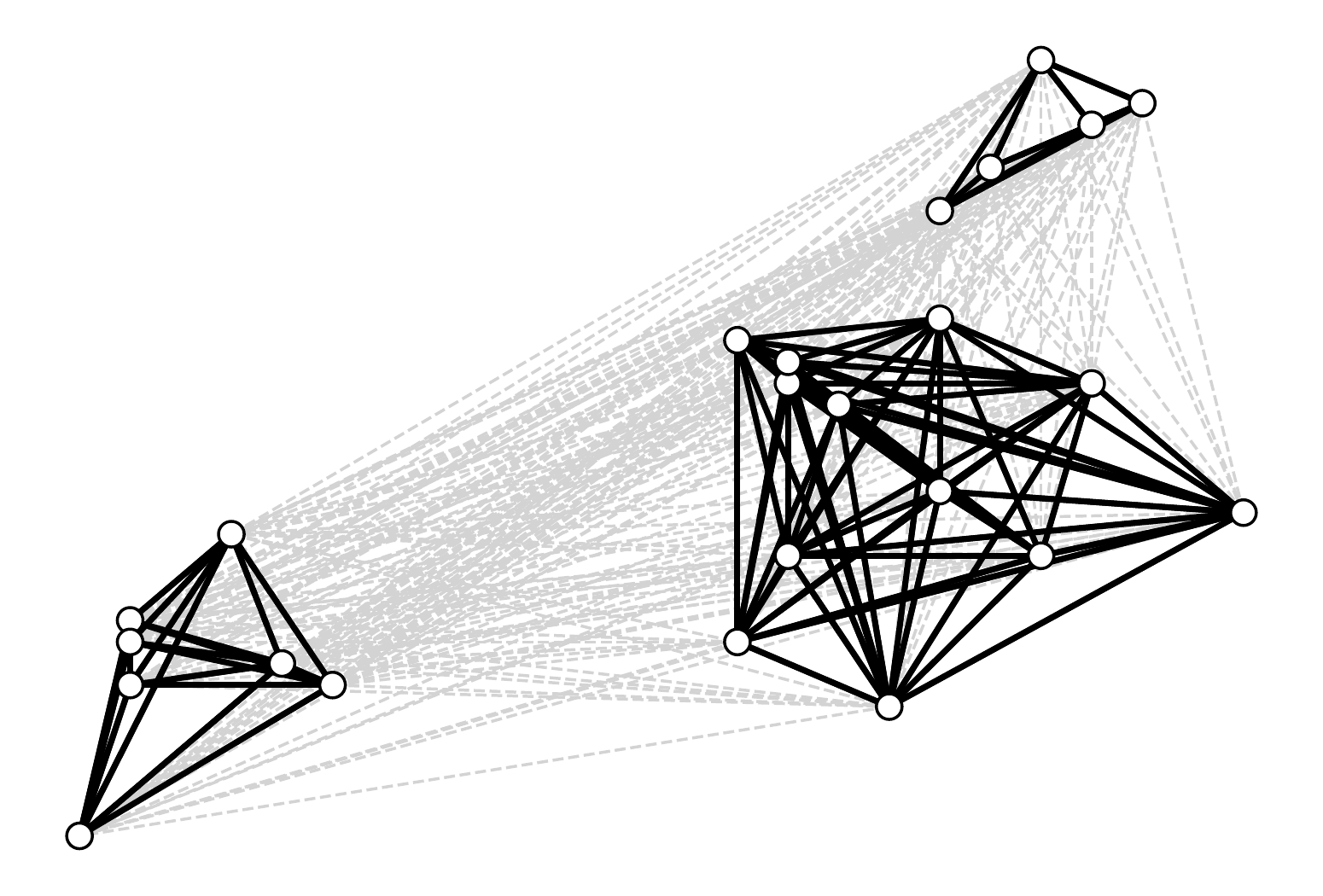} & 
            \includegraphics[width=0.25\linewidth]{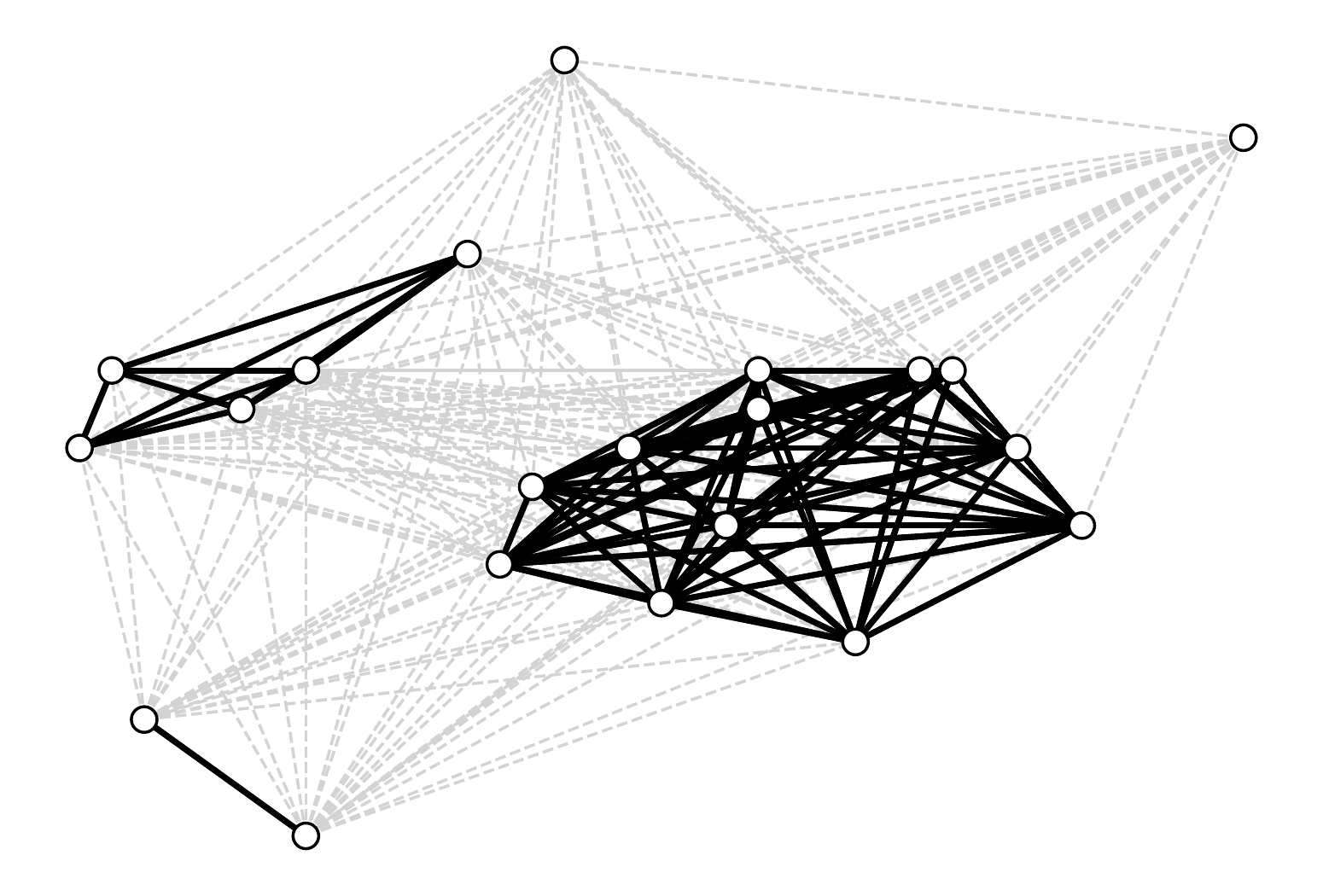} \\
            (d) OPT 1. &
            (e) OPT 2. & 
            (f) OPT 3. \\
    
        \end{tabular}
        \caption{
            Samples of the IrisMP Graph Dataset.
            (a)-(c) depict problem instances.
            Red edges have negative weights.
            (d)-(f) depict optimal solutions.
            Gray edges are cut.
        }
        \label{fig:3-samples-iris}
        \vspace*{0.23cm}
        \setlength\tabcolsep{0pt}
        \centering
        \begin{tabular}{@{\hspace{0.3cm}}c@{}c@{}c@{}}
            \includegraphics[width=0.25\linewidth]{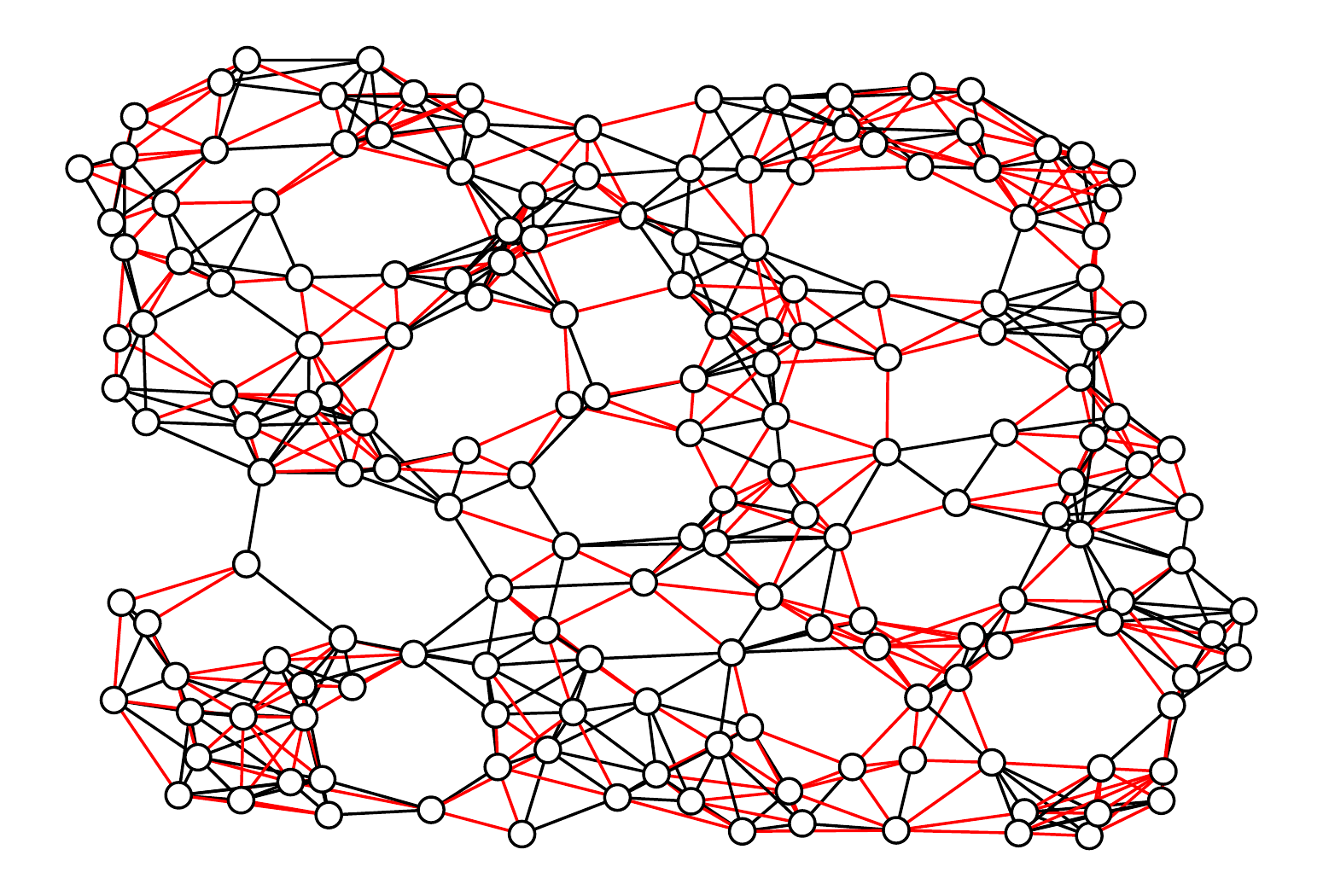} &
            \includegraphics[width=0.25\linewidth]{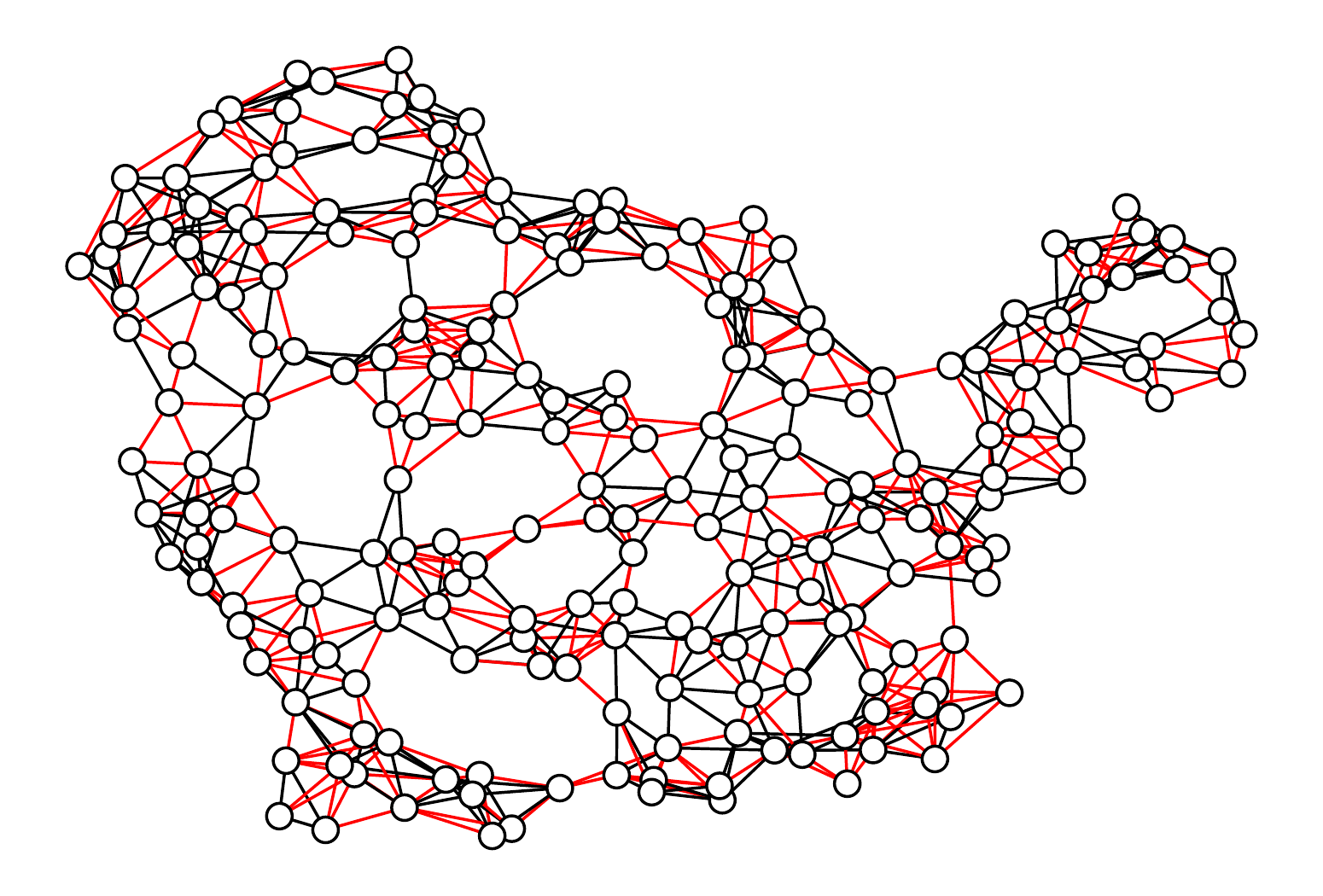} & 
            \includegraphics[width=0.25\linewidth]{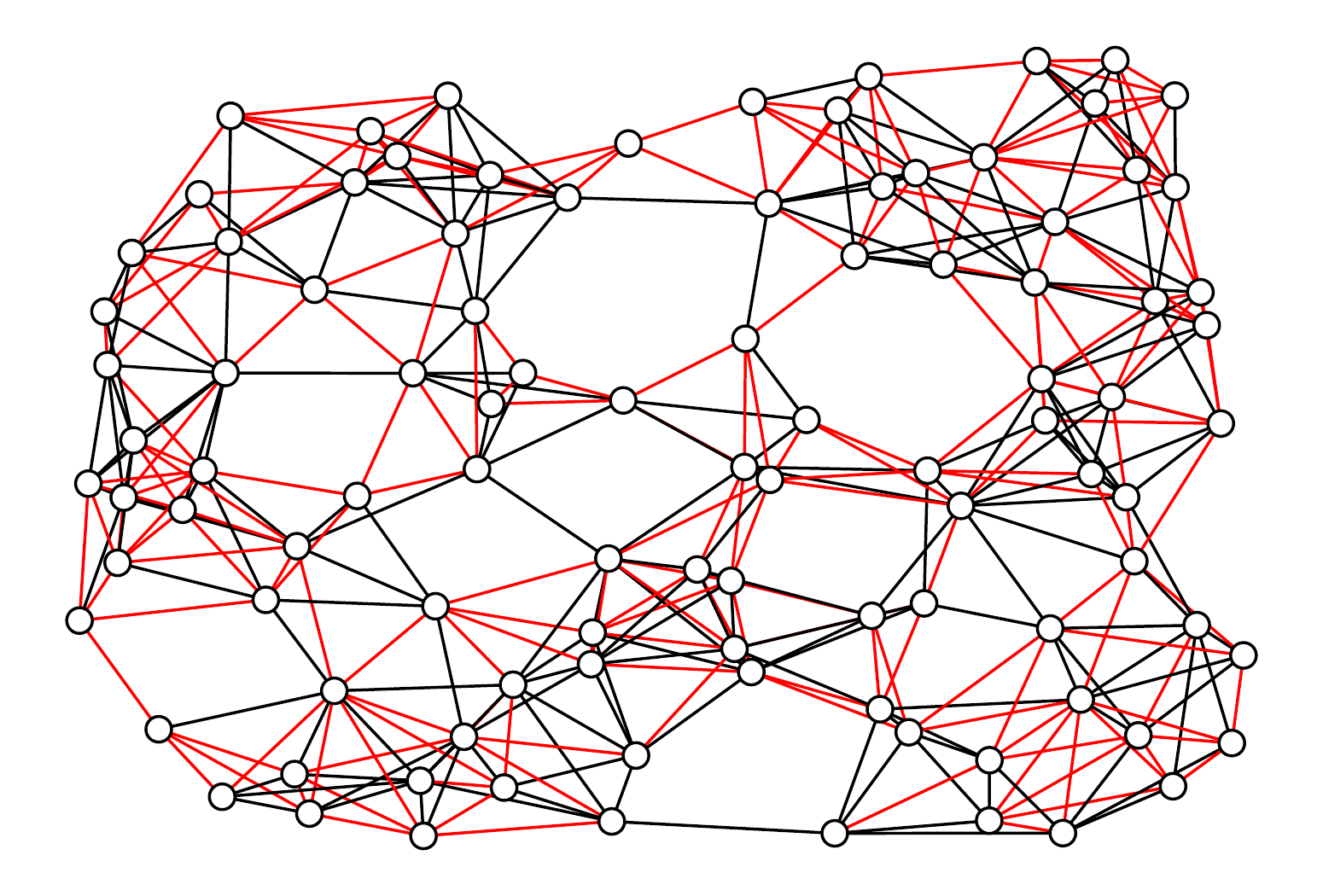} \\
            (a) Graph 1. &
            (b) Graph 2. & 
            (c) Graph 3. \\
            
            \multicolumn{3}{l}{} \\
            \includegraphics[width=0.25\linewidth]{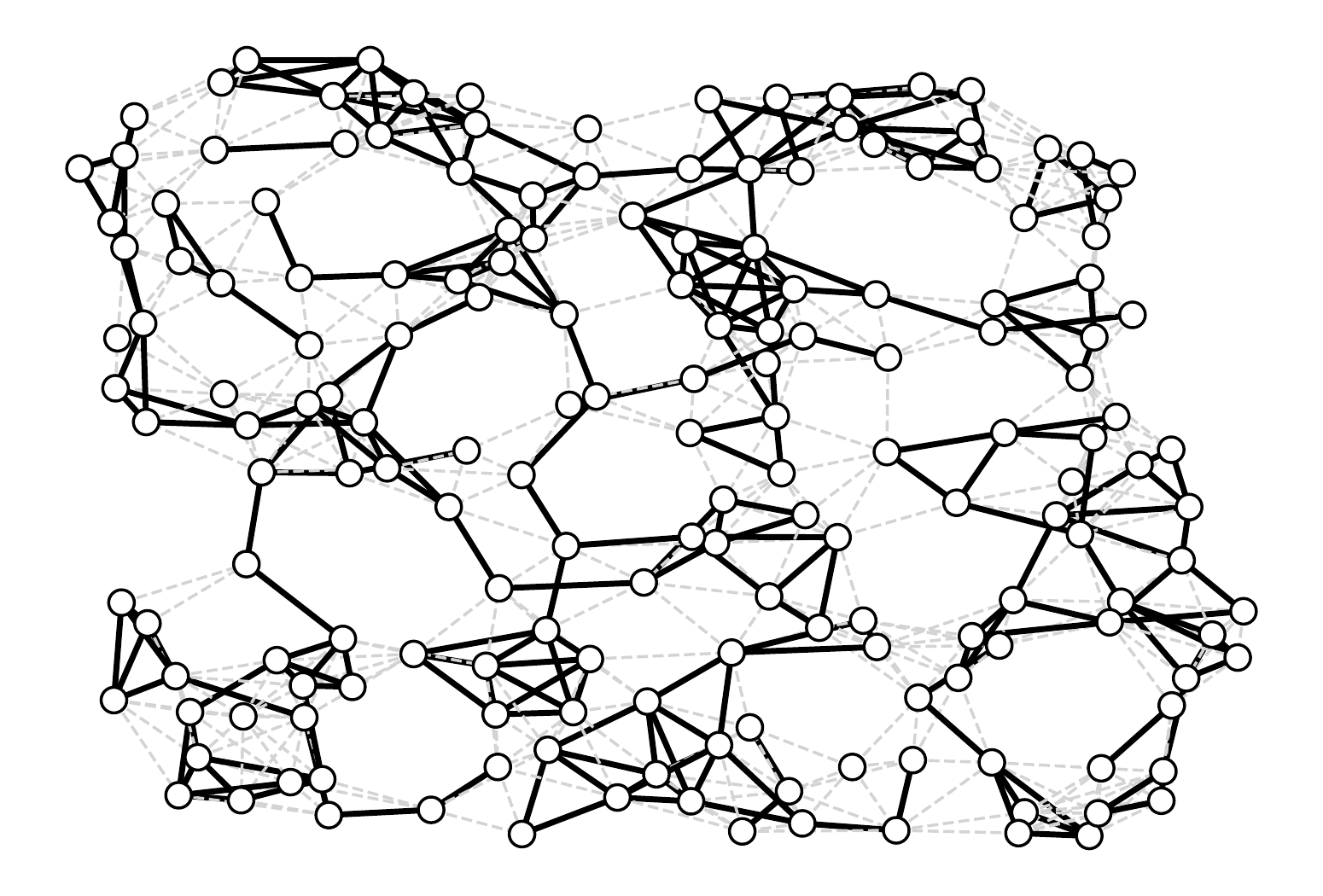} &
            \includegraphics[width=0.25\linewidth]{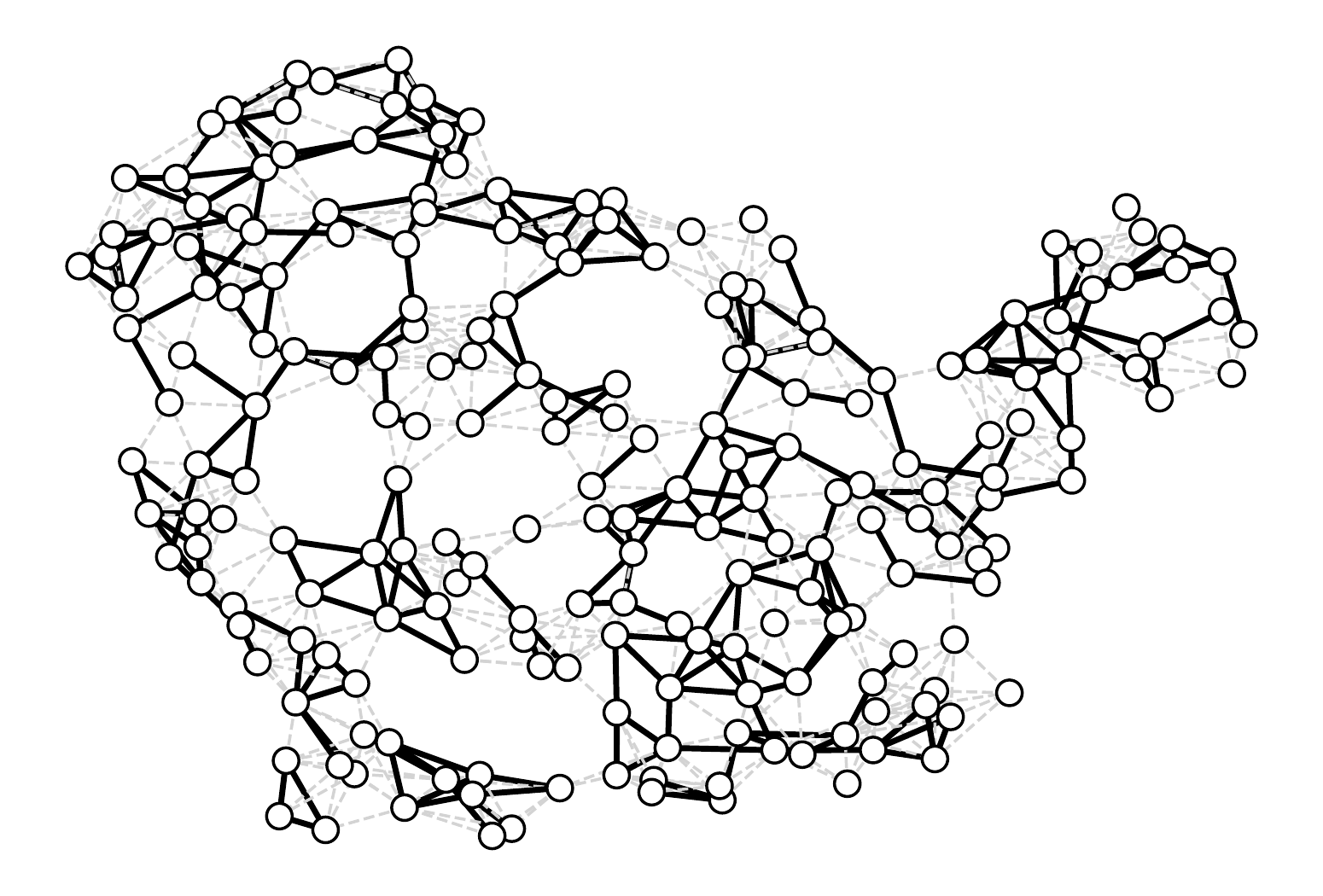} & 
            \includegraphics[width=0.25\linewidth]{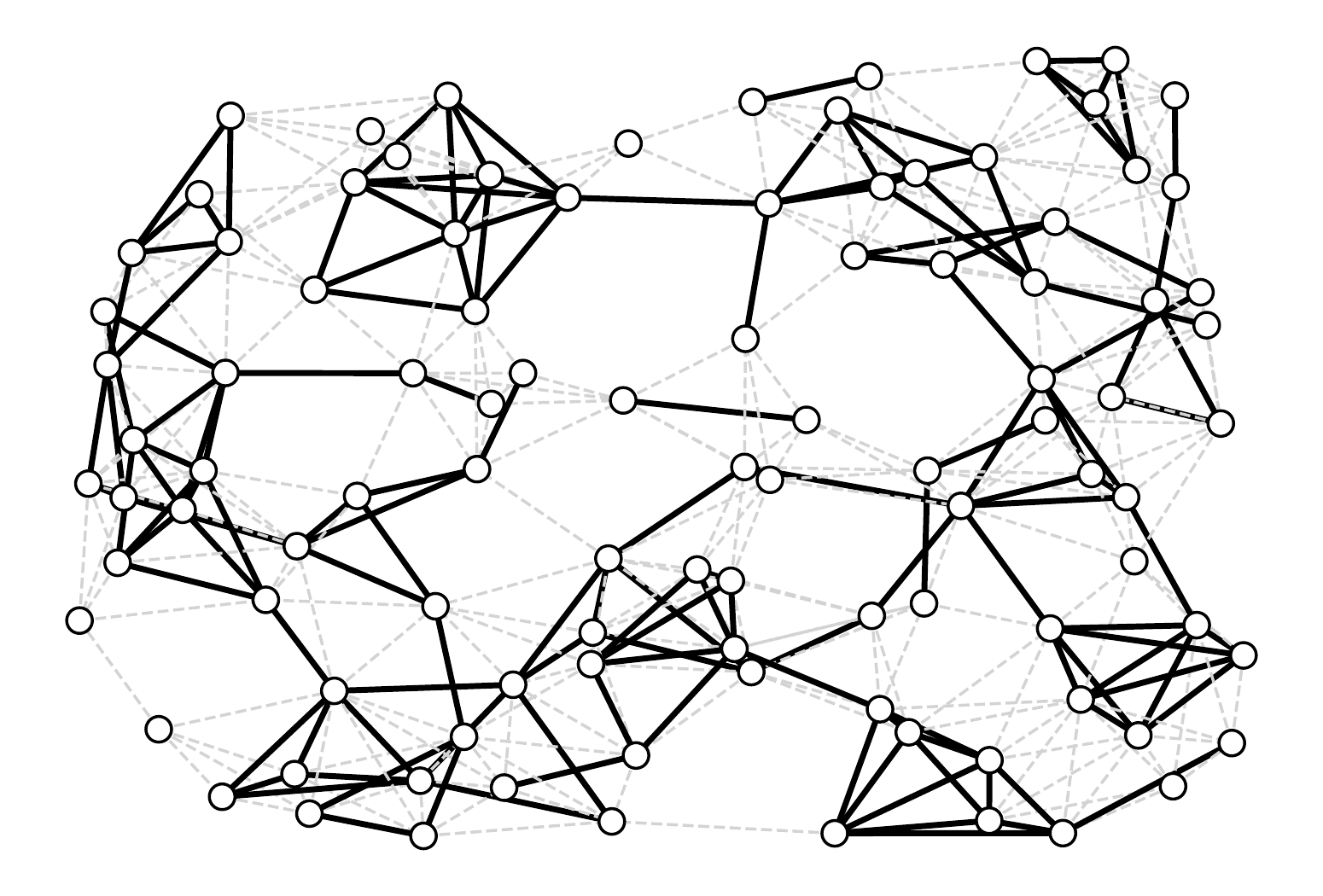} \\
            (d) OPT 1. &
            (e) OPT 2. & 
            (f) OPT 3. \\
    
        \end{tabular}
        \caption{
            Samples of the RandomMP 
            dataset.
            (a)-(c) depict problem instances.
            Red edges have negative weights.
            (d)-(f) depict optimal solutions.
            Gray edges are cut.
        }
        \label{fig:3-samples-random}
\end{figure}
\paragraph{Training Datasets}
While the multicut problem is ubiquitous in many real world applications, the amount of available annotated problem instances is scarce and domain specific. Therefore, in order train and test a general purpose model, we generated two synthetic datasets, \emph{IrisMP} and \emph{RandomMP}, with complementary connectivity statistics, of $22000$ multicut instances each.

The first dataset, IrisMP, consists of multicut instances on complete graphs based on the Iris flower dataset \cite{1936Iris}.
The generation procedure is described in the Appendix. 
Each problem instance consists of 
$120$ to $276$ edges.
Three graphs with their respective optimal solutions are depicted in
\autoref{fig:3-samples-iris}.
To complement the IrisMP dataset, we generated a second dataset that contains sparse but larger problem instances with $180$ nodes on average, called RandomMP.
The generation procedure is described in the Appendix.
Examples are depicted in
\autoref{fig:3-samples-random}.


{
\begingroup 
\afterpage{
\begin{landscape}
\setlength\tabcolsep{4.5pt} 
\begin{table*}[ht]
    \caption{
        Results on the test datasets.
        We compare different GNN variants, heuristics (GAEC) \cite{2015Heuristics}, LP-solver \cite{2015InferenceTechniques}, and ILP-solver \cite{2015InferenceTechniques}.
        The performance is evaluated as optimal objective ratio $\uparrow$ and is averaged over all datasets via harmonic mean to account for generalizability.
        The last column shows the total runtime $\downarrow$ over all datasets in milliseconds.
        OOM indicates insufficient memory.
        OOT indicates no termination within 24hrs.
        Neither OOM nor OOT are considered in the runtime (marked with *).
    }
    \label{tab:test}
    \begin{center}\begin{small}
        \begin{tabular}{@{}cc|c@{\hspace{0.1cm}}c@{\hspace{0.1cm}}c@{\hspace{0.1cm}}c@{\hspace{0.2cm}}|ccccc|c|cc@{}}
        \toprule
        && \multicolumn{4}{c|}{\textbf{Solver}} & \multicolumn{5}{c}{\textbf{Test Dataset}} & &  \multicolumn{2}{|c}{\textbf{Runtime [s]}} \\
        && Variant & Depth & $\alpha$ & $l$ & IrisMP & RandomMP & BSDS300 & CREMI & Knott & h.mean & forward & total \\
        \midrule
\multirow{15}{*}{\rotatebox[origin=c]{90}{\textbf{Proposed learned solvers}}}&
        \multirow{8}{*}{\rotatebox[origin=c]{90}{\textbf{IrisMP}}}
& \texttt{GCN\_W\_BN} & $12$ & $0.001$ & $3$ & $0.9834$ & $0.7188$ & $\mathbf{0.8912}$ & $\mathbf{0.7255}$ & $\mathbf{0.6902}$ & $\mathbf{0.7865}$ & $0.5$ & $4.4$ \\ 
&& \texttt{GIN0\_W\_BN} & $12$ & $0.01$ & $3$ & $\mathbf{0.9905}$ & $\mathbf{0.7387}$ & $0.8474$ & $0.5464$ & $0.0000$ & $0.0000$ & $0.0$ & $4.0$ \\ 
&& \texttt{Signed\_W\_BN} & $12$ & $0.01$ & $3$ & $0.9878$ & $0.2526$ & $0.6451$ & $0.5154$ & $0.3808$ & $0.4510$ & $1.3$ & $5.3$ \\ 
&& \texttt{RGGCN\_HE} & $12$ & $0.01$ & $3$ & $0.7976$ & $0.1449$ & $0.4655$ & $0.1544$ & $0.1735$ & $0.2218$ & $0.1$ & $4.1$ \\ 
&& \texttt{GT} & $12$ & $0.001$ & $3$ & $0.7940$ & $0.2964$ & $0.6360$ & $0.4037$ & $0.6038$ & $0.4836$ & $0.1$ & $4.0$ \\ 
        \cmidrule{3-14}
&& \multicolumn{4}{c@{\hspace{0.1cm}}|}{\texttt{LR}} & $0.6769$ & $0.1118$ & $0.6824$ & $0.2689$ & $0.0366$ & $0.1164$ & & N/A \\
&& \multicolumn{4}{c@{\hspace{0.1cm}}|}{\texttt{MLP}} & $0.6626$ & $0.3127$ & $0.7139$ & $0.2789$ & $0.1493$ & $0.3051$ & & N/A \\
        \cmidrule{2-14}
&        \multirow{7}{*}{\rotatebox[origin=c]{90}{\textbf{RandomMP}}}
& \texttt{GCN\_W\_BN} & $20$ & $0.01$ & $8$ & $\mathbf{0.9762}$ & $\mathbf{0.9041}$ & $\mathbf{0.9204}$ & ${0.8440}$ & $\mathbf{0.7870}$ & $\mathbf{0.8815}$ & $0.9$ & $4.8$ \\ 
&& \texttt{GIN0\_W\_BN} & $20$ & $0.01$ & $8$ & $0.9528$ & $0.8693$ & $0.9109$ & $0.4812$ & $0.0000$ & $0.0000$ & $0.0$ & $4.0$ \\ 
&& \texttt{Signed\_W\_BN} & $20$ & $0.01$ & $8$ & $0.9709$ & $0.8695$ & $0.8825$ & $0.4653$ & $0.6408$ & $0.7120$ & $2.3$ & $6.3$ \\ 
&& \texttt{RGGCN\_HE} & $20$ & $0.01$ & $8$ & $0.9703$ & $0.8787$ & $0.8352$ & $0.5593$ & OOM  & - & $0.1$ & $2.7$* \\ 
        \cmidrule{3-14}
&& \multicolumn4{c@{\hspace{0.1cm}}|}{\texttt{LR}} & $0.8035$ & $0.3938$ & $0.7958$ & $\mathbf{0.9260}$ & $0.7335$ & $0.6681$ & & N/A \\
&& \multicolumn4{c@{\hspace{0.1cm}}|}{\texttt{MLP}} & $0.8985$ & $0.3099$ & $0.6804$ & $0.4845$ & $0.1517$ & $0.3457$ & & N/A \\
        \midrule
        \multirow{5}{*}{\rotatebox[origin=c]{90}{}}
&& \multicolumn{4}{c@{\hspace{0.1cm}}|}{\texttt{GAEC}} & $0.9836$ & $0.9780$ & $0.9997$ & $0.9958$ & $0.9968$ & $0.9907$ &  & $23.2$ \\
&& \multicolumn{4}{c@{\hspace{0.1cm}}|}{\texttt{Time-bounded GAEC}} & $0.3642$ & $0.0034$ & $0.0000$ & $0.1516$ & $0.0000$ & $0.0000$ &  & $6.3$ \\
&& \multicolumn{4}{c@{\hspace{0.1cm}}|}{\texttt{LP solver}} & $0.9882$ & $0.9525$ & $0.9979$ & $0.9998$ & OOT & - &  & $31~918.8$* \\
&& \multicolumn{4}{c@{\hspace{0.1cm}}|}{\texttt{ILP solver}} & $1.0000$ & $1.0000$ & $1.0000$ & $1.0000$ & $1.0000$ & $1.0000$ &  & $24~361.2 $ \\
        \bottomrule
        \end{tabular}
    \end{small}\end{center}
\end{table*}
\end{landscape}
}
\endgroup}

\section{Experiments}
\label{sec:results}
We evaluate all models trained on IrisMP and RandomMP and provide runtime as well as objective value evaluations, where we compare the proposed GCN to GIN and SGCN-based, edge-weight enabled models (see appendix for details) as well as to RGGCN \cite{2019GatedGCNTSP} and GTN \cite{2020GraphTransformer}. Then, we provide an ablation study on the proposed GCN-based edge representation mapping and the multicut loss.

\subsection{Evaluation on Test Data}
We evaluate our models on three segmentation benchmarks: a graph-based image segmentation dataset \cite{2011Probabilistic} based on the Berkeley Segmentation Dataset (BSDS300) \cite{BSDS300} consisting of $100$ test instances, a graph-based volume segmentation dataset \cite{2012ClosedSurface} (Knott3D) containing $24$ volumes, and $3$ additional test instances based on the challenge on Circuit Reconstruction from Electron Microscopy Images (CREMI) \cite{2017CREMI} that contains volumes of electron microscopy images of fly brains.
BSDS300 and Knott3D instances are available as part of a benchmark containing discrete energy minimization problems, called OpenGM \cite{2015InferenceTechniques}.


\vspace{0.1cm}
\noindent\textbf{Implementation Details\hspace{0.1cm}}
We train the proposed MPNN-based solvers with (adapted) GCN, GIN, SGCN, RGGCN and GT backbones in different settings, where we uniformly set the node representation dimensionality to $128$.
We set the depth of the MPNN to $12$ for IrisMP and $20$ for RandomMP.
CCL is applied with $\alpha \in \{0,0.01,0.001\}$ and chordless cycle length up to $8$. All of our experiments are conducted on MEGWARE Gigabyte G291-Z20 servers with NVIDIA Quadro RTX 8000 GPUs.
If not stated otherwise, we consider as performance metric $m$ the \emph{optimal objective ratio} achieved, hence $m=\max(0,w(\mathbf{y})/w(\mathbf{\tilde{y}})) \in [0,1]$.

\paragraph{Results}
In \autoref{tab:test}, we show the results on all test datasets of the best models based on the evaluation objective value after rounding, and thereby compare models trained on IrisMP and models trained on RandomMP. 
In general, sparser problems (RandomMP and established test datasets) are harder for the solvers to generalize to. This is likely due to the longer chordless cycles that the model needs to consider to ensure feasibility. Overall, our GCN-based model provides the best generalizability over all test datasets both when trained on IrisMP and RandomMP. 
We compare the GNN-based solvers to different baselines.
First, we train logistic regression (LR) and MLPs as edge classifiers directly on the training data (concatenation of node features and edge weights).
All our learned models outperform these baselines significantly.
This indicates that MPNNs provide meaningful topological information to the edge classifier that facilitates solving MP instances.
Second, we compare against Branch \& Cut LP and ILP solvers as well as GAEC.
In terms of objective value, GCN-based solvers are on par with heuristics and LP solvers on complete graphs, even when trained on sparse graphs.
On general graphs, ILP solvers and GAEC issue lower energies, and, as expected, training on complete graphs does not generalize well to sparse graphs.
However, the wall-clock runtime comparison shows that GCN-based solvers are faster by an order of $10^3$ than ILP and LP solvers.
They are also significantly faster than the fast and greedy GAEC heuristic.
We further compared to a time-constrained version of GAEC, where we set the available time budget to the runtime of the GCN-based solver.
The result shows that the trade-off between smaller energies and smaller runtime is in favor of the GCN-based solver.
In the Appendix, we report additional experiments for our proposed GCN-based model on domain specific training and show that task specific priors can be learned efficiently from only few training samples.

{
\begingroup
\setlength\tabcolsep{4.5pt} 
\begin{table}[t]
    \caption{
        Wall-clock runtime $\downarrow$ and objective values $\downarrow$ of MPNN-based solver vs. GAEC, LP and ILP on a growing, randomly-generated graph.
        OOT indicates no termination within 24hrs.
    }
    \label{tab:runtimes_gcn}
    \centering
            \begin{scriptsize}
            \begin{tabular}{r|rr|rr|rr|rr}
                \toprule
                & \multicolumn{2}{c|}{GAEC} & \multicolumn{2}{c|}{LP} & \multicolumn{2}{c|}{ILP} & \multicolumn{2}{c}{GCN\_W\_BN} \\
                Nodes & [ms] & Objective & [ms] & Objective & [ms] & Objective & [ms] & Objective \\
                \midrule
      $10^1$  &   $\mathbf{0}$ &        $-29$ &      ${6}$    &        $-24$  &          $11$ &     $-30$  &           $29$  &        $-29$ \\
     $10^2$  & $\mathbf{4}$ &       $-327$ &        ${191}$  &       $-246$  &         $273$ &    $-330$  &           $26$  &       $-276$ \\
    $10^3$  &$\mathbf{24}$ &      $-3051$ &       ${6585}$  &      $-2970$  &        $1299$ &   $-3093$  &           $29$  &      $-2643$ \\
  $10^4$  &       $228$  &    $-32~264$ &       $688~851$ &    $-31~531$  &      $18~604$ &  $-32~682$ &   $\mathbf{78}$ &    $-27~552$ \\
 $10^5$  &      $2534$  &   $-323~189$ &    \multicolumn{2}{c|}{OOT}     &   $2~171~134$ & $-328~224$ &  $\mathbf{557}$ &   $-269~122$ \\
$10^6$ &    $35~181$  & $-3~401~783$ &    \multicolumn{2}{c|}{OOT}     &   \multicolumn{2}{c|}{OOT} & $\mathbf{8713}$ & $-2~182~589$ \\
                \bottomrule
            \end{tabular}
            \end{scriptsize}
\end{table}
\endgroup
}
Next, we conduct a scalability study on random graphs with increasing number of nodes, generated according to the RandomMP dataset. Results are shown in \autoref{tab:runtimes_gcn}.
While the GAEC is fast for small graphs, the GCN-based solver scales better and returns solutions significantly faster for larger graphs.
LP and ILP solvers are not able to provide solutions within 24hrs for larger instances.
It is noteworthy that GNN-based solvers spend $75$-$99\%$ of their runtime rounding the solutions.
Hence, GNN-based solvers are already more scalable and still have a large potential for improvement in this regard, while GAEC and LP/ILP solvers are already highly optimized for runtime.

Next, we ablate on the {GCN} aggregation functions, loss and network depths.

\paragraph{Edge-weighted GCNs}

First, we determine the impact of each adjustment to the GCN update function.
In \autoref{tab:gcn-ablation} we show the results of this ablation study.
While vanilla GCN is not applicable in the MP setting, simply removing the Laplacian from \autoref{eq:3-gcn} provides a first baseline.
We observe that adding edge weights ($w_{u,v}$) to \autoref{eq:3-gcn} improves the performance on the test split of the training data substantially.
However, the model is not able to generalize to different graph statistics.
By adding the signed normalization term ($\left({\overline{\text{deg}}(u)\overline{\text{deg}}(v)}\right)^{-1/2}$) we arrive at \autoref{eq:3-gcn-signed}, achieving improved generalizability.
Removing edge weights from \autoref{eq:3-gcn-signed} deteriorates performance and generalizability.
Thus both changes are necessary to enable GCN in the MP setting.

\begin{figure}[t]
    \centering
    \setlength\tabcolsep{0pt} 
    \begin{tabular}{ccc}
        \includegraphics[height=3cm]{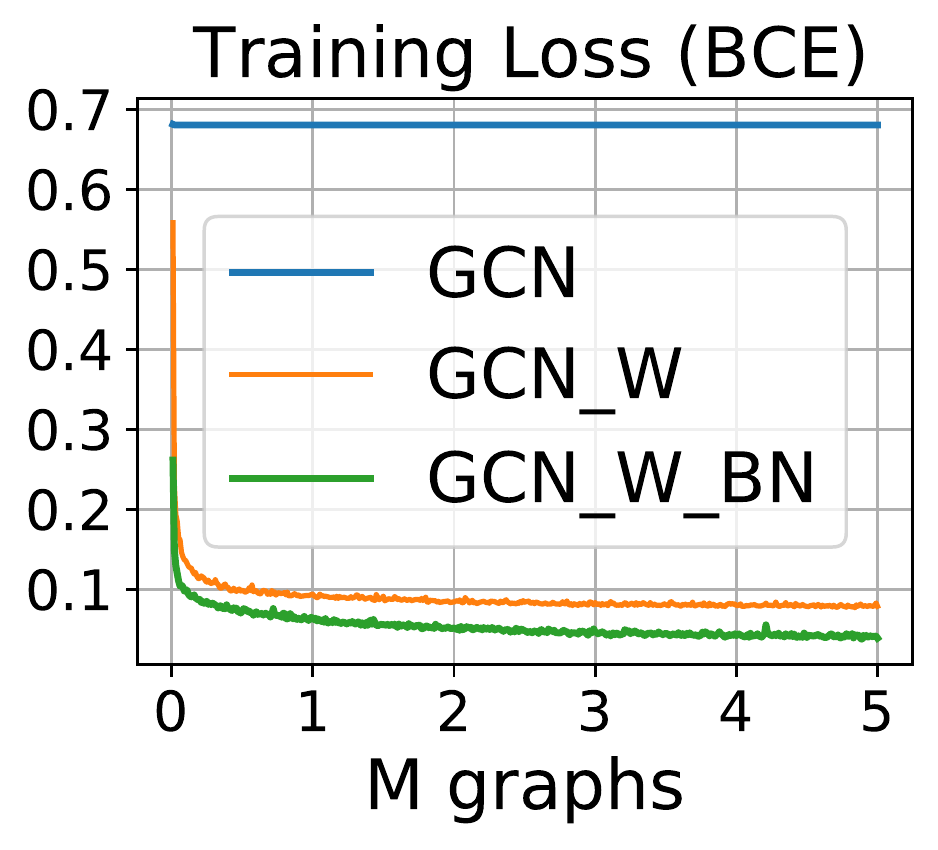} & 
        \includegraphics[height=3cm]{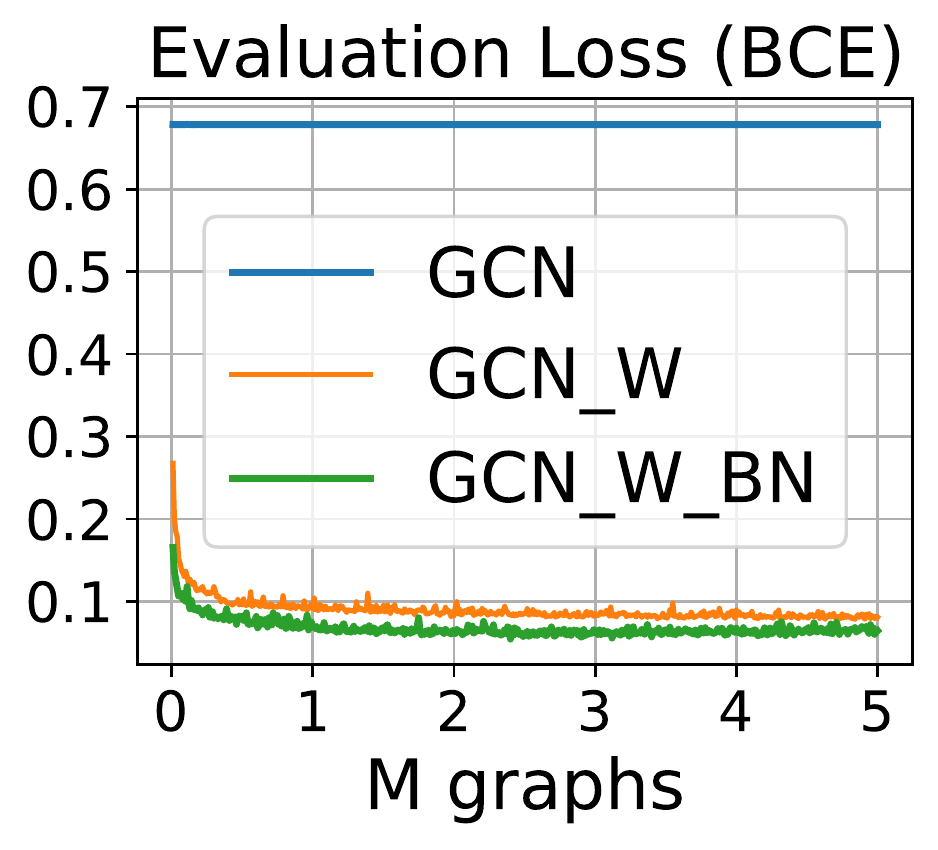} & 
        \includegraphics[height=3cm]{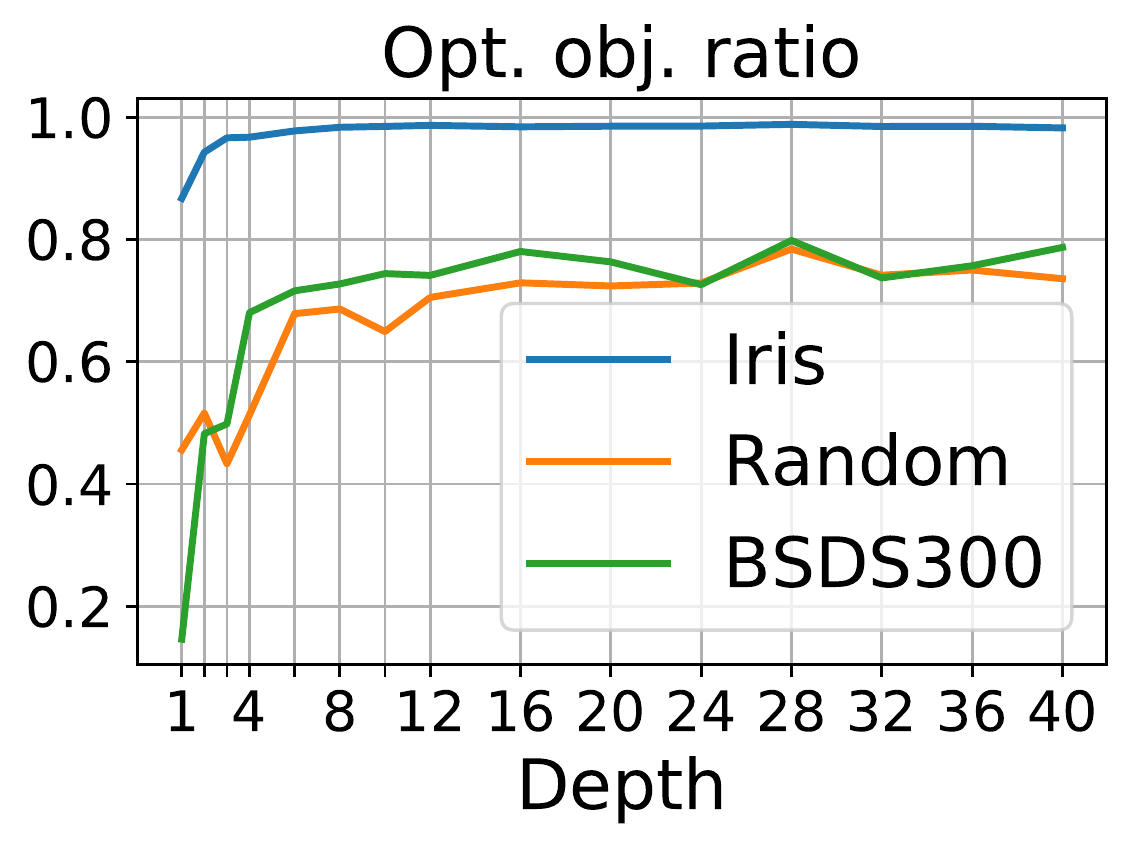} \\
        (a) Training loss. &
        (b) Evaluation loss. &
        (c) Network Depth.
    \end{tabular}
    \caption{
        (a) Training and (b) evaluation loss while training variants of {GCN} on IrisMP.
        Each plot compares the variants with (GCN\_W) and without (GCN) edge weights in the aggregation, and GCN\_W with batch normalization (GCN\_W\_BN).
        (c) Results in terms of optimal objective ratio on the evaluation data when training GCN\_W\_BN with varying depths.
    }
    \label{fig:4-training-iris-loss}
\end{figure}

Additionally, we compare GCNs with edge weights and \emph{signed} graph Laplacian normalization, trained with batch normalization, to the plain GCN model~\cite{2017GCN}.
To this end, we train on the IrisMP dataset and set the model width to $128$ and its depth to $12$.
Here, we set $\alpha=0$, hence, we do not apply {CCL}.
\autoref{fig:4-training-iris-loss} shows the results of this experiment. The corresponding plots for SGCN and GIN are given in the Appendix.
The variants with edge features achieve lower losses than without edge features, and batch normalization improves the loss further.
In fact, the original GCN is not able to provide any meaningful features for the edge classification network. The proposed extensions enable these networks to find meaningful node representations for the multicut problem.

\begin{figure}[t]
    \centering
    \setlength\tabcolsep{0pt} 
    \begin{tabular}{ccc}
        \includegraphics[height=4.0cm]{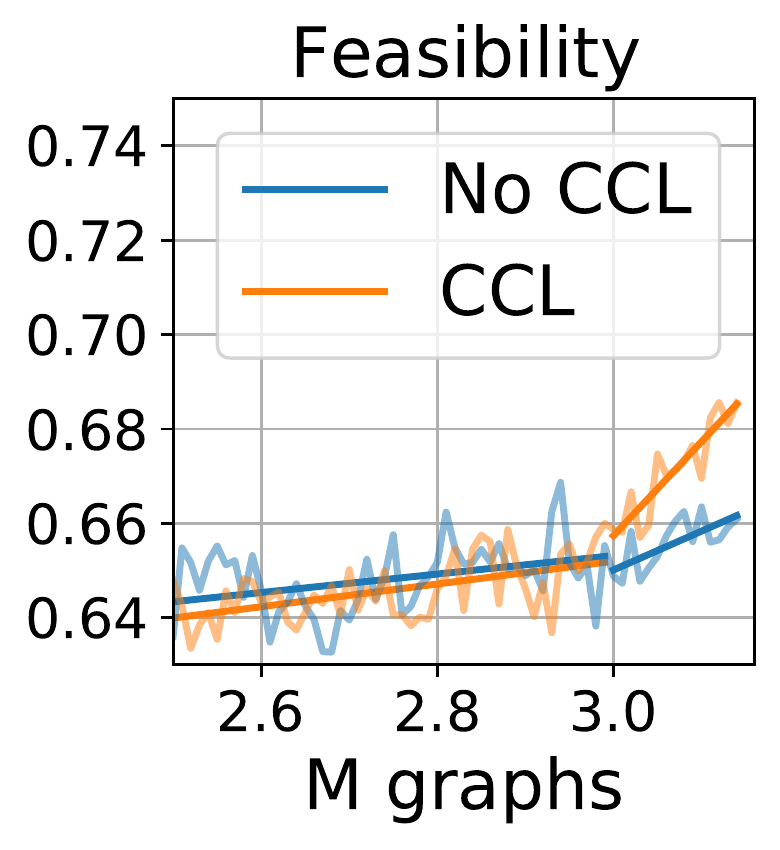} &
        \includegraphics[height=4.0cm]{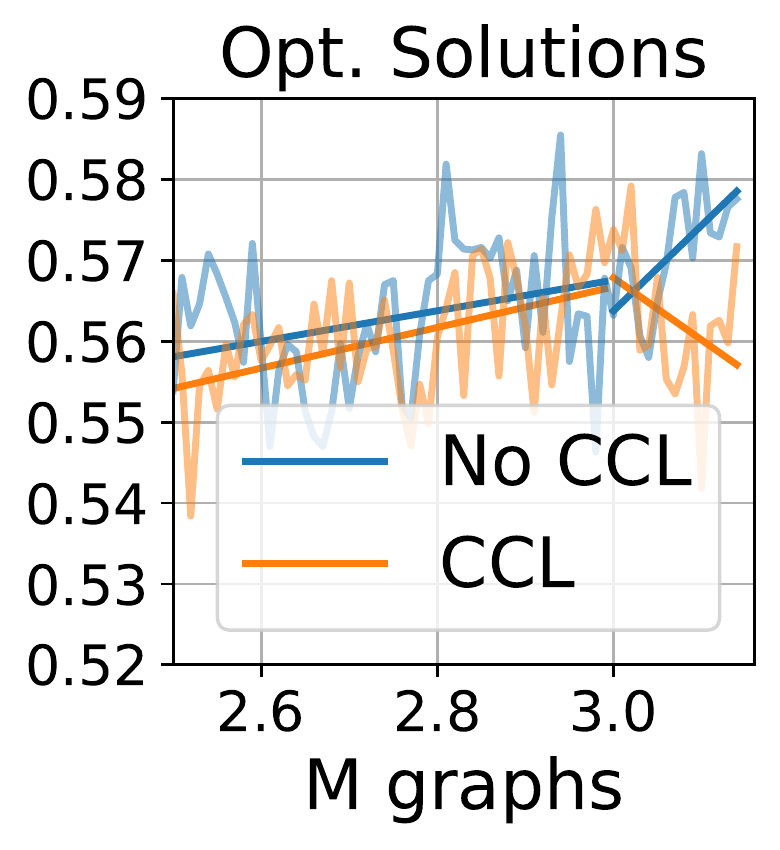} & 
        \includegraphics[height=4.0cm]{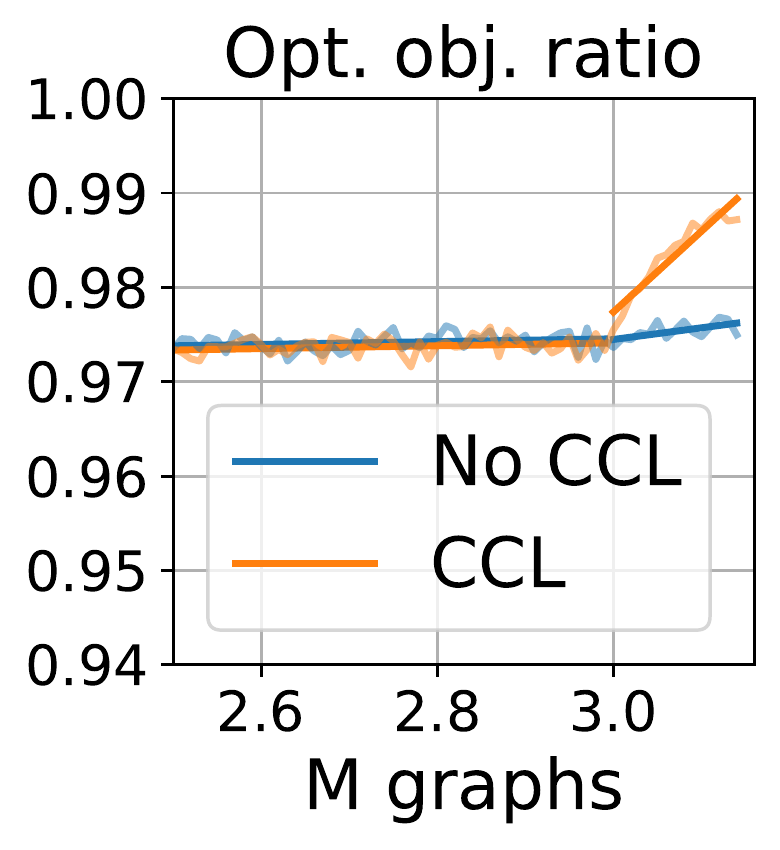} \\
        
        (a) Feasibility. &
        (b) Optimality. &
        (c) Objective. 
        \end{tabular}
    \caption{
        (a) Ratio of feasible solutions before repairing,
        (b) Ratio of optimal solutions, and 
        (c) Optimal objective ratio, for GCN\_W\_BN on RandomMP, applying CCL after $3M$ instances.
    }
    \label{fig:4-training-iris-eta-gcn}
\end{figure}
\subsection{Ablation Study}

{
\begingroup
\setlength\tabcolsep{4.5pt}
\begin{table}
    \caption{
        Ablation study with GCN \cite{2017GCN} trained on IrisMP without CCL..
        Additional comparison to vanilla versions of GIN0 \cite{2019GIN}, and MPNN \cite{2017MPNN}.
        We report the performance on the test data in terms of optimal objective ratio $\uparrow$.
    }
    \label{tab:gcn-ablation}
    \centering
        \begin{small}
            \begin{tabular}{p{2.20cm}c@{\hspace{0.1cm}}c@{\hspace{0.1cm}}c@{\hspace{0.1cm}}c@{\hspace{0.1cm}}c@{\hspace{0.1cm}}}
                \toprule
                Variant & IrisMP & RandMP & BSDS300 & CREMI & Knott \\
                \midrule
                GCN  &\multicolumn{5}{c@{\hspace{0.1cm}}}{
                Not applicable: Laplacian may not exist.
                } \\
                - Laplacian & 0.41 & 0.18 & 0.00 & 0.49 & 0.00 \\
                + edge weights & 0.95 & 0.18 & 0.40 & 0.57 & 0.19 \\
                + signed norm. & \textbf{0.96} & \textbf{0.67} & \textbf{0.75} & \textbf{0.74} & \textbf{0.68} \\
                \midrule
                = GCN\_W & \textbf{0.96} & \textbf{0.67} & \textbf{0.75} & \textbf{0.74} & \textbf{0.68} \\
                - edge weights & 0.64 & 0.05 & 0.00 & 0.48 & 0.00 \\
                \midrule
                GIN0 & 0.41 & 0.04 & 0.07 & 0.48 & 0.00 \\
                MPNN & 0.93 & 0.45 & 0.48 & 0.49 & 0.06 \\
                \bottomrule
            \end{tabular}
        \end{small}
\end{table}
\endgroup
}
\paragraph{Number of Convolutional Layers}
Next, we evaluate the effect of depth of the GCN model when trained on the IrisMP dataset and evaluated on IrisMP, RandomMP as well as BSDS300.
\autoref{fig:4-training-iris-loss}(c) shows the results after varying the depth in increasing step sizes up to a depth of $40$.
The results suggest that increasing the depth improves the objective value up to a certain point.
In the case of IrisMP graphs with diameter $1$ and lengths of chordless cycles of at most $3$, increasing the depth beyond $10$ has no obvious effect.
This is an important observation, because \cite{2018DeeperInsights} raise concerns that GCN models can suffer from over-smoothing such that 
learned representations might become indistinguishable.

\paragraph{Cycle Consistency Loss}
Here, we evaluate the effect of applying the cycle consistency loss from Eq.~\eqref{eq:ccloss} by comparing models where CCL is applied after $3M$ instances to models solely trained without CCL.
Figures \ref{fig:4-training-iris-eta-gcn}(a) and (b) show the progress of the ratio of feasible solutions and ratio of optimal solutions found during training.
As soon as {CCL} is applied, the ratio of feasible solutions increases while the ratio of optimal solutions decreases.
Hence, CCL induces a trade-off between finding feasible and optimal solutions, where the model is forced to find feasible solutions to avoid the penalty, and as a consequence, settles for suboptimal relaxated solutions.
However, the objective value after rounding improves, which is most relevant because these values correspond to feasible solutions. This indicates that the model's upper bound on the optimal energy is higher while the relaxation is tighter when CCL is employed. 
See the Appendix for an ablation on $\alpha$ in~\autoref{eq:ccloss}.

\paragraph{Meaningful Embeddings}
In \autoref{fig:5-embeddings} we visualize the node embedding space given by our best performing model on an IrisMP instance.
Plotting the cosine similarity between all nodes reflects the resulting clusters.
This shows that the model is able to distinguish nodes based on their connectivity.
We show further examples in the Appendix.

\section{Conclusion}
In this paper, we address the minimum cost multicut problem using feed forward MPNNs.
To this end, we provide appropriate model and training loss modifications.
Our experiments on two synthetic and two real datasets with various GCN architectures show that the proposed approach provides highly efficient solutions even to large instances and scales better than highly optimized primal feasible heuristics (GAEC), while providing competitive energies.
Another significant advantage of our learning-based approach is the ability to provide gradients for downstream tasks, which we assume will inherently improve inferred solutions.

\bibliography{ref}
\bibliographystyle{splncs04}

\newpage
\appendix
\section{Appendix}

In this supplementary material, we provide several additional details, ablations and visualizations. We provide
\begin{itemize}
\item an example graph from an image segmentation problem, providing some intuition on the practical quality of results; 
\item details on the generation of the contributed training datasets IrisMP and RandomMP;
\item adapted update functions for GCN, for GIN and SGCN that we used for the evaluation in Table 1 of the main paper, as well as training and evaluation losses for these models with and without the proposed modifications and batch norm;
\item experiments for domain specific finetuning of our models. While the domain specific training data is very scarce, these experiments show the promis of learnable multicut solvers;
\item an ablation study on the choice of the hyperparamter $\alpha$, that weights the two loss terms in Eq.(8);
\item additional embedding space visualizations, similar to Fig.~1 in the main paper;
\item additional details to our training settings.
\end{itemize}

\section{Multicut Segmentation Example}
\begin{figure}
  \begin{tabular}{@{}c@{}c@{}}
        \includegraphics[width=0.49\linewidth]{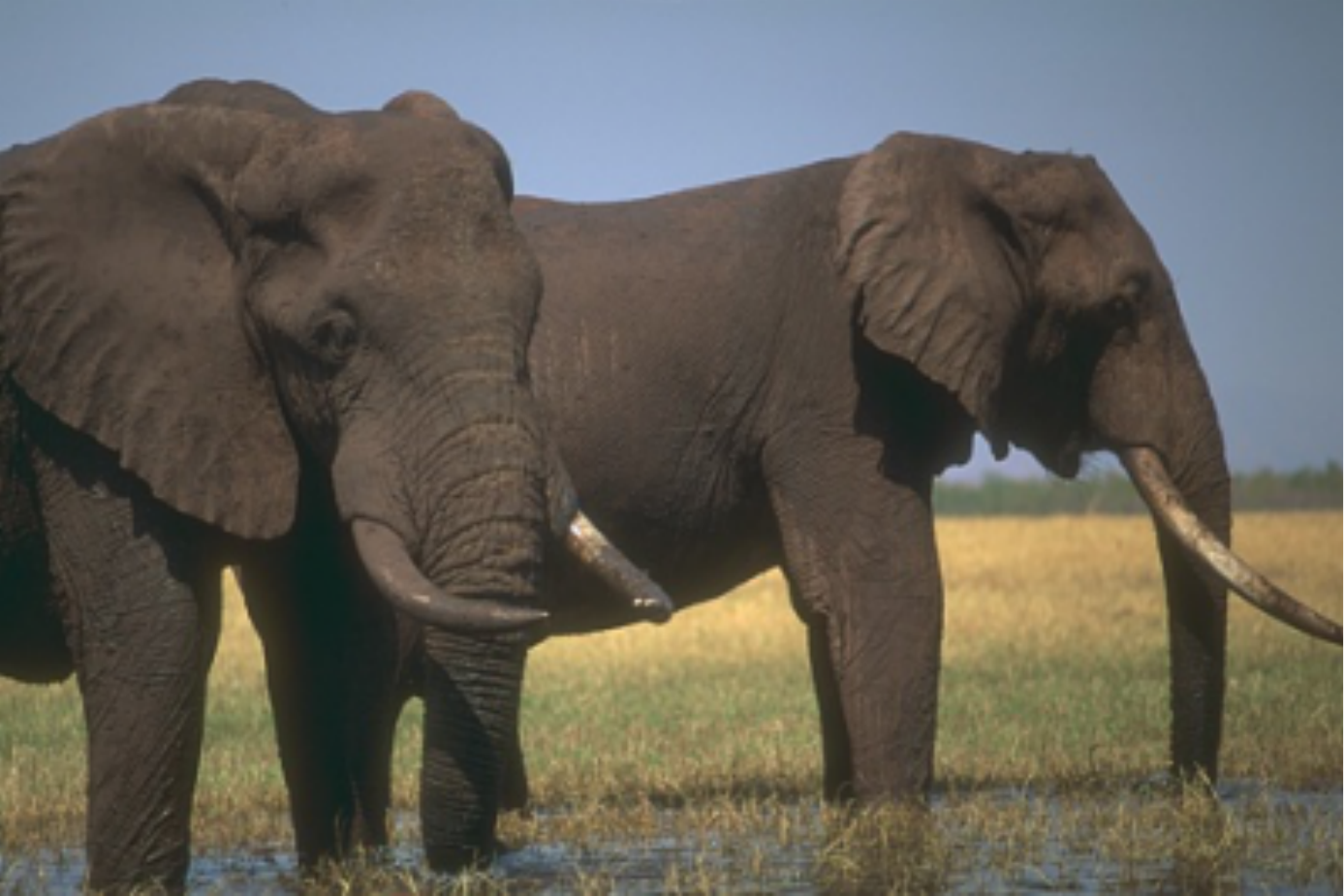}& \includegraphics[width=0.49\linewidth]{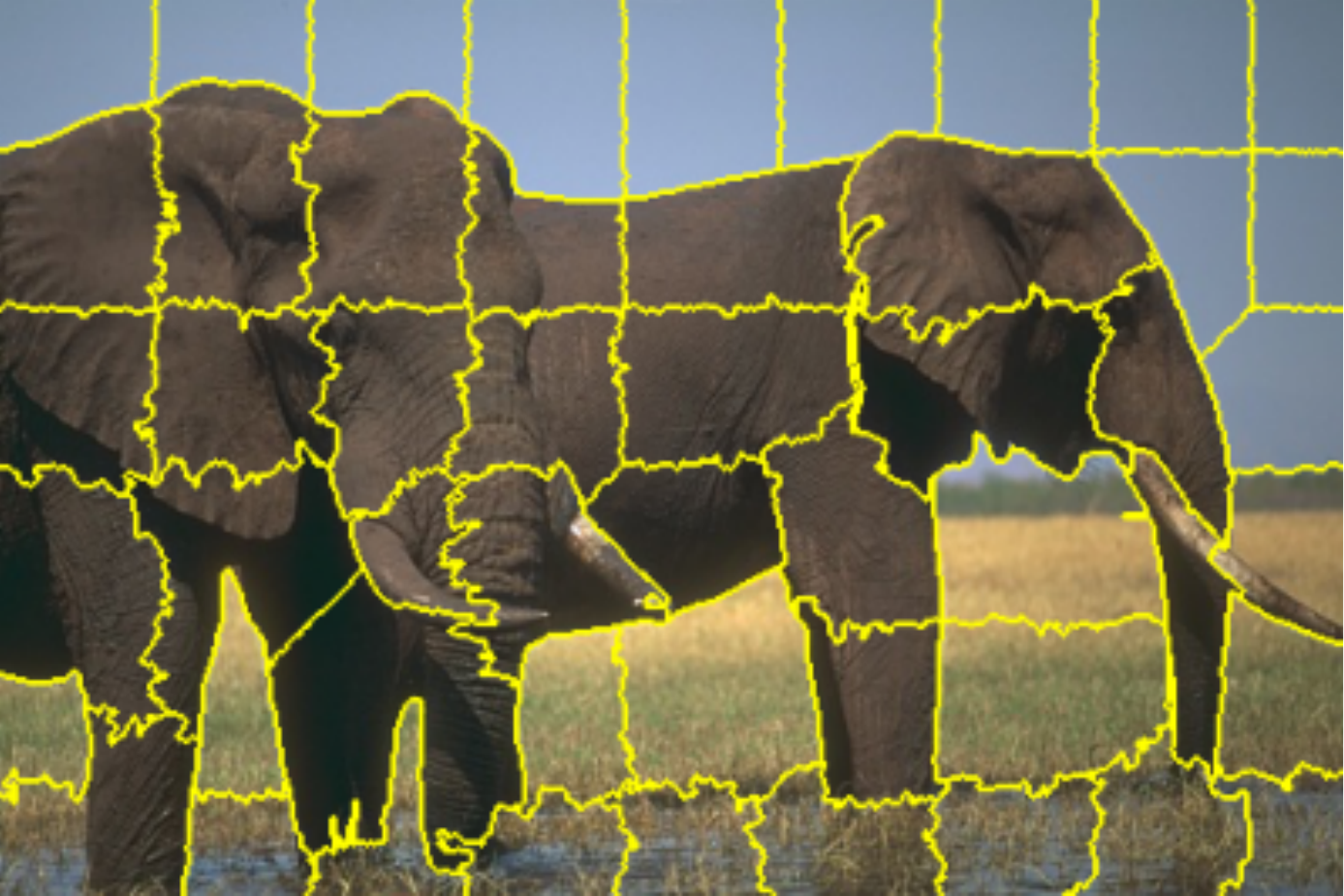}\\
        a) Original image.&
      b) Watershed transformed.\\
       
        \includegraphics[width=0.49\linewidth]{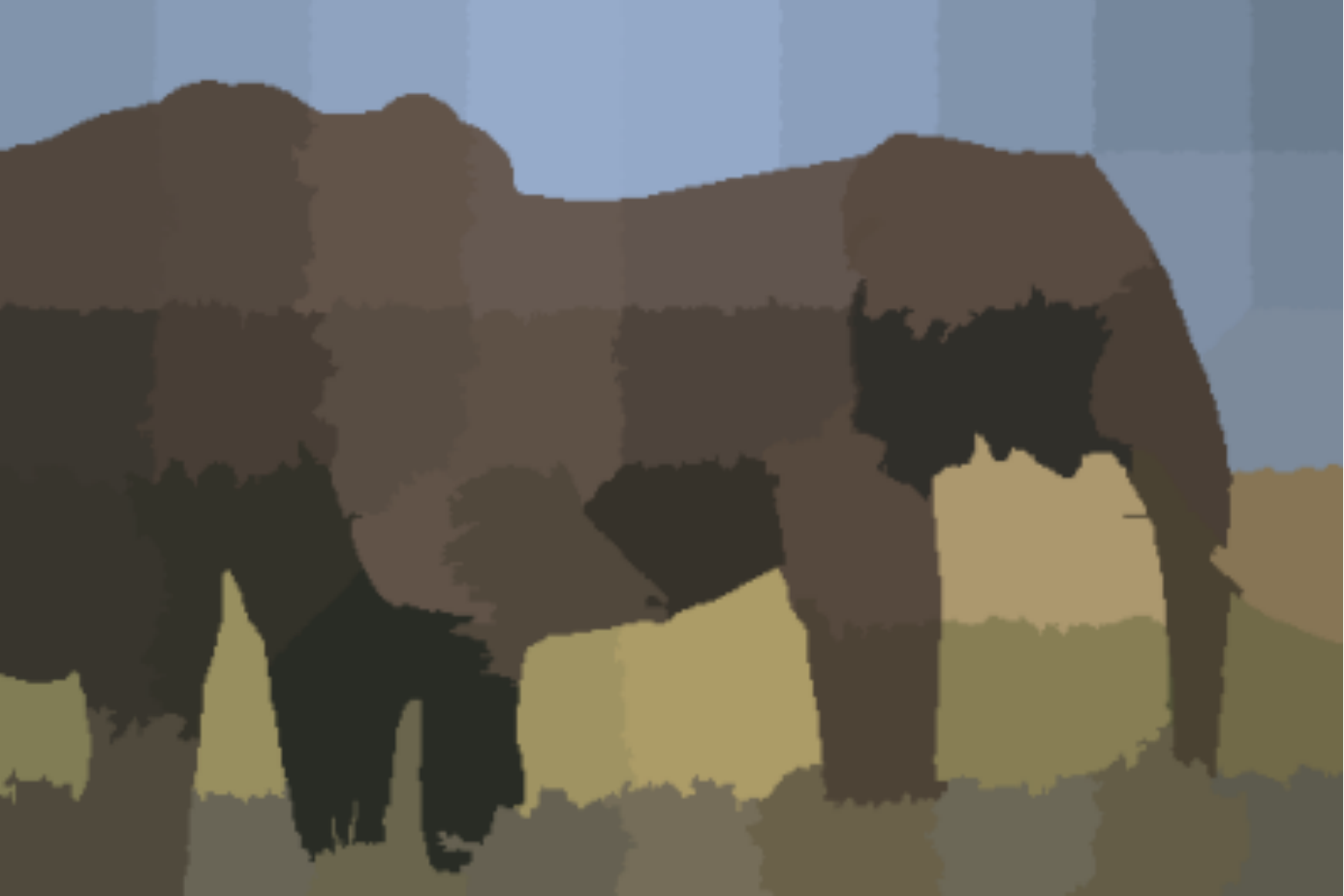}&
        \includegraphics[width=0.49\linewidth]{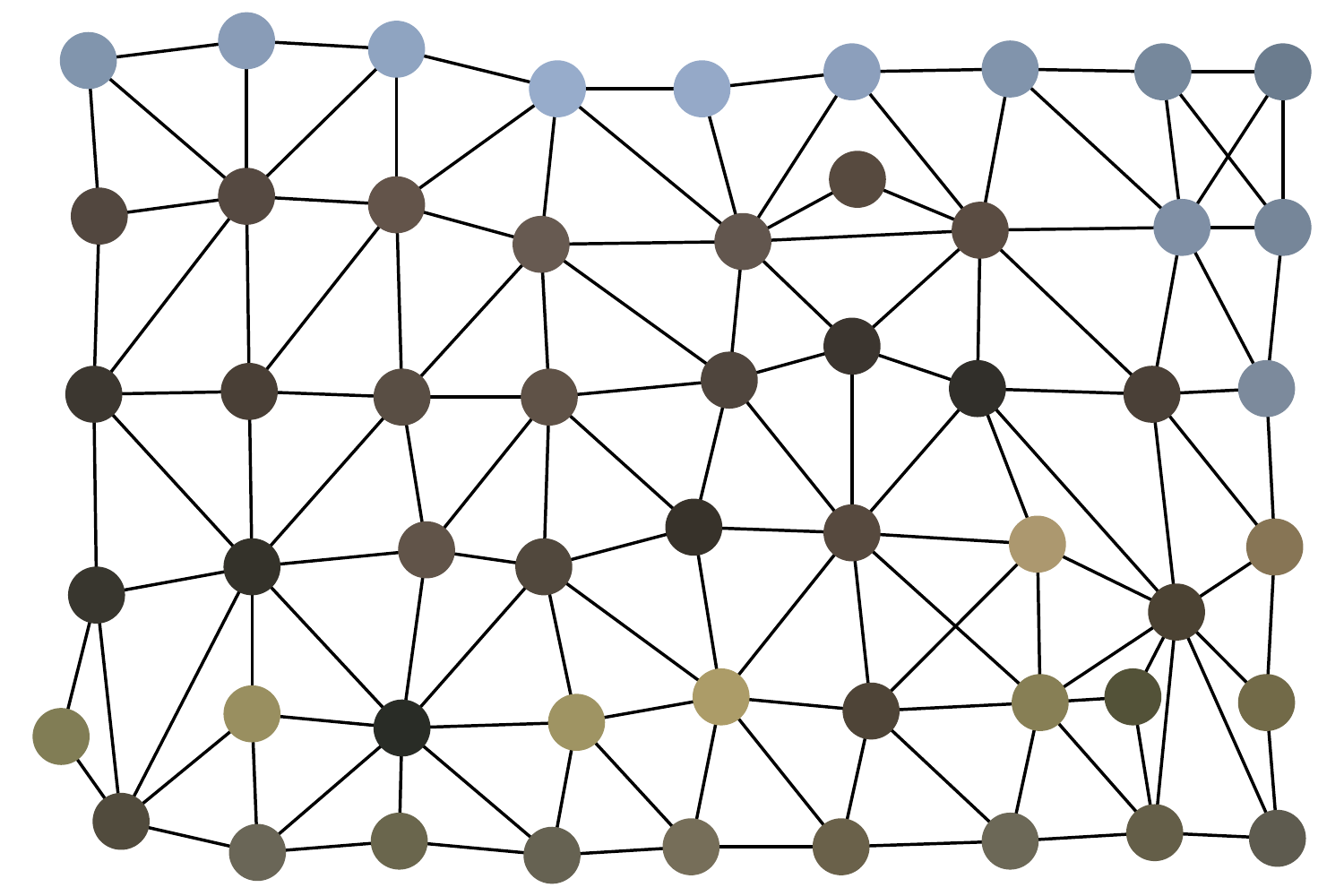}\\
        c) Superpixel image.&
        d) adjacency graph. 
 
\end{tabular}
    \caption{Expressing image segmentation by superpixelization as a multicut problem instance.}
    \label{fig:2-decomposition}
\end{figure}

\begin{figure}
    \centering
    \begin{tabular}{@{}c@{}c@{}}
        \includegraphics[width=0.49\linewidth]{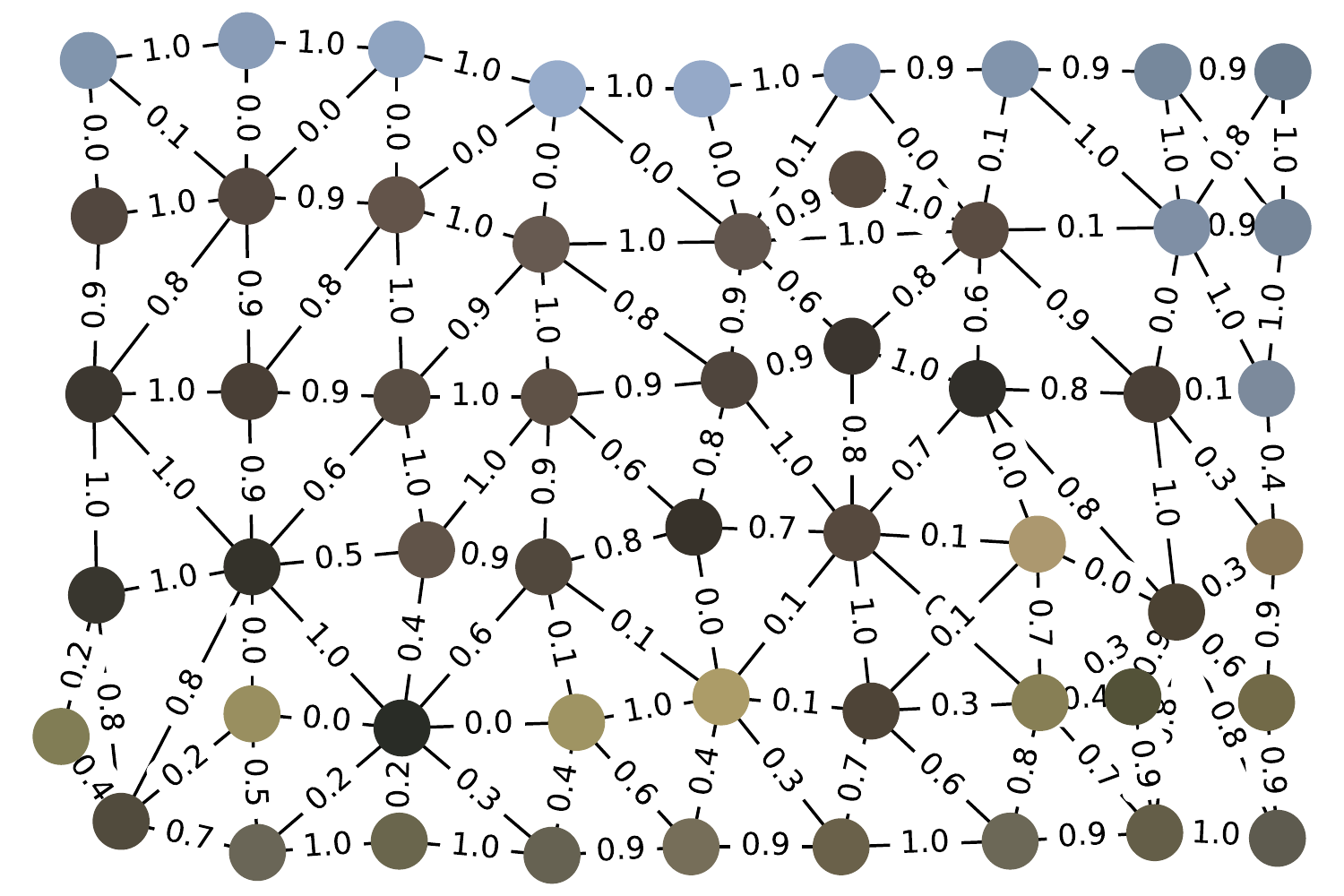}& \includegraphics[width=0.49\linewidth]{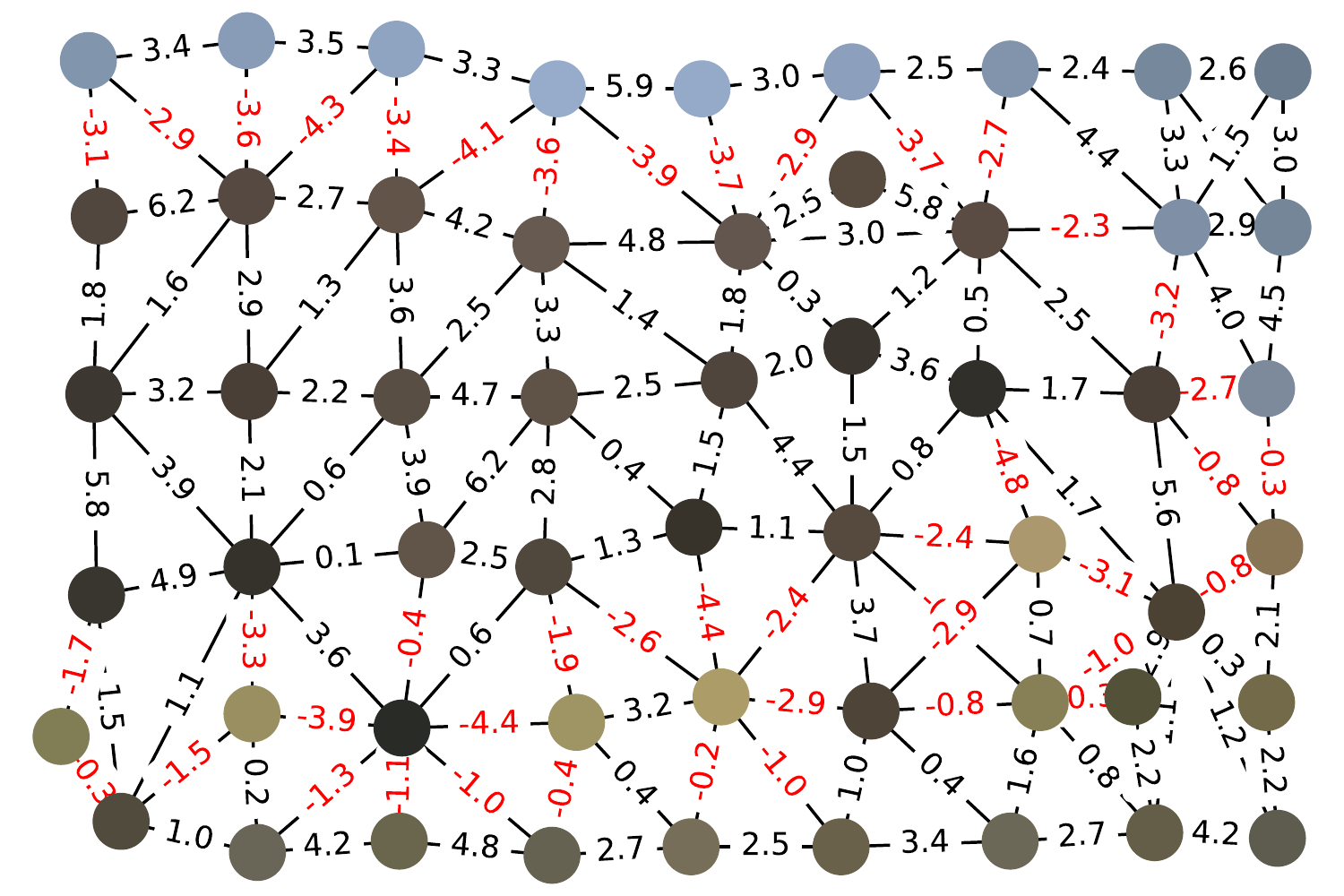} \\
        (a) Similarities& (b) real-valued weights\\
    \end{tabular}
    \caption{Example graph (a) Weights are similarities based on a Gaussian kernel. (b) Weights are log-odds of (a) and define a multicut problem instance.}
    \label{fig:a-elephants-graphs}
\end{figure}

For visualization purposes, we generate a small graph based on image segmentation of a training sample from the Berkeley Segmentation Dataset, BSDS300~\cite{BSDS300}, given in \autoref{fig:2-decomposition}(a). 
First, the gradients of the original image are computed using a Sobel filter.
Then, the watershed transformation is computed with $50$ desired markers, and a compactness of $0.0014$.
This results in the image consisting of $54$ segments as shown in
\autoref{fig:2-decomposition}(b).
Then the image is superpixelated computing the mean color of each resulting segment.
E.g., $c_i = \sum_{x\in S_i}x / |S_i|$ is the mean color of superpixel $i$, where $S_i$ is the set of pixels it contains.
$x$ is the color information of a pixel, yielding
\autoref{fig:2-decomposition}(c).
From these superpixels in region adjacency graph is constructed by connecting superpixels with a squared spatial distance less than $2$.
A positive weight $w_{ij}$ between two superpixels $i$ and $j$ is calculated using color similarity by applying a Gaussian kernel%
\footnote{\url{https://scikit-image.org/docs/dev/api/skimage.future.graph.html\#skimage.future.graph.rag\_mean\_color}}
with $\sigma=0.1$ on their color distance: $w_{ij}=e^{-|c_i-c_j| / \sigma}$.
The resulting graph is shown in \autoref{fig:a-elephants-graphs}.
From positive weights that represent superpixel similarity, positive and negative edge weights for the multicut problem can be derived using the logit function.

\begin{figure}[h]
    \centering
    \setlength\tabcolsep{0pt} 
    \begin{tabular}{cc}
       
        \includegraphics[width=0.48\linewidth]{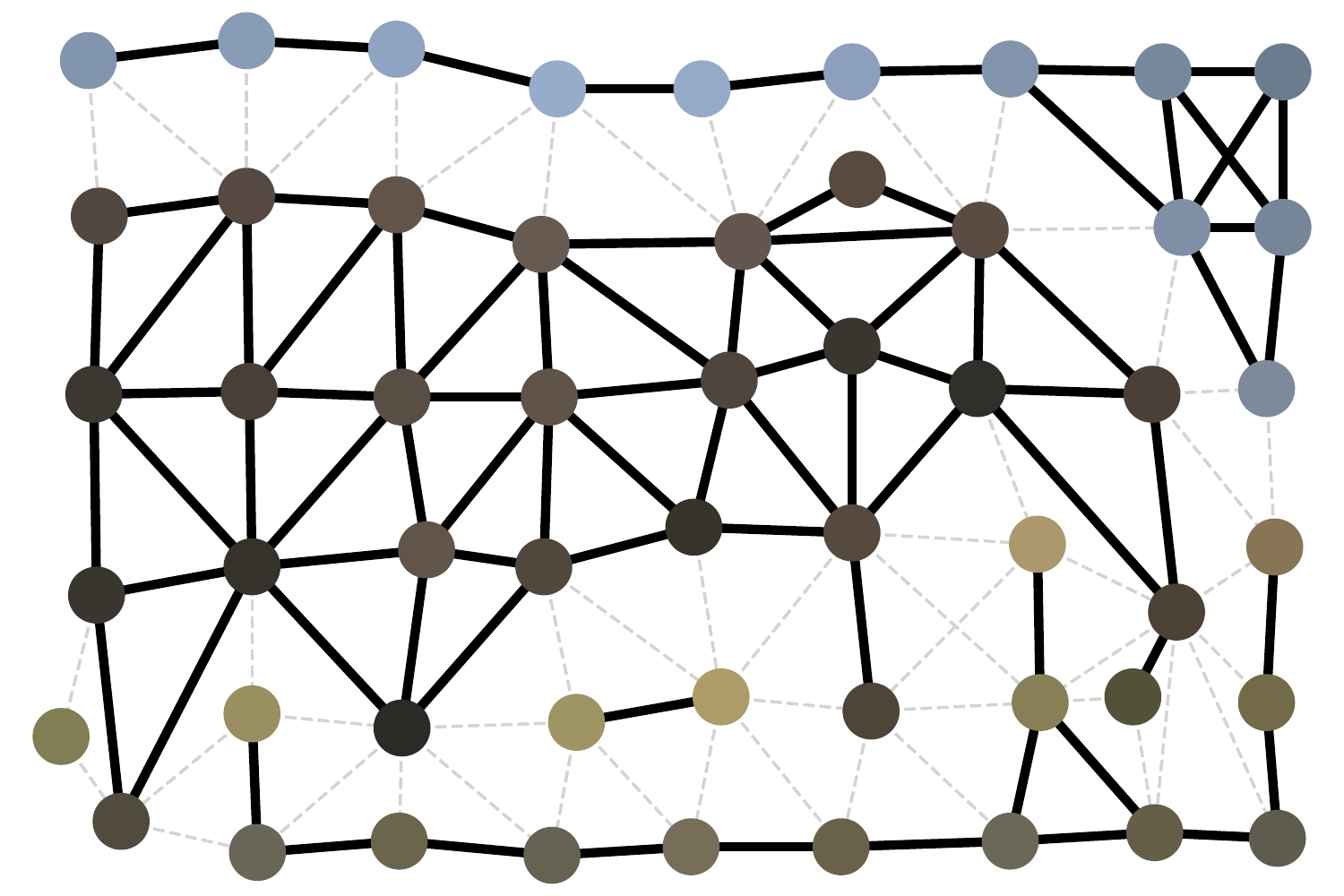} &
        \includegraphics[width=0.48\linewidth]{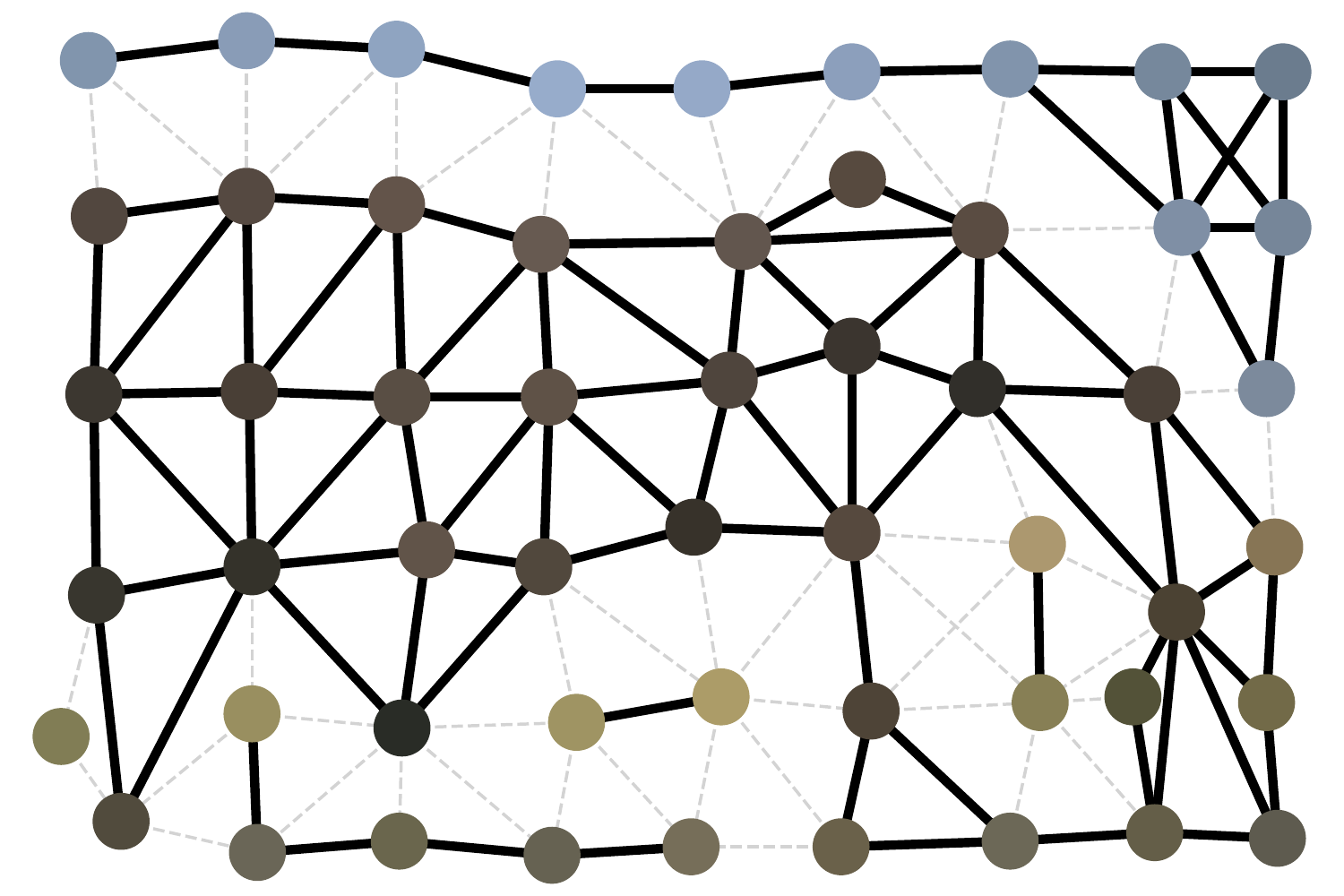} \\
        (a) ILP solution. &
        (b) GAEC solution. \\
    
        \includegraphics[width=0.4\linewidth]{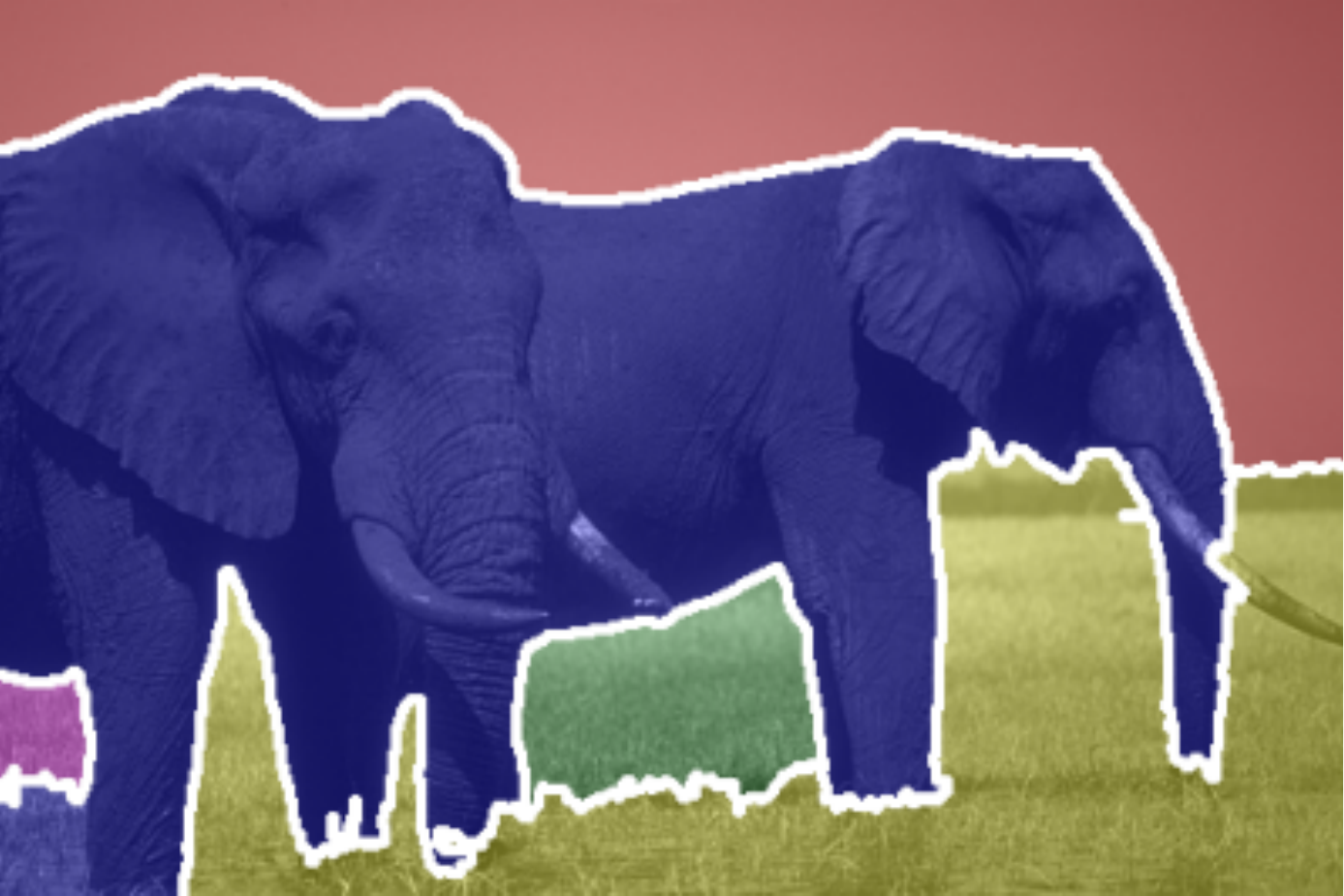} &
        \includegraphics[width=0.4\linewidth]{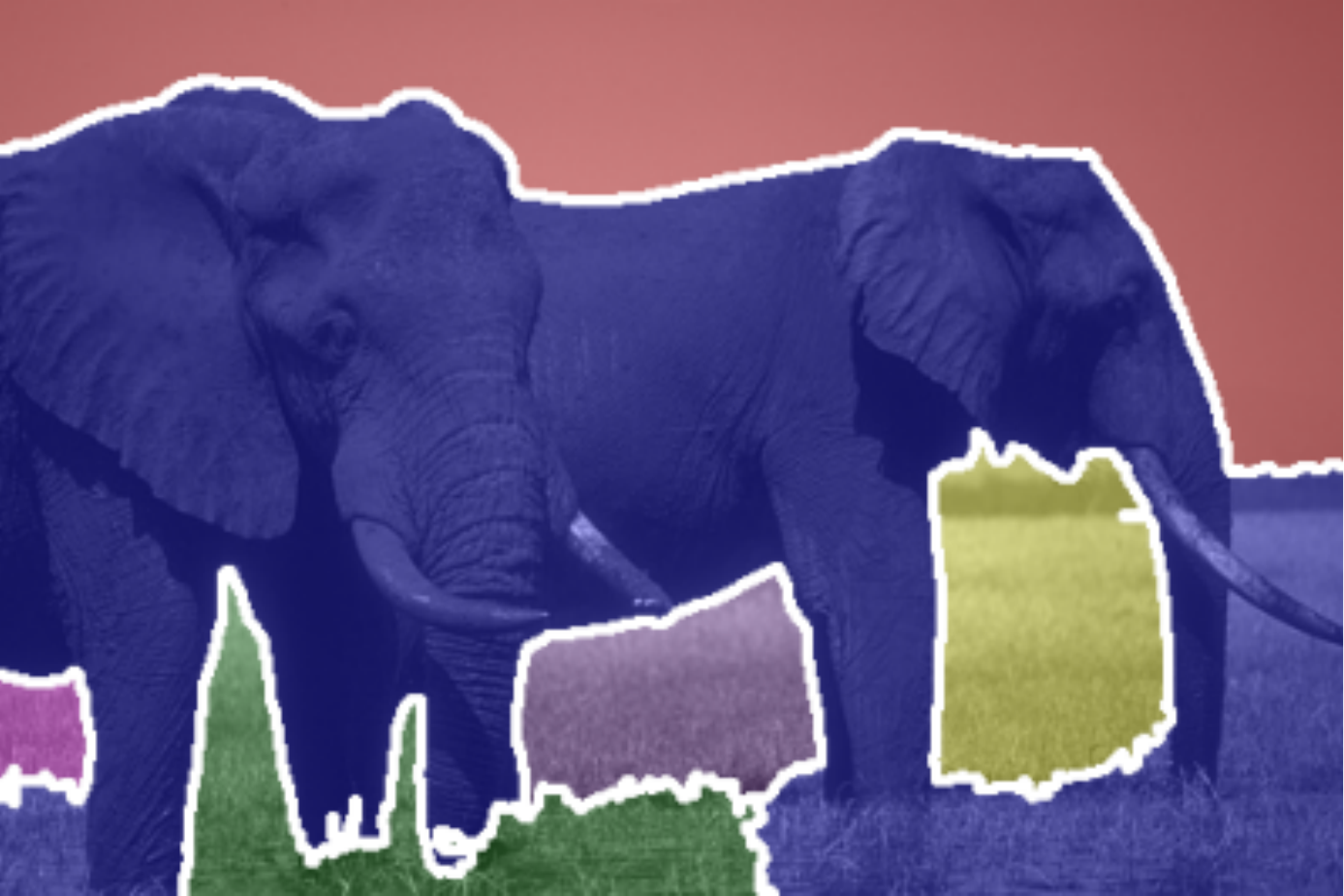} \\
        (c) ILP segmentation. &
        (d) GAEC segmentation. \\

    \end{tabular}
    \caption{
        Multicuts of the graph using different solvers.
        (a) shows the optimal cut, and (b) shows the cut computed with GAEC.
        Dotted, gray lines indicate that the edge is cut, and solid, black lines indicate otherwise.
        (c) and (d) depict the resulting segmentations.
    }
    \label{fig:multicut-solutions}
\end{figure}

Since the graph is small, we can compute its optimal solution using an ILP solver 
(see \autoref{fig:multicut-solutions}(a)).
The optimal objective value is denoted by $c$. 
The optimal solution yields an objective value of $c=-106.998$, while GAEC performs slightly worse with $c=-106.587$, thus providing optimality ratio $c_r=0.9961$.
\vfill
\begin{figure}[H]
    \centering
    \begin{tabular}{@{}r@{}l@{}}
        \includegraphics[width=0.49\linewidth]{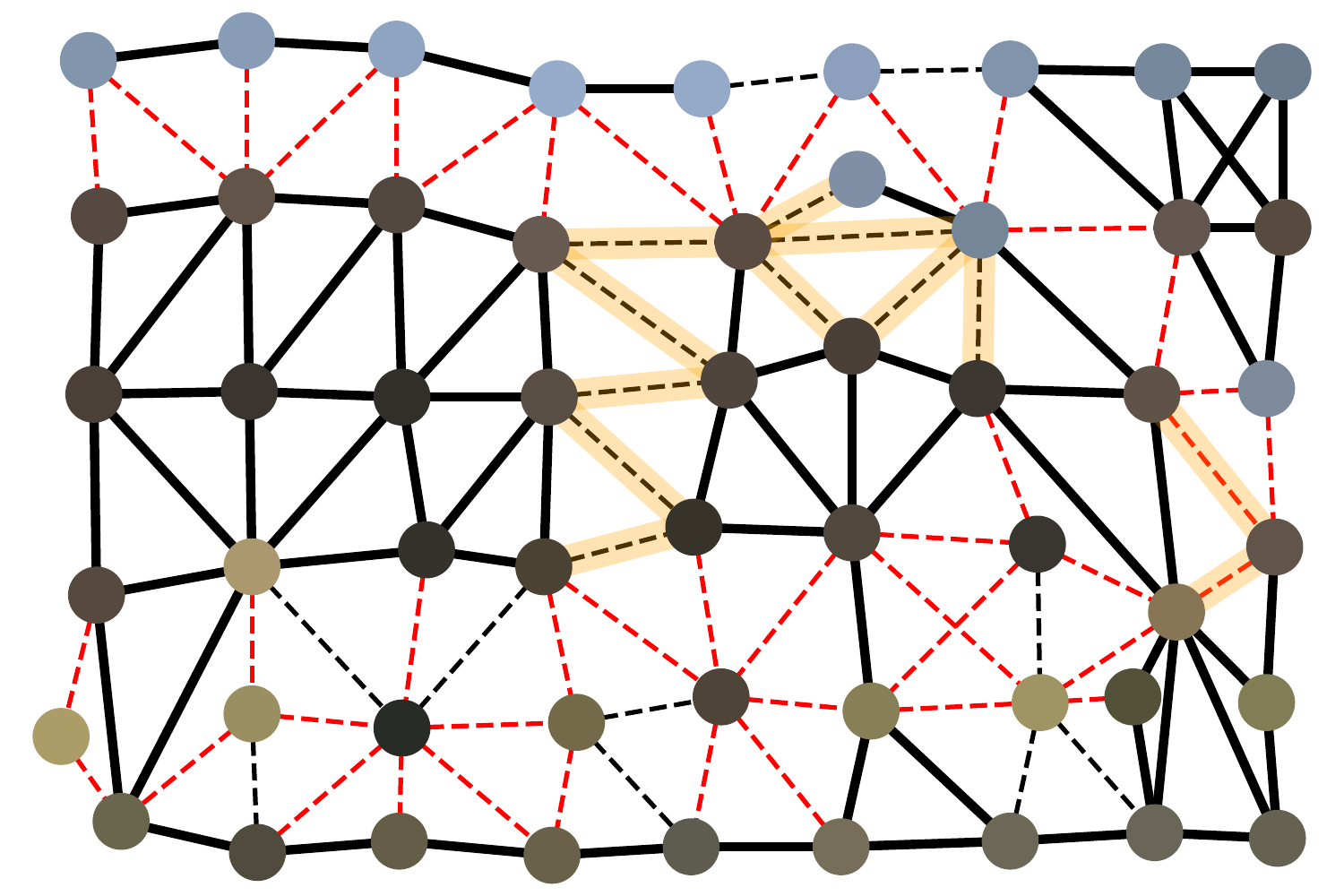} &
        \includegraphics[width=0.49\linewidth]{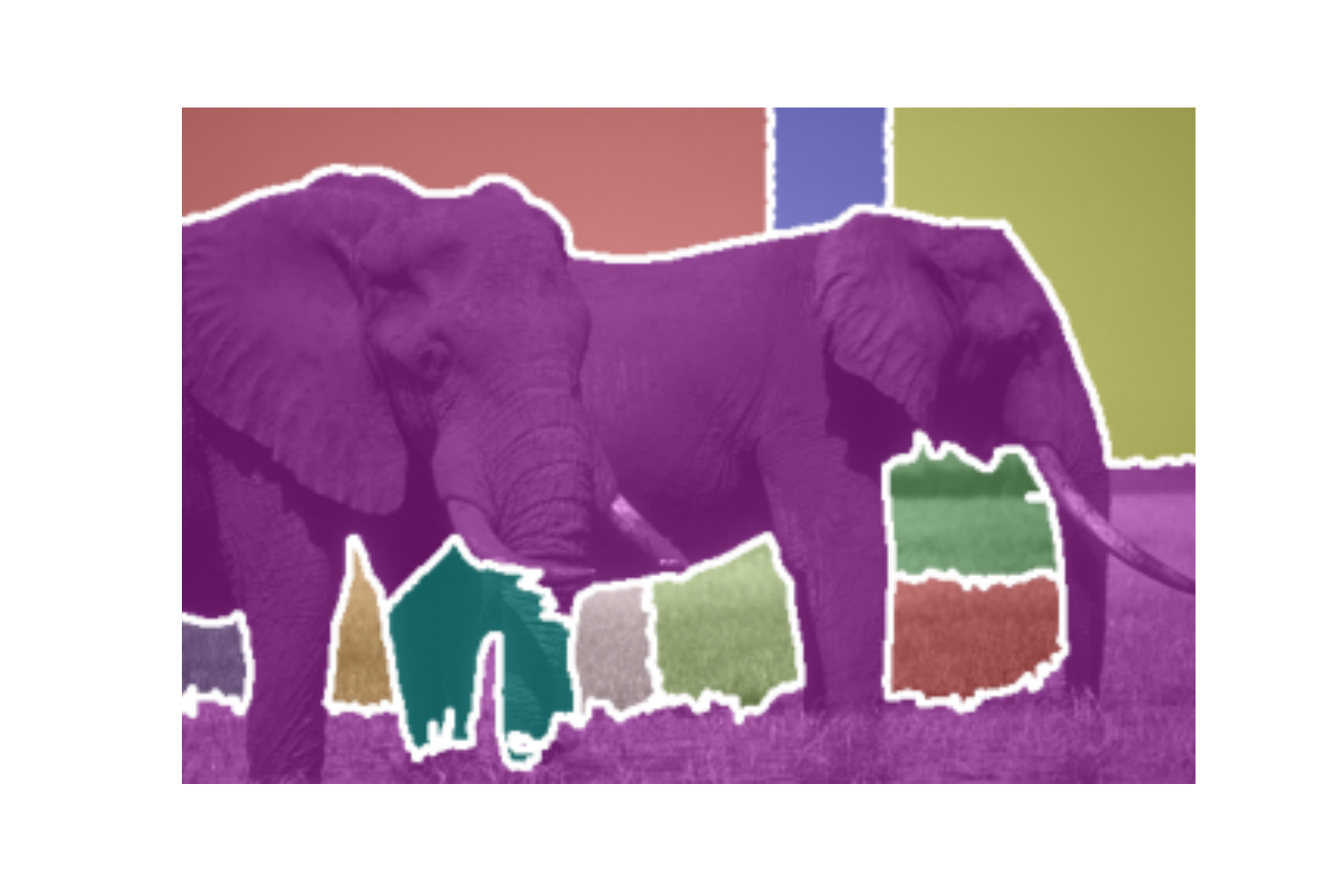} \\
        \multicolumn{2}{l}{(a) \texttt{GCN\_W\_BN}, Iris, $\text{depth}=12$,  $c_r=0.8948$.} \\
        \includegraphics[width=0.49\linewidth]{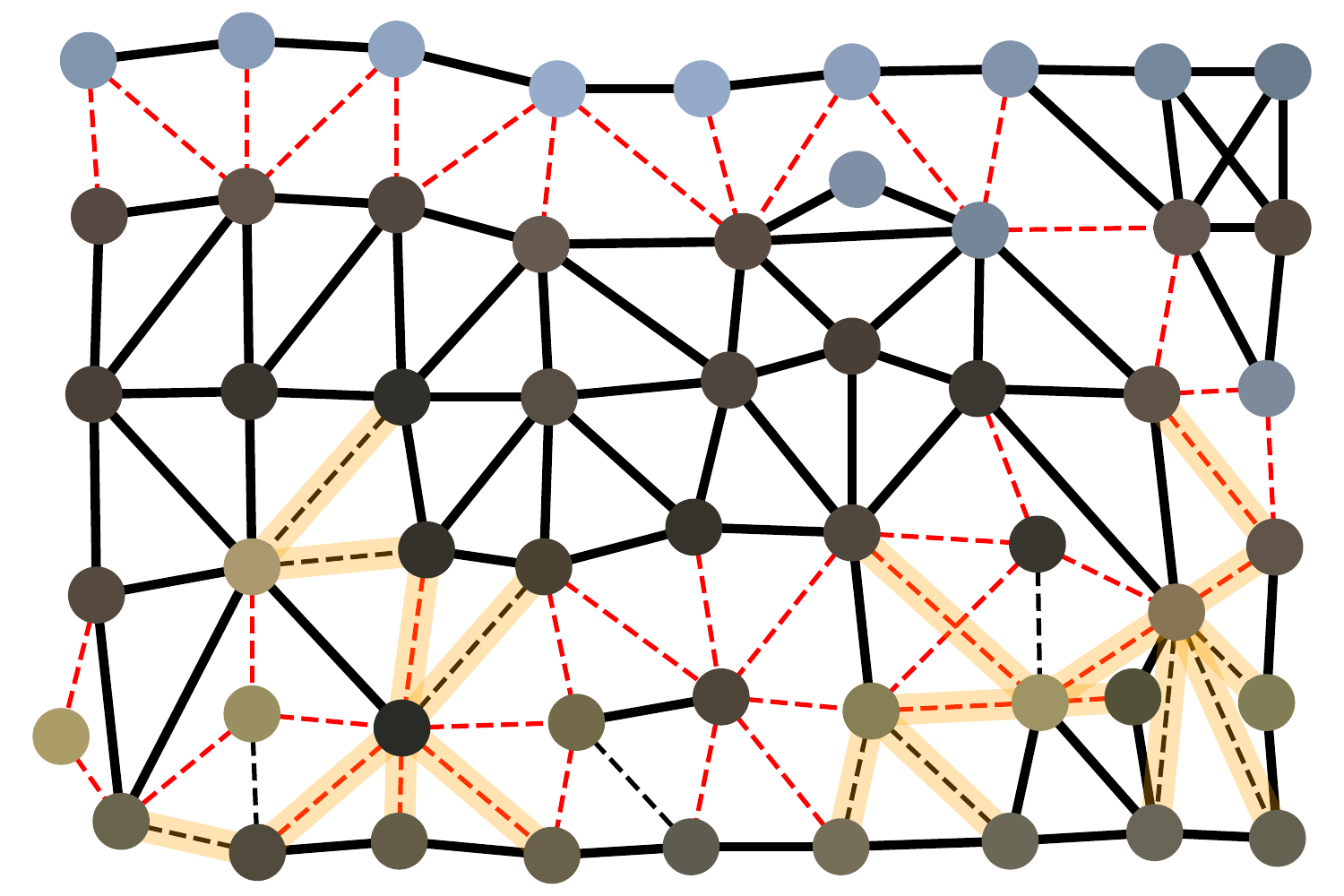} &
        \includegraphics[width=0.49\linewidth]{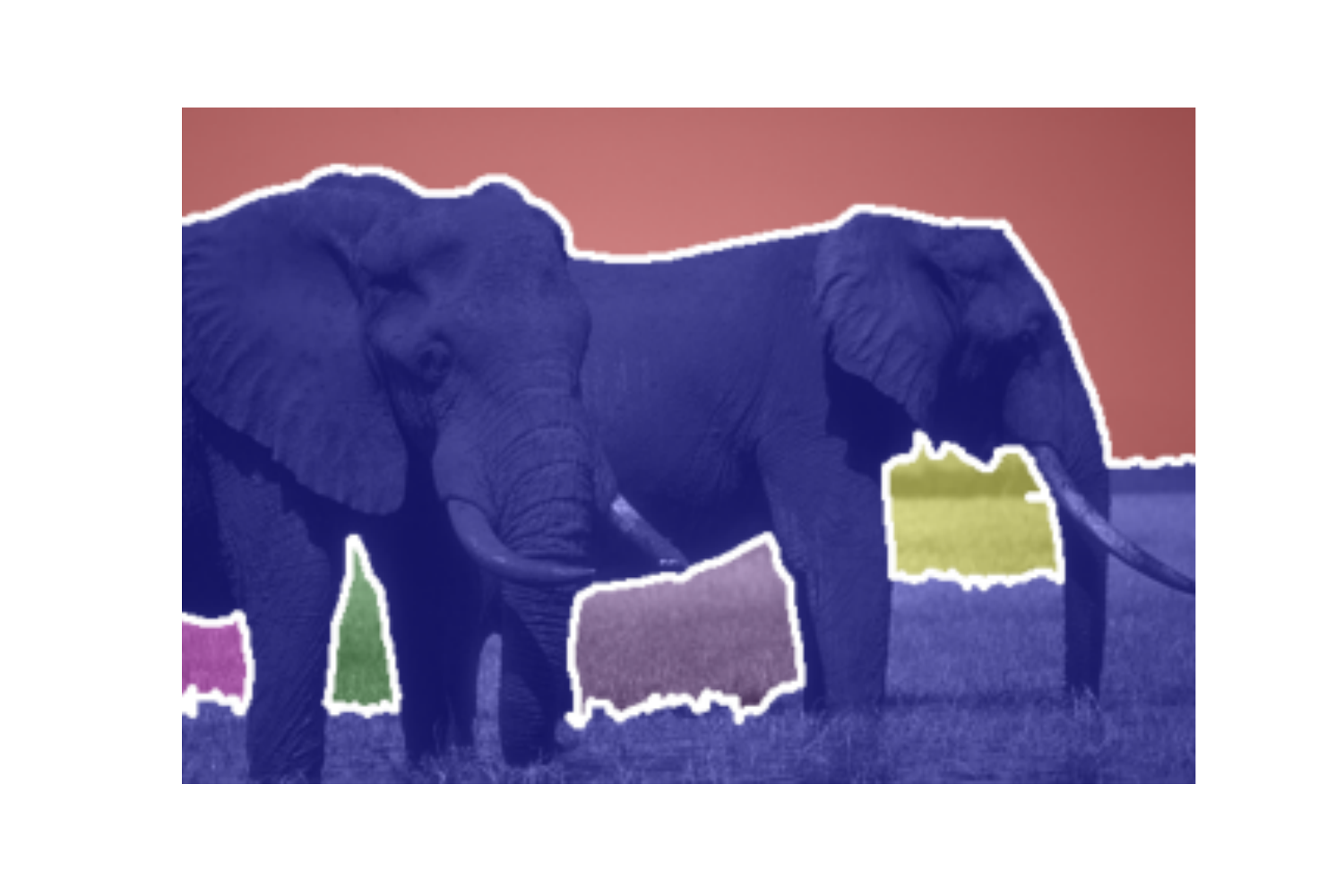} \\
        \multicolumn{2}{l}{(a) \texttt{GCN\_W\_BN}, Random, $\text{depth}=20$,  $c_r=0.9899$.} \\
    \end{tabular}
    \caption{
        Multicuts and resulting segmentations of the example graph by two of our trained models.
        Highlighted edges are removed by the model without partitioning the corresponding nodes.
        Hence, those cuts violate cycle consistencies and the edges are reinstated after rounding.
    }
    \label{fig:4-cuts}
\end{figure}
\noindent
For comparison, we show multicuts computed by a two of our models in
\autoref{fig:4-cuts}.
While the resulting optimality ratios are lower, the results are still of comparable practical quality, especially when considering the model trained on RandomMP.

\section{Training Datasets}
\subsection{IrisMP}
\label{sec:A-IrisMP}
The first dataset we generated, IrisMP, consists of multicut problem instances on complete graphs based on the Iris flower dataset \cite{1936Iris}.
This dataset is a well-known multivariate dataset containing $4$ different measurements, namely the width and length of sepal and petal for $3$ different species of flowers.
For each species the dataset contains $50$ samples.
For each graph we drew $2$ dimensions uniformly at random, and then uniformly drew $16$ to $24$ data points that we used as nodes.
We connected all the nodes and computed edge weights for each connection.
Edge weights are computed in three steps.
First, we computed the $L_2$ distance between two nodes.
Then we used a Gaussian kernel with $\sigma=0.6$ to convert the distances into similarity measures.
Since the resulting similarity is in $[0,1]$, we applied the logit function to retrieve positive as well as negative edge weights.
Since all graphs are fully connected it is sufficient for this dataset to only consider triangles to ensure cycle consistency. 
Since the number of edges increases quadratically with the number of nodes in such complete graphs,
we kept the number of drawn nodes small. IrisMP instances consist of $20$ nodes on average.
The resulting dataset contains $20~000$ instances for training.
Additionally, we generated $1000$ graphs each for validation and test splits.

\begin{table}[H]
    \caption{IrisMP statistics.}
    \label{tab:statistics-iris}
    \begin{center}
    \begin{scriptsize}
    \begin{tabular}{c|c|c|c}
    \toprule
    \multicolumn{1}{c}{}&\multicolumn{3}{c}{\textbf{Graph Stats}} \\
    \midrule
    Split & {Train} & {Eval} & {Test} \\
    \midrule
${|\mathcal{D}|}$&$20$~$000$&$1000$&$1000$\\
    \midrule
$\overline{|N|}$&$20 \pm 3$&$20 \pm 3$&$20 \pm 3$\\
$\overline{|E|}$&$194 \pm 50$&$196 \pm 49$&$192 \pm 50$\\
    \midrule
    \multicolumn{1}{c}{}&\multicolumn{3}{c}{\textbf{Weights}} \\
    \midrule
$\overline{w_{\text{min}}}$&\multicolumn{1}{r|}{$-4.51 \pm 0.29$}&\multicolumn{1}{r|}{$-4.51 \pm 0.33$}&\multicolumn{1}{r}{$-4.51 \pm 0.31$}\\
$\overline{w_{\text{avg}}}$&\multicolumn{1}{r|}{$-0.41 \pm 0.41$}&\multicolumn{1}{r|}{$-0.42 \pm 0.41$}&\multicolumn{1}{r}{$-0.40 \pm 0.42$}\\
$\overline{w_{\text{max}}}$&\multicolumn{1}{r|}{$4.57 \pm 0.12$}&\multicolumn{1}{r|}{$4.57 \pm 0.12$}&\multicolumn{1}{r}{$4.57 \pm 0.11$}\\
    \midrule
    \multicolumn{1}{c}{}&\multicolumn{3}{c}{\textbf{Objective Values}} \\
    \midrule
$c(\mathbf{y}^-)$&\multicolumn{1}{r|}{$-1.169$}&\multicolumn{1}{r|}{$-1.177$}&\multicolumn{1}{r}{$-1.168$}\\
$c(\mathbf{\tilde{y}})$&\multicolumn{1}{r|}{$-1.103$}&\multicolumn{1}{r|}{$-1.111$}&\multicolumn{1}{r}{$-1.100$}\\
$c(\mathbf{1})$&\multicolumn{1}{r|}{$-0.407$}&\multicolumn{1}{r|}{$-0.426$}&\multicolumn{1}{r}{$-0.405$}\\
$c(\mathbf{y}^+)$&\multicolumn{1}{r|}{$0.762$}&\multicolumn{1}{r|}{$0.751$}&\multicolumn{1}{r}{$0.762$}\\
    \bottomrule
    \end{tabular}
    \end{scriptsize}
    \end{center}
\end{table}

\subsection{RandomMP}
\label{sec:A-RandomMP}
To complement the IrisMP dataset, we generated a second dataset that contains sparse but larger problem instances in terms of the number of nodes, called RandomMP.
To generate this dataset, we employed the following procedure. 
First, we sampled the number of nodes from a normal distribution with $\mu=180$ and $\sigma=30$.
Then, for each node, we sampled its coordinates on a 2D plane uniformly in $[0,1]$ for each coordinate.
We connected nodes based on the $k$ nearest neighbors, where $k$ is drawn from a normal distribution with $\mu=6$ and $\sigma=2$.
However, we constrained the minimum number of neighbors of each node to $1$ so that the graph is connected.
We computed edge weights based on the $L_2$ distance on the plane.
Then, we subtracted the median from all edge weights, to achieve an approximate distribution of $50\%$ positive as well as $50\%$ negative connections.
Again, we generated a training dataset of size $20~000$ 
and splits of $1000$ for validation and testing.

\begin{table}[h]
    \caption{RandomMP statistics.}
    \label{tab:statistics-random}
    \begin{center}
    \begin{small}
    \begin{tabular}{c|c|c|c}
    \toprule
    \multicolumn{1}{c}{}&\multicolumn{3}{c}{\textbf{Graph Stats}} \\
    \midrule
    Split & {Train} & {Eval} & {Test} \\
    \midrule
    ${|\mathcal{D}|}$&$20$~$000$&$1000$&$1000$\\
    \midrule
$\overline{N}$&$180 \pm 30$&$180 \pm 29$&$179 \pm 31$\\
$\overline{E}$&$686 \pm 115$&$685 \pm 114$&$684 \pm 118$\\
    \midrule
    \multicolumn{1}{c}{}&\multicolumn{3}{c}{\textbf{Weights}} \\
    \midrule
    $\overline{w_{\text{min}}}$&\multicolumn{1}{r|}{$-9.36 \pm 1.09$}&\multicolumn{1}{r|}{$-9.37 \pm 1.08$}&\multicolumn{1}{r}{$-9.39 \pm 1.12$}\\
    $\overline{w_{\text{avg}}}$&\multicolumn{1}{r|}{$-0.00 \pm 0.10$}&\multicolumn{1}{r|}{$0.00 \pm 0.10$}&\multicolumn{1}{r}{$0.00 \pm 0.10$}\\
    $\overline{w_{\text{max}}}$&\multicolumn{1}{r|}{$9.38 \pm 1.09$}&\multicolumn{1}{r|}{$9.37 \pm 1.10$}&\multicolumn{1}{r}{$9.38 \pm 1.10$}\\
    \midrule
    \multicolumn{1}{c}{}&\multicolumn{3}{c}{\textbf{Objective Values}} \\
    \midrule
$c(\mathbf{y}^-)$&\multicolumn{1}{r|}{$-1.196$}&\multicolumn{1}{r|}{$-1.193$}&\multicolumn{1}{r}{$-1.195$}\\
$c(\mathbf{\tilde{y}})$&\multicolumn{1}{r|}{$-0.844$}&\multicolumn{1}{r|}{$-0.842$}&\multicolumn{1}{r}{$-0.841$}\\
$c(\mathbf{1})$&\multicolumn{1}{r|}{$-0.001$}&\multicolumn{1}{r|}{$0.002$}&\multicolumn{1}{r}{$0.002$}\\
$c(\mathbf{y}^+)$&\multicolumn{1}{r|}{$1.196$}&\multicolumn{1}{r|}{$1.195$}&\multicolumn{1}{r}{$1.197$}\\
    \bottomrule
    \end{tabular}
    \end{small}
    \end{center}
\end{table}
\newpage

\section{Test Datasets}
\begin{centering}
\begin{minipage}{.5\linewidth}
    \begin{table}[H]
        \caption{BSDS300 statistics.}
        \label{tab:statistics-seg2d}
        \begin{center}
        \begin{small}
        \begin{tabular}{c|c}
        \toprule
        \multicolumn{2}{c}{\textbf{Graph Stats}} \\
        \midrule
        ${|\mathcal{D}|}$&$100$\\
        \midrule
    $\overline{N}$&$1551 \pm 777$\\
    $\overline{E}$&$4432 \pm 2$~$269$\\
        \midrule
        \multicolumn{2}{c}{\textbf{Weights}} \\
        \midrule
    $\overline{w_{\text{min}}}$&\multicolumn{1}{r}{$-10.77 \pm 3.74$}\\
    $\overline{w_{\text{avg}}}$&\multicolumn{1}{r}{$0.48 \pm 0.57$}\\
    $\overline{w_{\text{max}}}$&\multicolumn{1}{r}{$10.27 \pm 2.32$}\\
        \midrule
        \multicolumn{2}{c}{\textbf{Objective Values}} \\
        \midrule
    $c(\mathbf{y}^-)$&\multicolumn{1}{r}{$-0.738$}\\
    $c(\mathbf{\tilde{y}})$&\multicolumn{1}{r}{$-0.669$}\\
    $c(\mathbf{1})$&\multicolumn{1}{r}{$0.497$}\\
    $c(\mathbf{y}^+)$&\multicolumn{1}{r}{$1.235$}\\
        \bottomrule
        \end{tabular}
        \end{small}
        \end{center}
    \end{table}
\end{minipage}
\begin{minipage}{.5\linewidth}
    \begin{table}[H]
        \caption{CREMI statistics.}
        \label{tab:statistics-cremi}
        \begin{center}
        \begin{small}
        \begin{tabular}{c|c}
        \toprule
        \multicolumn{2}{c}{\textbf{Graph Stats}} \\
        \midrule
    ${|\mathcal{D}|}$&$3$\\
        \midrule
    $\overline{N}$&$31$~$988 \pm 4769$ \\
    $\overline{E}$&$212$~$686 \pm 24$~$767$ \\
        \midrule
        \multicolumn{2}{c}{\textbf{Weights}} \\
        \midrule
    $\overline{w_{\text{min}}}$&\multicolumn{1}{r}{$-6.91 \pm 0.00$}\\
    $\overline{w_{\text{avg}}}$&\multicolumn{1}{r}{$-0.15 \pm 0.26$}\\
    $\overline{w_{\text{max}}}$&\multicolumn{1}{r}{$79.52 \pm 12.63$}\\
        \midrule
        \multicolumn{2}{c}{\textbf{Objective Values}} \\
        \midrule
    $c(\mathbf{y}^-)$&\multicolumn{1}{r}{$-1.013$}\\
    $c(\mathbf{\tilde{y}})$&\multicolumn{1}{r}{$-1.002$}\\
    $c(\mathbf{1})$&\multicolumn{1}{r}{$-0.135$}\\
    $c(\mathbf{y}^+)$&\multicolumn{1}{r}{$0.878$}\\
        \bottomrule
        \end{tabular}
        \end{small}
        \end{center}
    \end{table}
\end{minipage}
\end{centering}



\newpage
\section{Modified Update Functions}

In the main paper, we extend the Graph Convolutional Network (GCN) \cite{2017GCN} such that it is able to process weighted graphs.
In addition to that, we extend Signed Graph Convolutional Network (SGCN) \cite{2018Signed} and Graph Isomorphic Network (GIN) \cite{2019GIN} in a similar way to be able to produce the results in Table~1 of the main paper.
\autoref{tab:adjustments} shows our modifications to the update functions of those networks.

\label{sec:A-update-functions}
\begin{table*}[h]
    \caption{
        Modifications highlighted in blue to MPNNs to account for signed edge weights.
    }
    \label{tab:adjustments}
    \centering
        \begin{small}
            \begin{tabular}{c|c}
                \toprule
                \textbf{Network} & \textbf{Modified Update Function} \\
                \midrule
                    \textbf{GCN} &
                    $\mathbf{h}_\mathbf{u}^{(t)} = g^{(t)}_\theta \left(
                    \mathbf{h}_\mathbf{u}^{(t-1)} +
                    \sum_{v\in\mathcal{N}(u)} \textcolor{blue}{w_{v,u}} \cdot \textcolor{blue}{\left({\overline{\text{deg}}(u)\overline{\text{deg}}(v)}\right)^{-1/2}}
                    \right)$ \\
                \midrule
                    \multirow{9}{*}{\textbf{SGCN}} &
                    \multicolumn{1}{l}{
                        $\mathbf{h}_\mathbf{u}^{B(t)} = g^{B(t)}_\theta \Bigg(
                        \mathbf{h}_\mathbf{u}^{B(t-1)} \;+ $
                    } \\
                    & \multicolumn{1}{r}{
                        $\sum_{v\in\mathcal{N}^+(u)}
                        \frac{\textcolor{blue}{w_{v,u}} \cdot \mathbf{h}_\mathbf{v}^{B(t-1)}}{|\mathcal{N}^+(u)|} +
                        \sum_{v\in\mathcal{N}^-(u)}
                        \frac{\textcolor{blue}{w_{v,u}} \cdot \mathbf{h}_\mathbf{v}^{U(t-1)}}{|\mathcal{N}^-(u)|}
                        \Bigg)$
                    } \\
                    \cmidrule{2-2}
                    & \multicolumn{1}{l}{
                        $\mathbf{h}_\mathbf{u}^{U(t)} = g^{U(t)}_\theta \Bigg(
                        \mathbf{h}_\mathbf{u}^{U(t-1)} \;+ $
                    } \\
                    & \multicolumn{1}{r}{
                        $\sum_{v\in\mathcal{N}^+(u)}
                        \frac{\textcolor{blue}{w_{v,u}} \cdot \mathbf{h}_\mathbf{v}^{U(t-1)}}{|\mathcal{N}^+(u)|} +
                        \sum_{v\in\mathcal{N}^-(u)}
                        \frac{\textcolor{blue}{w_{v,u}} \cdot \mathbf{h}_\mathbf{v}^{B(t-1)}}{|\mathcal{N}^-(u)|}
                        \Bigg)$
                    } \\
                \midrule
                    \textbf{GIN} &
                    $\mathbf{h}_\mathbf{u}^{(t)} = g^{(t)}_\theta \left(
                    \left( 1 + \epsilon^{(t)} \right) \cdot \mathbf{h}_\mathbf{u}^{(t-1)}
                    +
                    \sum_{v\in\mathcal{N}(u)} \textcolor{blue}{w_{v,u}} \cdot \mathbf{h}_\mathbf{v}^{(t-1)}
                    \right)$ \\
                \bottomrule
            \end{tabular}
        \end{small}
\end{table*}

\begin{figure}[H]
    \centering
    \setlength\tabcolsep{0pt} 
    \begin{tabular}{ccc}
        \includegraphics[height=3.5cm]{images/experiments/gcn_train.pdf} &
        \includegraphics[height=3.5cm]{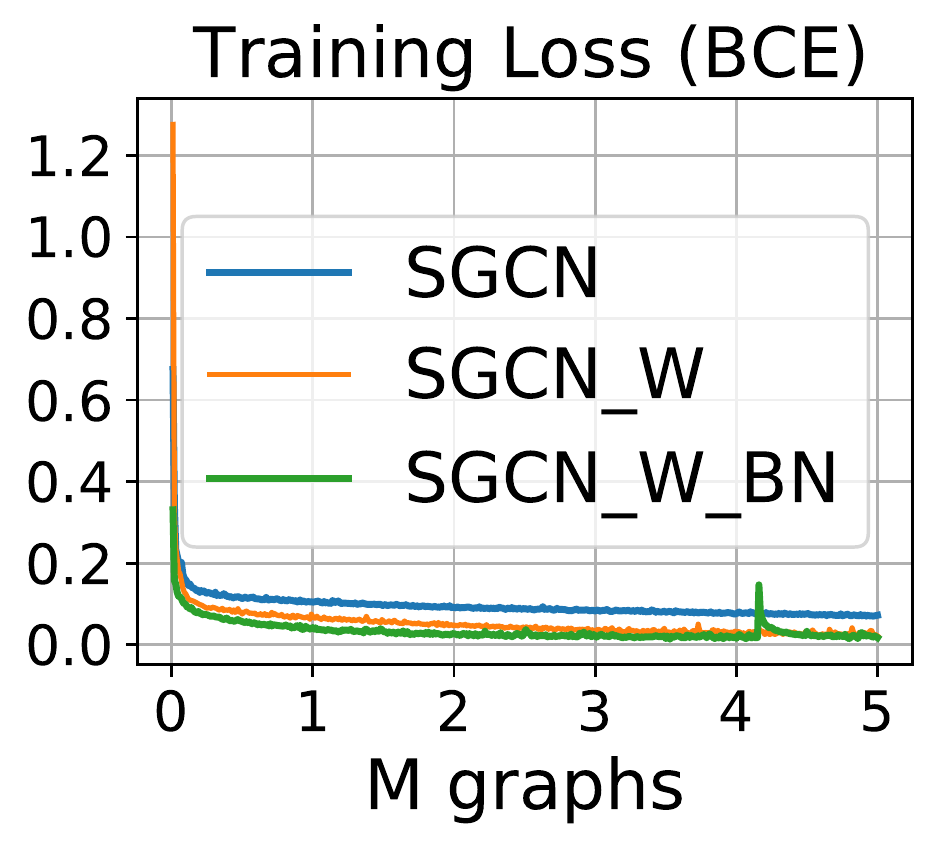} &
        \includegraphics[height=3.5cm]{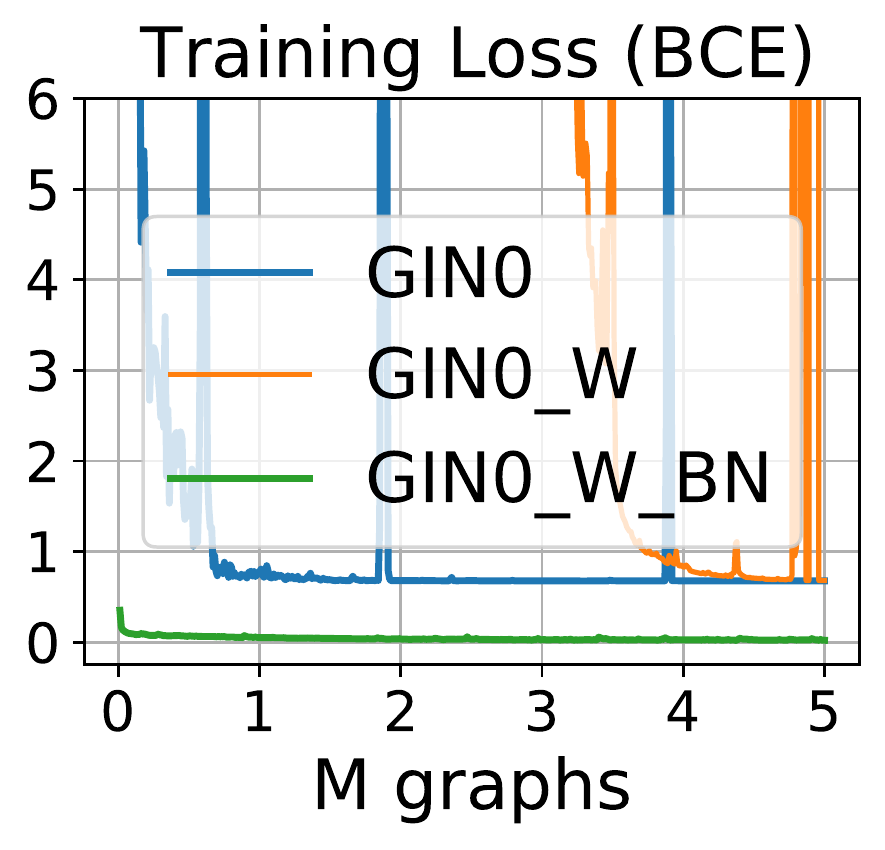} \\
        (a) GCN Training Loss. &
        (b) SGCN Training Loss. &
        (c) GIN Training Loss. \\
        \includegraphics[height=3.5cm]{images/experiments/gcn_eval.pdf} &
        \includegraphics[height=3.5cm]{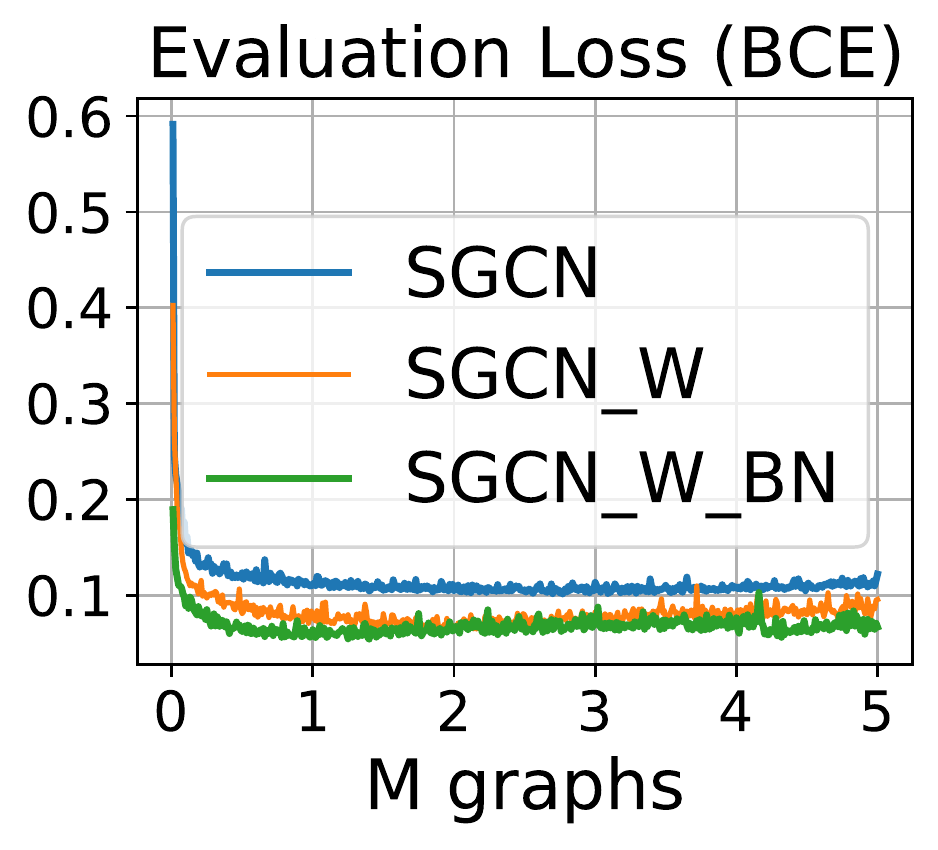} &
        \includegraphics[height=3.5cm]{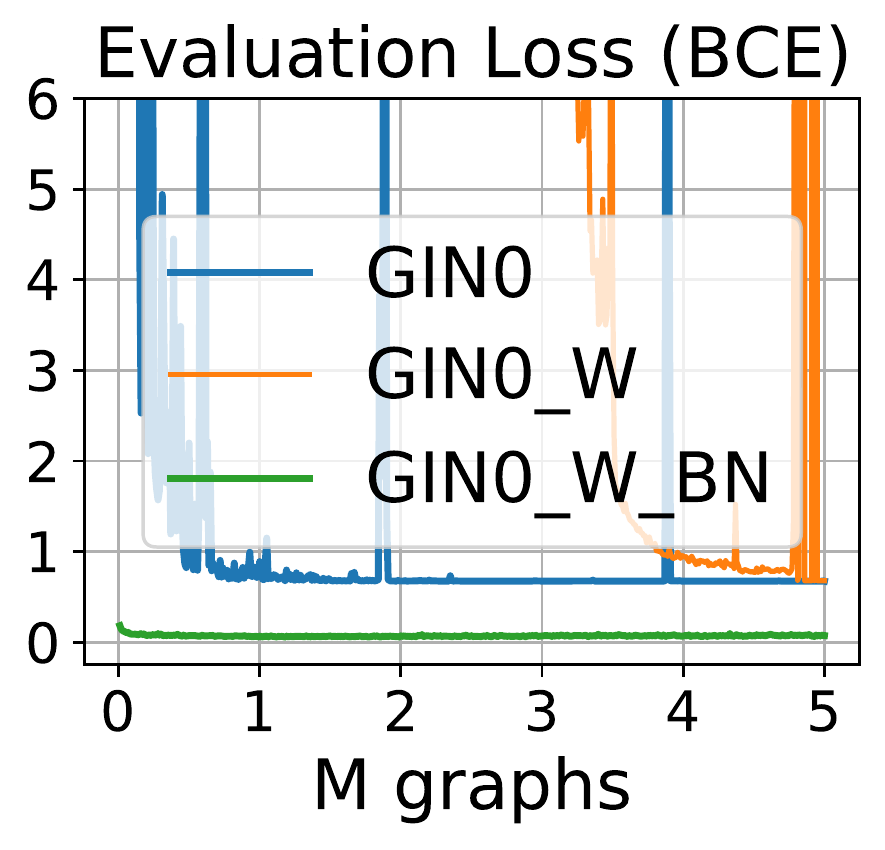} \\
        (d) GCN Evaluation Loss. &
        (e) SGCN Evaluation Loss. &
        (f) GIN Evaluation Loss.
    \end{tabular}
    \caption{
        Comparison of the training (a)-(c) and evaluation (d)-(f) losses of the networks with (indicated by \emph{\_W}) and without modified update functions, as well as with batch normalization (indicated by \emph{\_W\_BN}).
    }
    \label{fig:appendix-update-adjustments}
\end{figure}

In \autoref{fig:appendix-update-adjustments} we show the training and evaluation losses of the modified update functions in comparison to their vanilla versions.
We trained on IrisMP with $12$ message-passing iterations, set the dimensionality of node embeddings to $128$, and performed edge classification with an MLP that consists of $2$ hiddens layers with $256$ neurons each.
No CCL was applied.
Optimization was performed with Adam~\cite{2015Adam} ($0.001$ learning rate, $5\cdot10^{-4}$ weight decay, $(0.9, 0.999)$ betas) and a batch size of $200$.

\newpage
\begin{figure}[H]
    \centering
    \setlength\tabcolsep{0pt} 

        \begin{tabular}{c}
        \includegraphics[height=3.5cm]{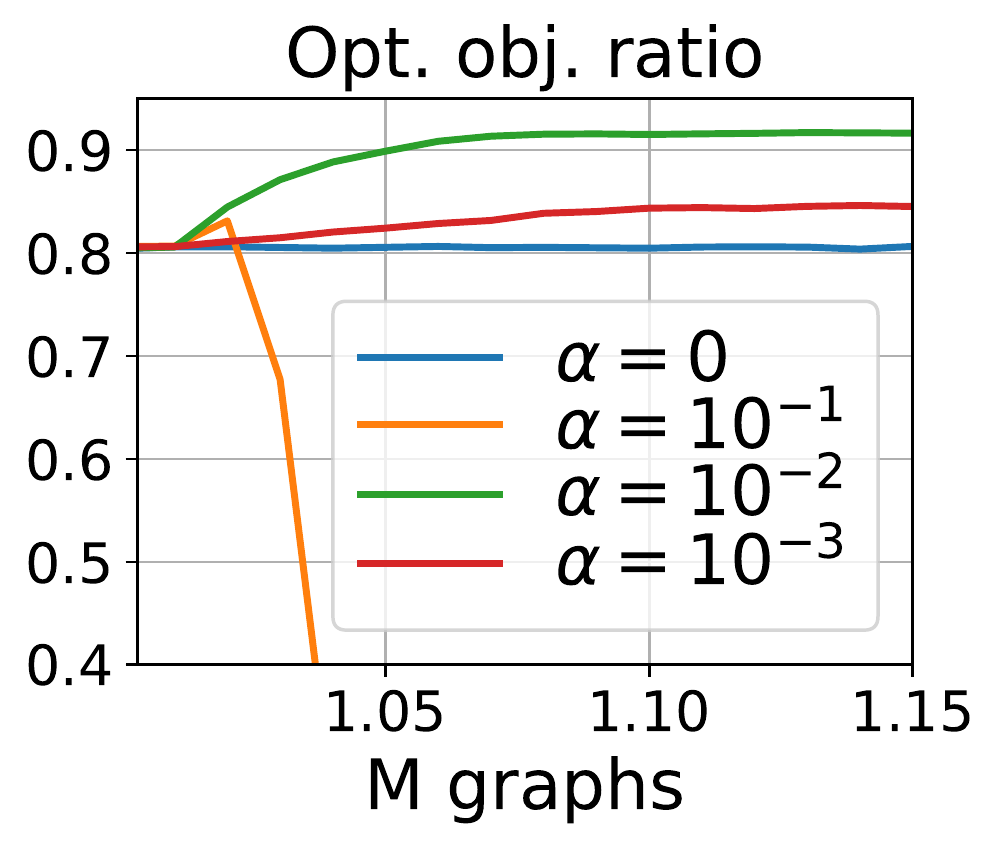} \\
         Hyperparameter $\alpha$.
    \end{tabular}
    \caption{
Training progress of the optimal objective ratio for different values of $\alpha$ when training GCN\_W\_BN on RandomMP and applying CCL after $1M$ instances.
    }
    \label{fig:4-training-iris-eta-gcn2}
\end{figure}
\newpage
\section{Finetuning Experiments}
Originally, generating synthetic training data for our method was mainly driven by the lack of suitable training data available. The benchmark datasets mentioned consist of only a few problem instances (BSDS300: 100, CREMI: 3, Knott: 24). Although BSDS300 consists of 100 training and 100 testing images, the multicut problem instances provided by the OpenGM benchmark are solely based on the test images. This is the case, because training images were used to train a model that derives edge weights for the testing images and are discarded afterwards. Nevertheless, we consider it interesting to see how the model behaves in this environment of scarce training data, and whether finetuning can help to boost model performance on a specific task. 
Therefore, we ran the following additional experiments: 
\begin{enumerate}[a)]
\item we trained GCN\_W\_BN from scratch on training splits of these datasets and 
\item we finetuned the best performing model of Table 1 (GCN\_W\_BN trained on RandomMP - referred to as RMP-GCN in the following) on training splits of these datasets. 
\end{enumerate}

We split BSDS300 into 70/20/10 (train/eval/test), CREMI into 2/1 (train/eval) and Knott into 18/6 (train/eval). The performance after training is evaluated on the eval split and compared to the performance as optimality ratio of the RandomMP trained GCN\_W\_BN (RMP) on the eval split. Results are given in 
\autoref{tab:my_labelapp}.
Please note that these results can not be directly compared to Table 1 since the whole datasets were evaluated in Table 1.
\begin{table}[ht]
    \centering
    \begin{tabular}{c|c cc}\toprule
    &RMP&from scratch&RMP + finetuned\\ \midrule
         BSDS300 & $0.8818$ & $0.8834$ & $0.8818$ \\
         Knott   & $0.8386$ & $0.8335$ & $0.9249$ \\
         CREMI   & $0.9068$ & $0.8081$ & $0.9365$ \\
         \bottomrule
    \end{tabular}
    \caption{Domain specific training of GCN\_W\_BN from scratch and finetuned on the scarce available training data for each task, compared to the general purpose model trained on RandomMP (RMP). Results show that domain specific properties can be learned, from few samples but pre-training can help in general.}
    \label{tab:my_labelapp}
\end{table}

 Still, the results indicate that when trained on BSDS300 from scratch, the model improves on the eval split from 0.8818 to 0.8834. We were not able to find models that outperform RMP-GCN trained on Knott and CREMI. For finetuning we tried three different settings: i) retraining all parameters (GCN+edge classifier), ii) only retraining edge classifier parameters and iii) only retraining the last layer of the edge classifier. We found that setting iii) worked best overall. As shown in
 \autoref{tab:my_labelapp},
 we found models that improved on Knott and CREMI. Yet, finetuning did not help to improve performance on BSDS300.


\section{Embedding Space Visualizations}
\label{sec:A-vis}

\begin{figure}[H]
    \centering
    \setlength\tabcolsep{0pt} 
    \begin{tabular}{cc}
        \includegraphics[height=3.5cm]{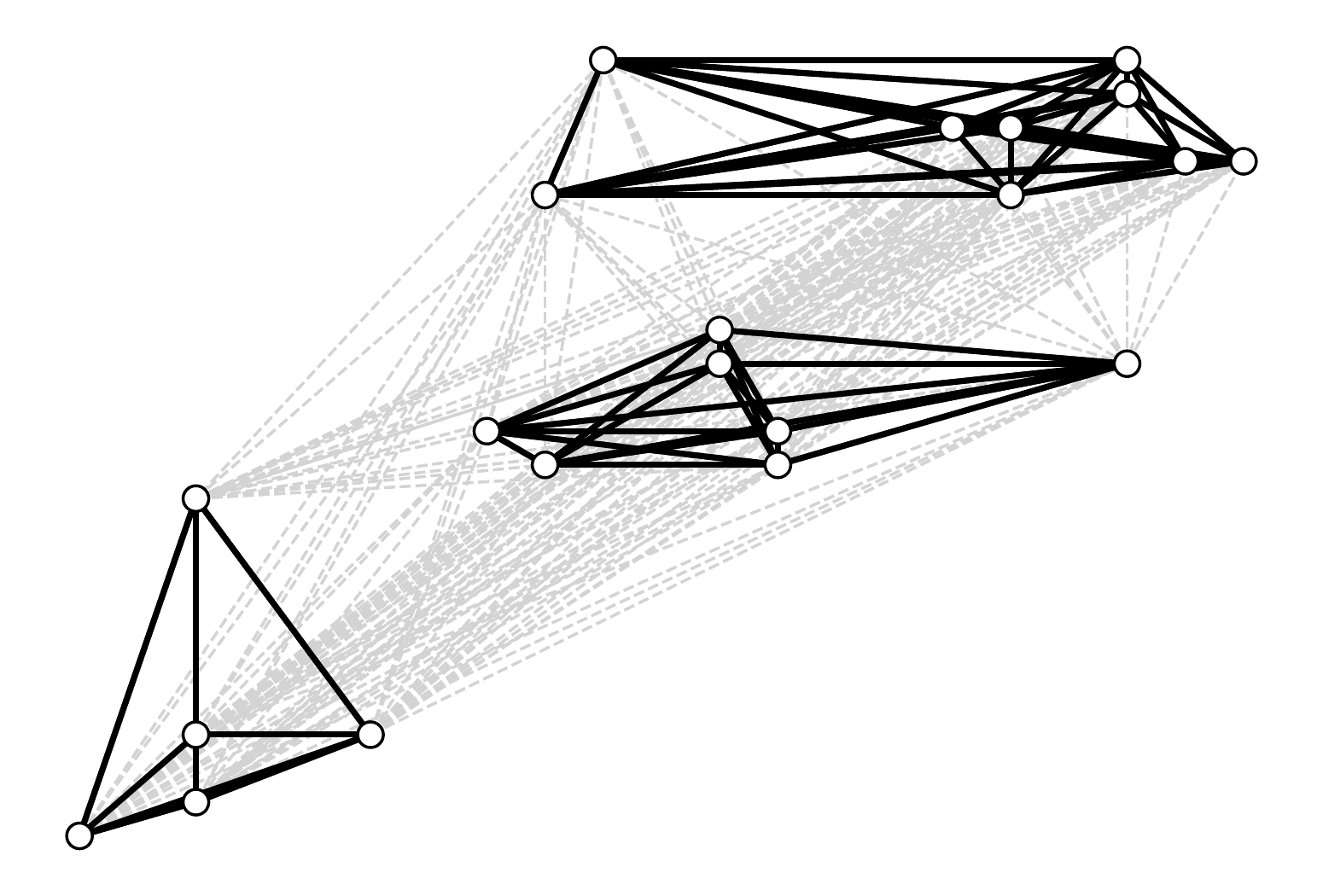} &
        \includegraphics[height=3.5cm]{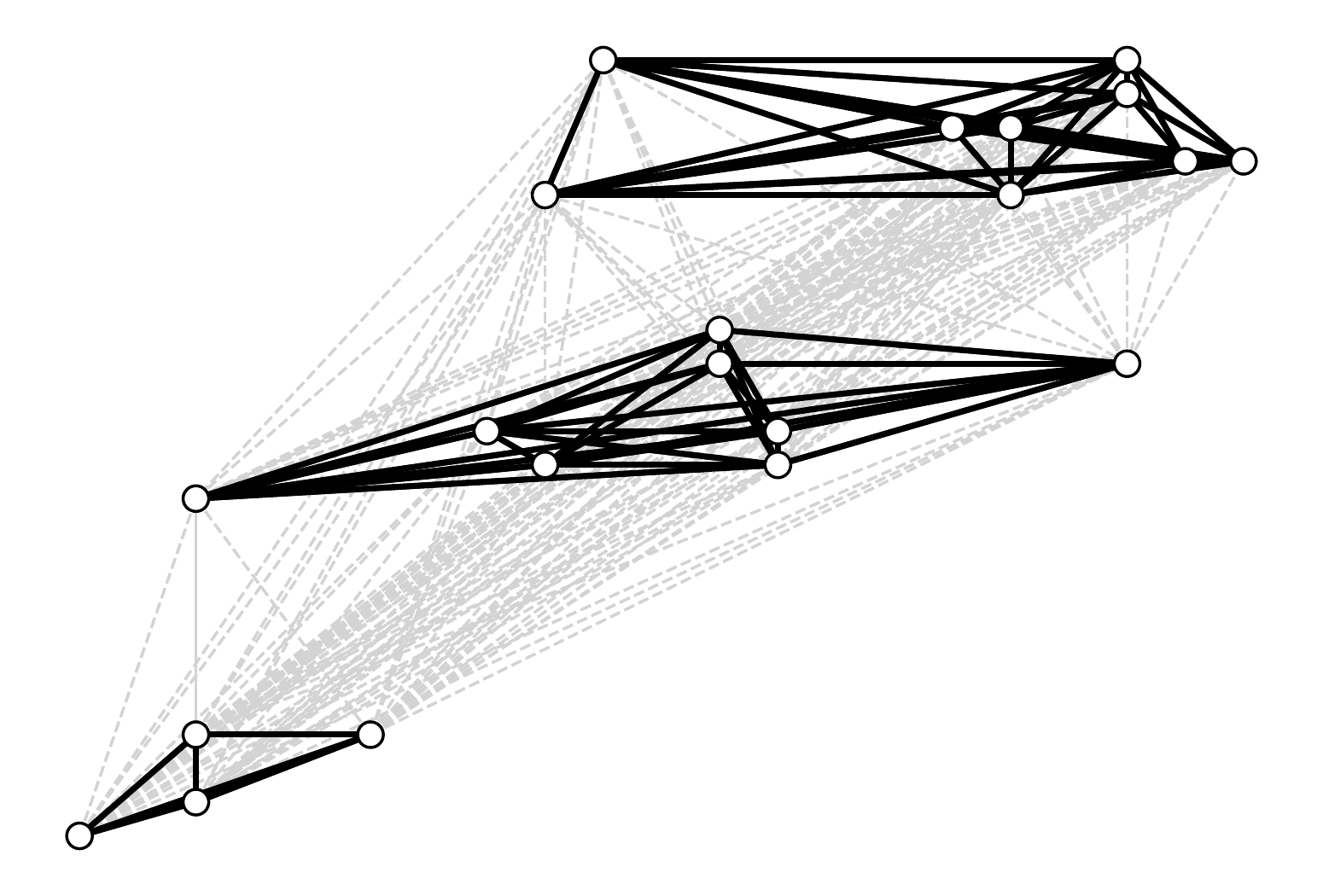} \\
        (a) Graph cut by model. &
        (b) Optimal solution. \\
        \includegraphics[height=4.0cm]{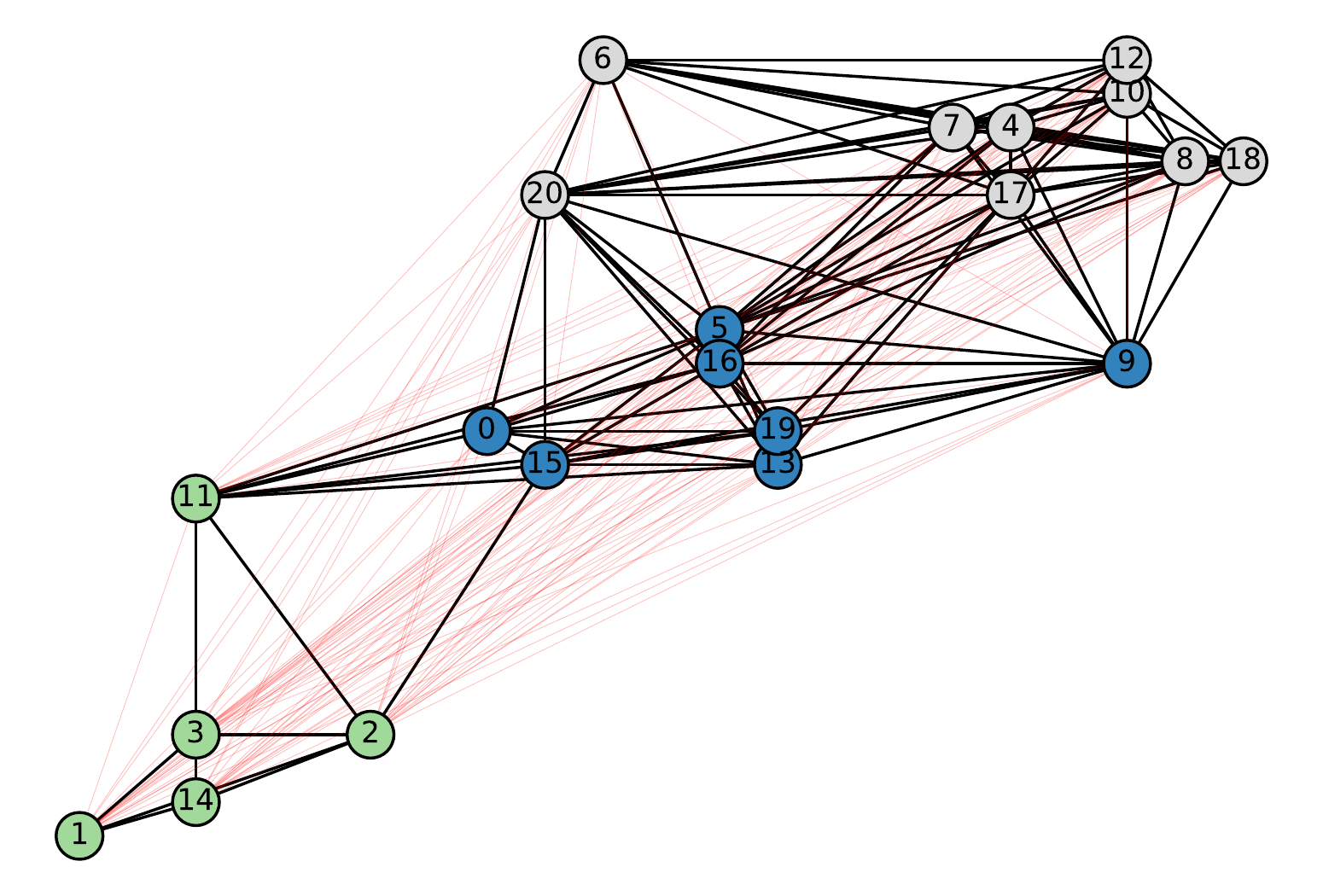} &
        \includegraphics[height=4.0cm]{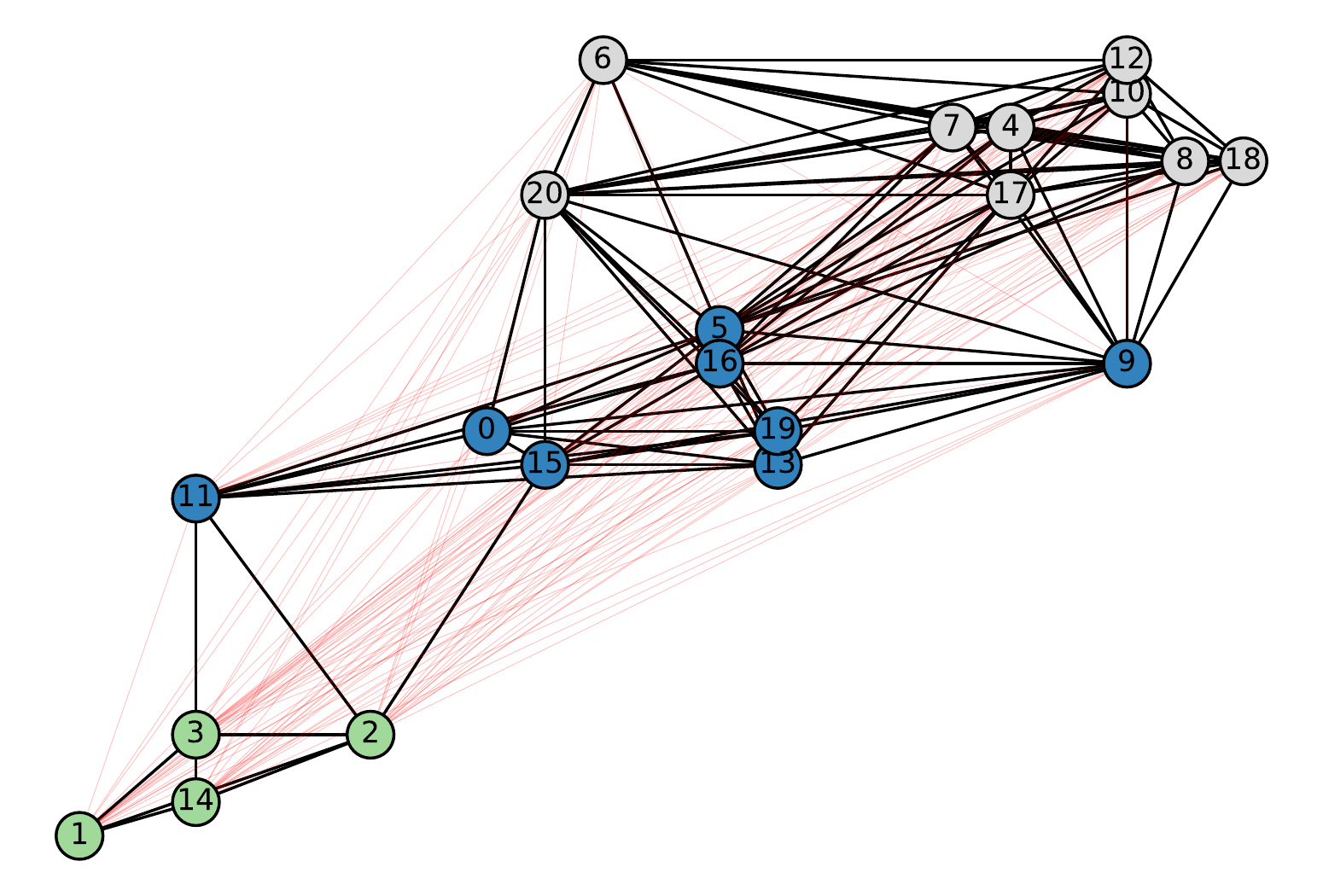} \\
        (c) Node clustering by model. &
        (d) Optimal solution. \\
        \includegraphics[height=4.6cm]{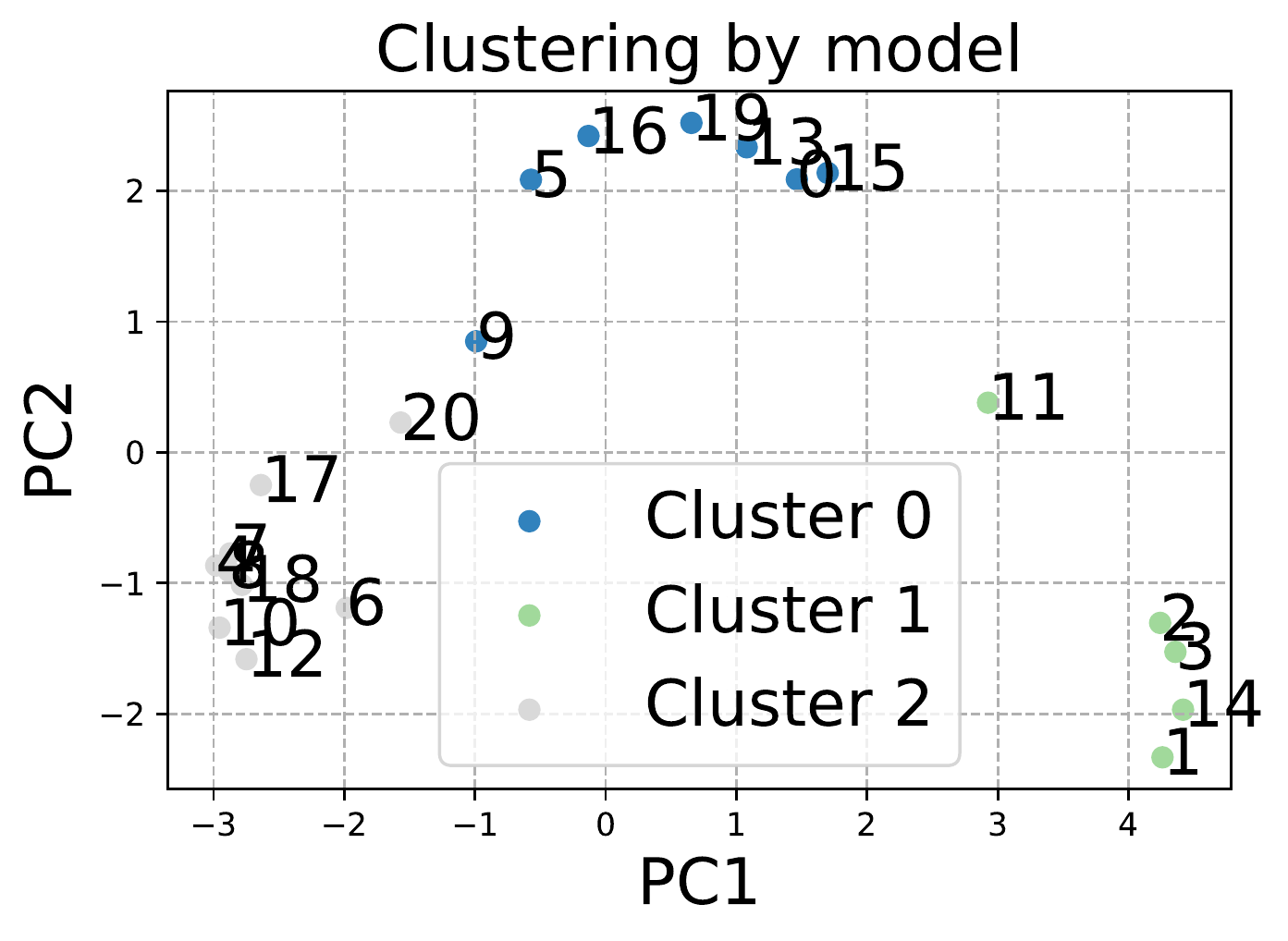} &
        \includegraphics[height=5cm]{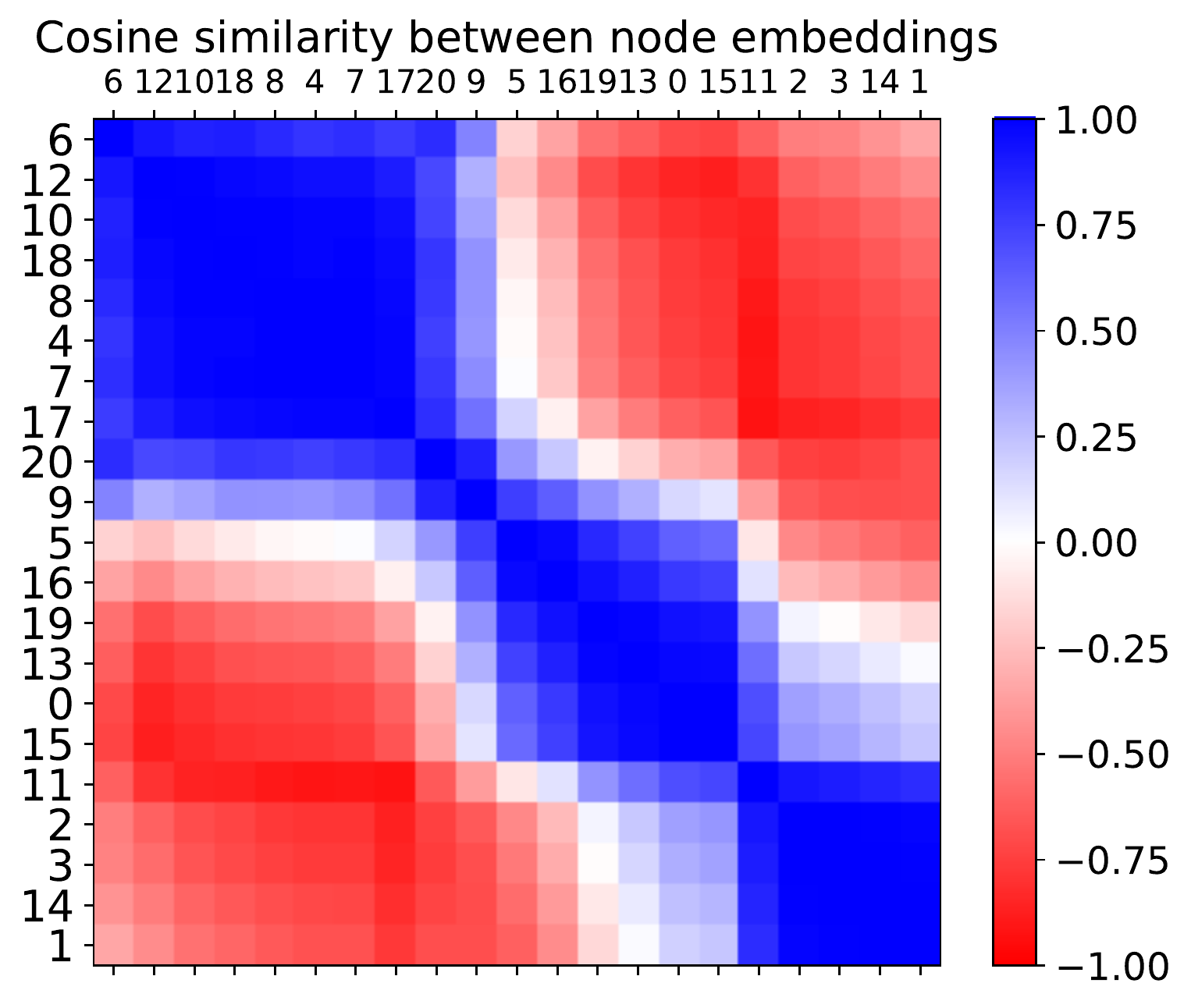} \\
        (e) Node clustering by model. &
        (f) Node embeddings. \\
    \end{tabular}
    \caption{
        (a) Graph cut solution computed by the model.
        (b) Graph cut of this graph according to the optimal solution.
        (c) Clustering of nodes according to the models' graph cut.
        (d) Clustering of nodes according to the optimal solution.
        (e) Node embeddings projected into a 2D space using PCA. Node colors according to the model prediction.
        (f) Cosine similarity between all node embeddings, ordered by similarity.
    }
    \label{fig:appendix-gcn-viz-0}
\end{figure}

\autoref{fig:appendix-gcn-viz-0} visualizes the results of our model (GCN\_W\_BN) on an IrisMP graph (\#0).

\begin{figure}[H]
    \centering
    \setlength\tabcolsep{0pt} 
    \begin{tabular}{cc}
        \includegraphics[height=3.5cm]{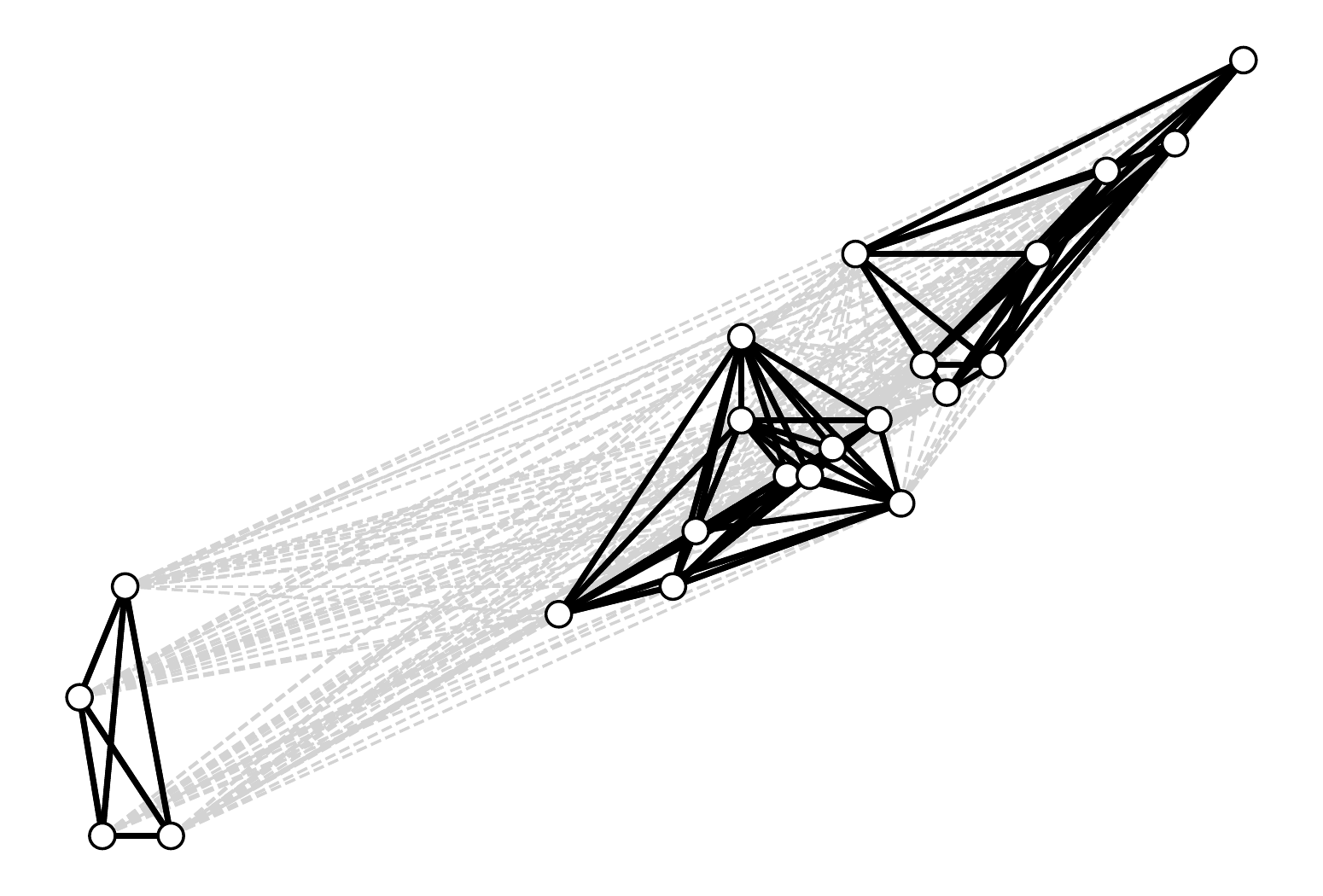} &
        \includegraphics[height=3.5cm]{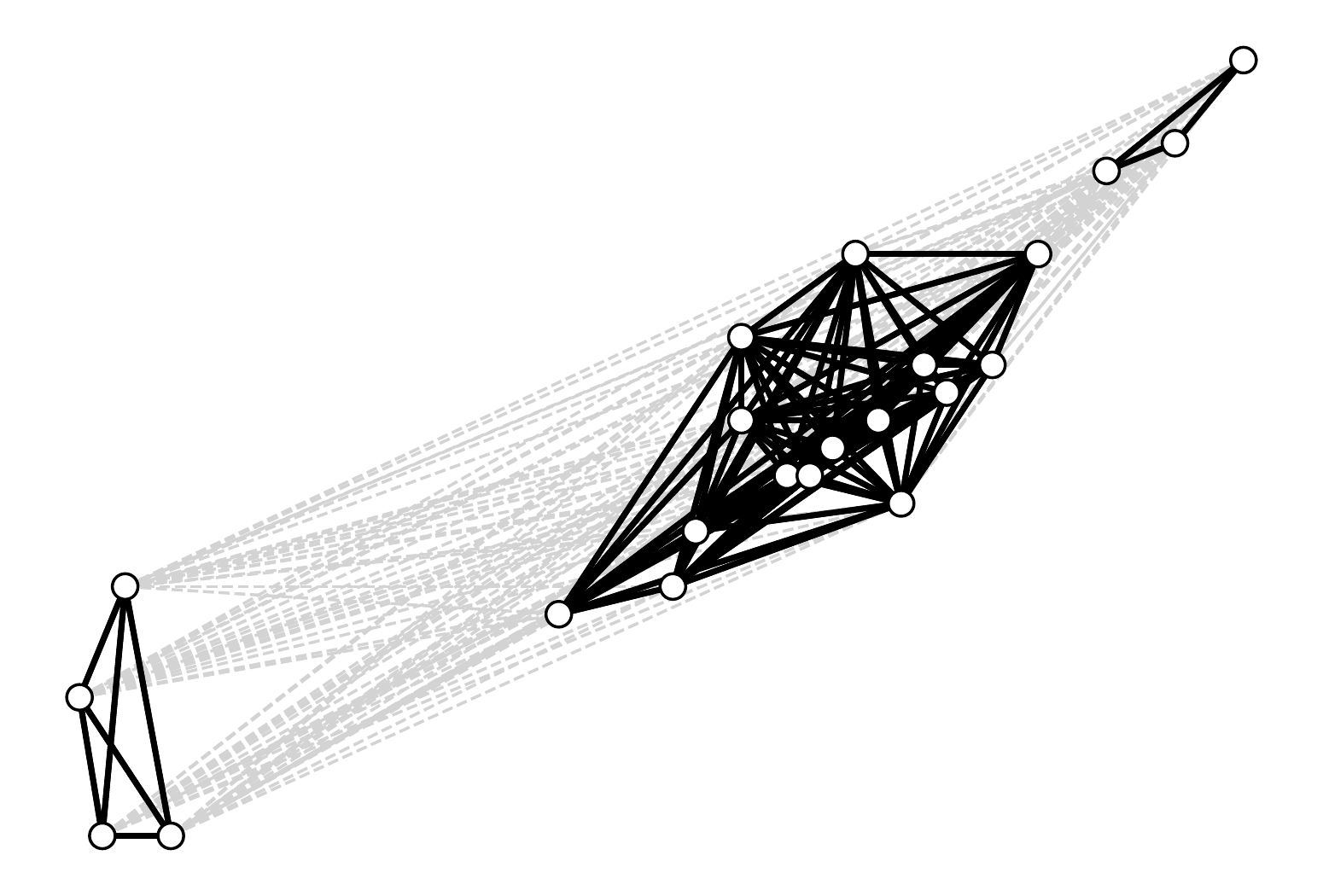} \\
        (a) Graph cut by model. &
        (b) Optimal solution. \\
        \includegraphics[height=4cm]{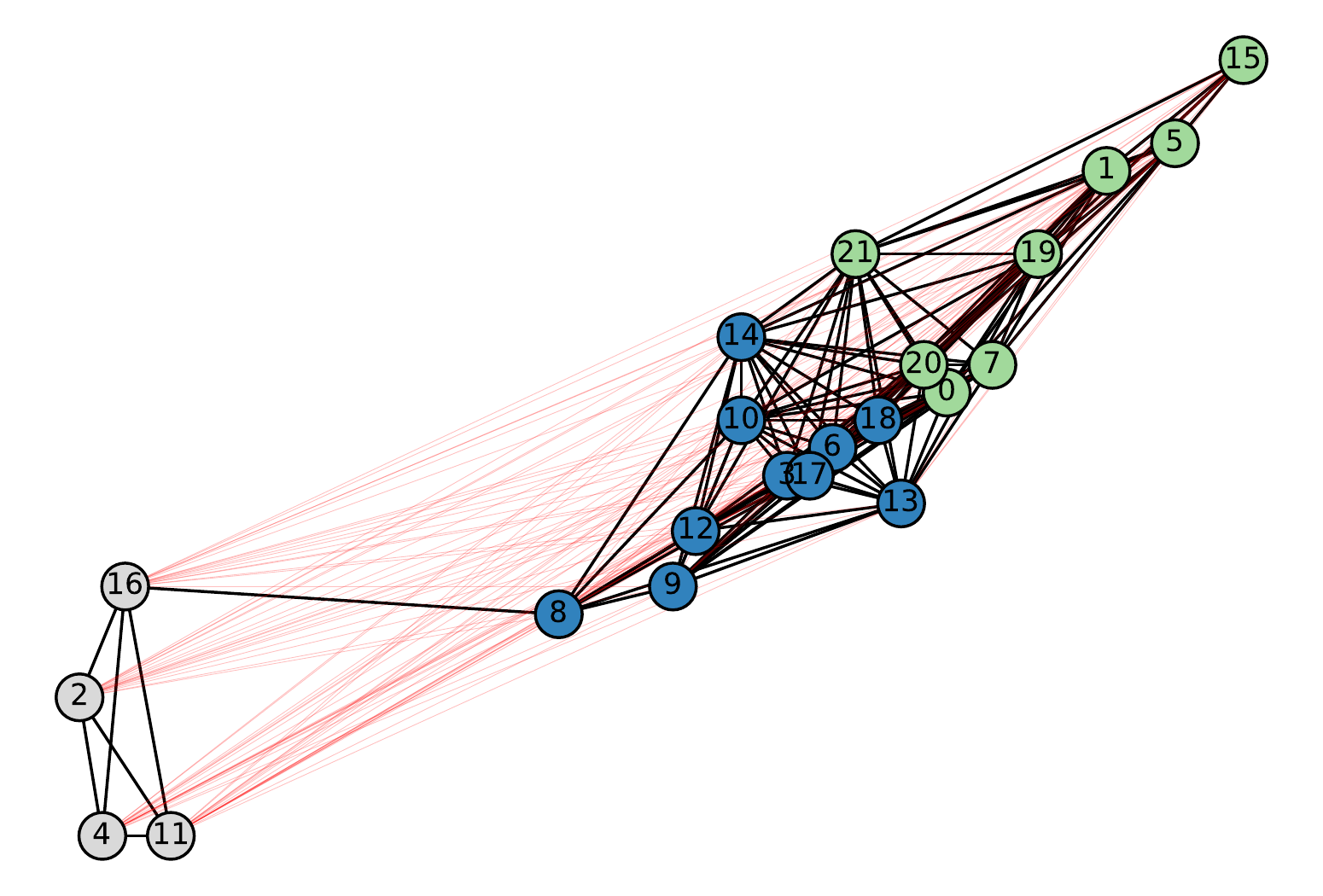} &
        \includegraphics[height=4cm]{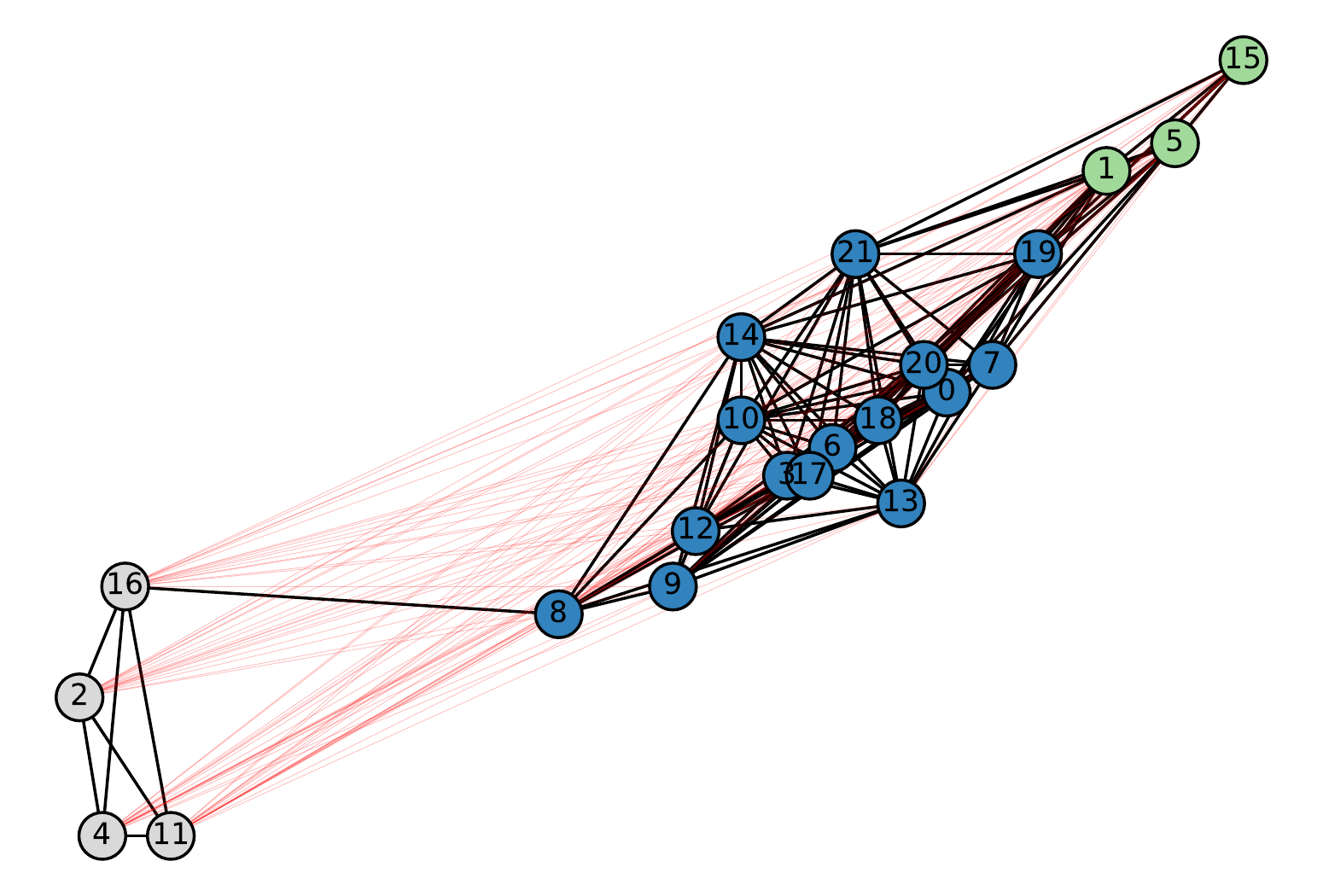} \\
        (c) Node clustering by model. &
        (d) Optimal solution. \\
        \includegraphics[height=4.6cm]{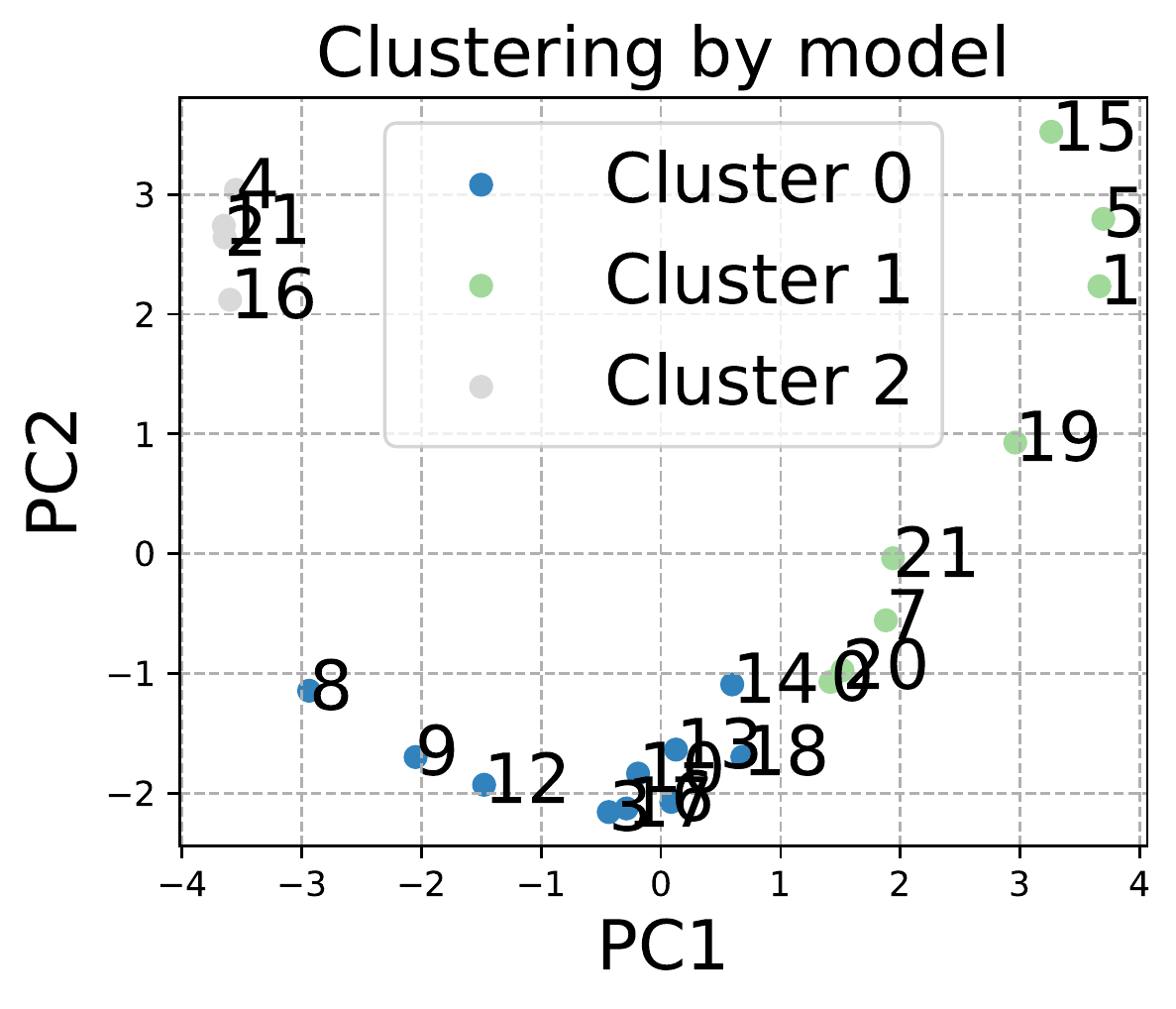} &
        \includegraphics[height=5cm]{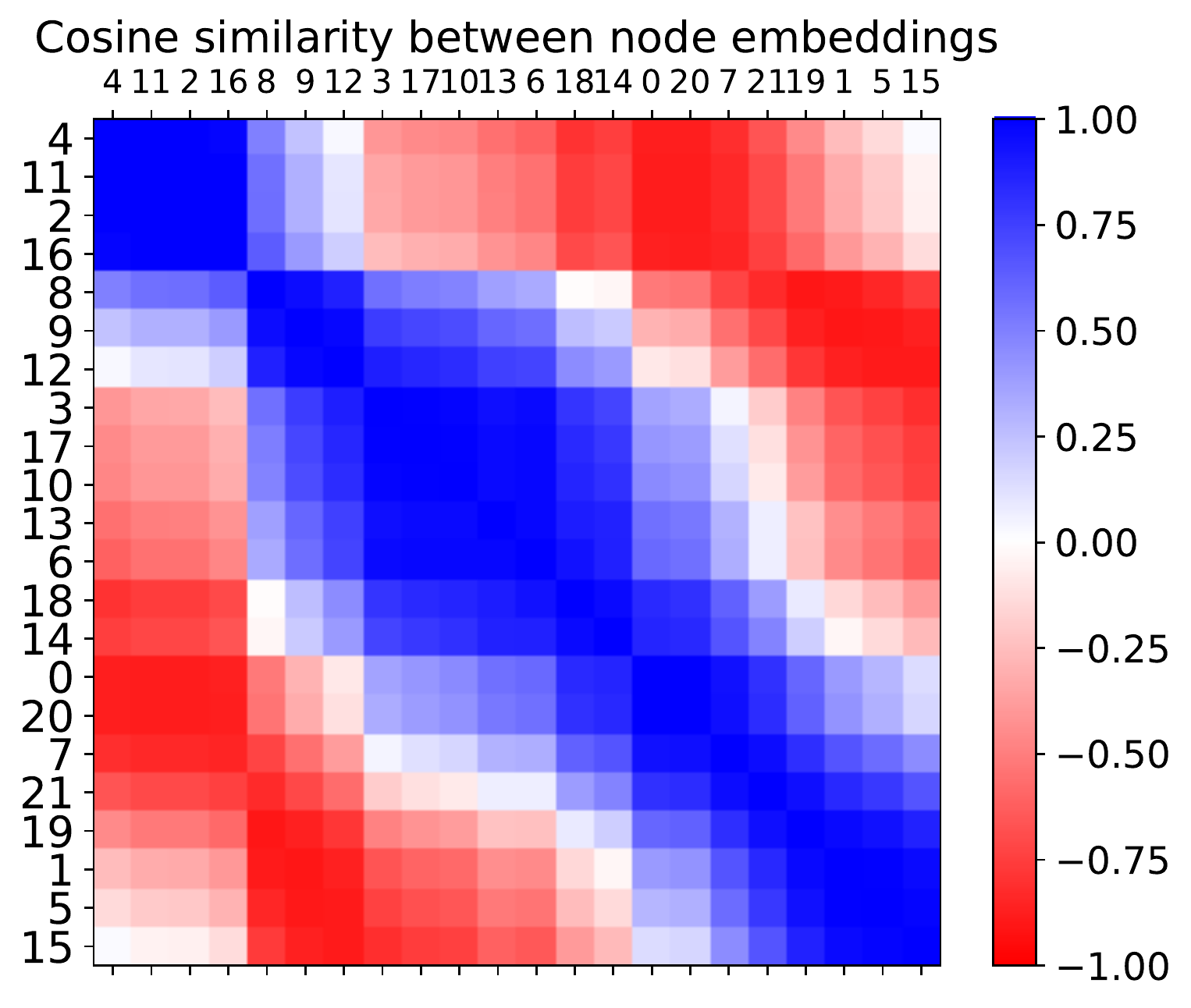} \\
        (e) Node clustering by model. &
        (f) Node embeddings. \\
    \end{tabular}
    \caption{
        (a) Graph cut solution computed by the model.
        (b) Graph cut of this graph according to the optimal solution.
        (c) Clustering of nodes according to the models' graph cut.
        (d) Clustering of nodes according to the optimal solution.
        (e) Node embeddings projected into a 2D space using PCA. Node colors according to the model prediction.
        (f) Cosine similarity between all node embeddings, ordered by similarity.
    }
    \label{fig:appendix-gcn-viz-1}
\end{figure}

\newpage
\noindent
\autoref{fig:appendix-gcn-viz-1} visualizes the results of our model (GCN\_W\_BN) on an IrisMP graph (\#1).

\begin{figure}[H]
    \centering
    \setlength\tabcolsep{0pt} 
    \begin{tabular}{cc}
        \includegraphics[height=3.5cm]{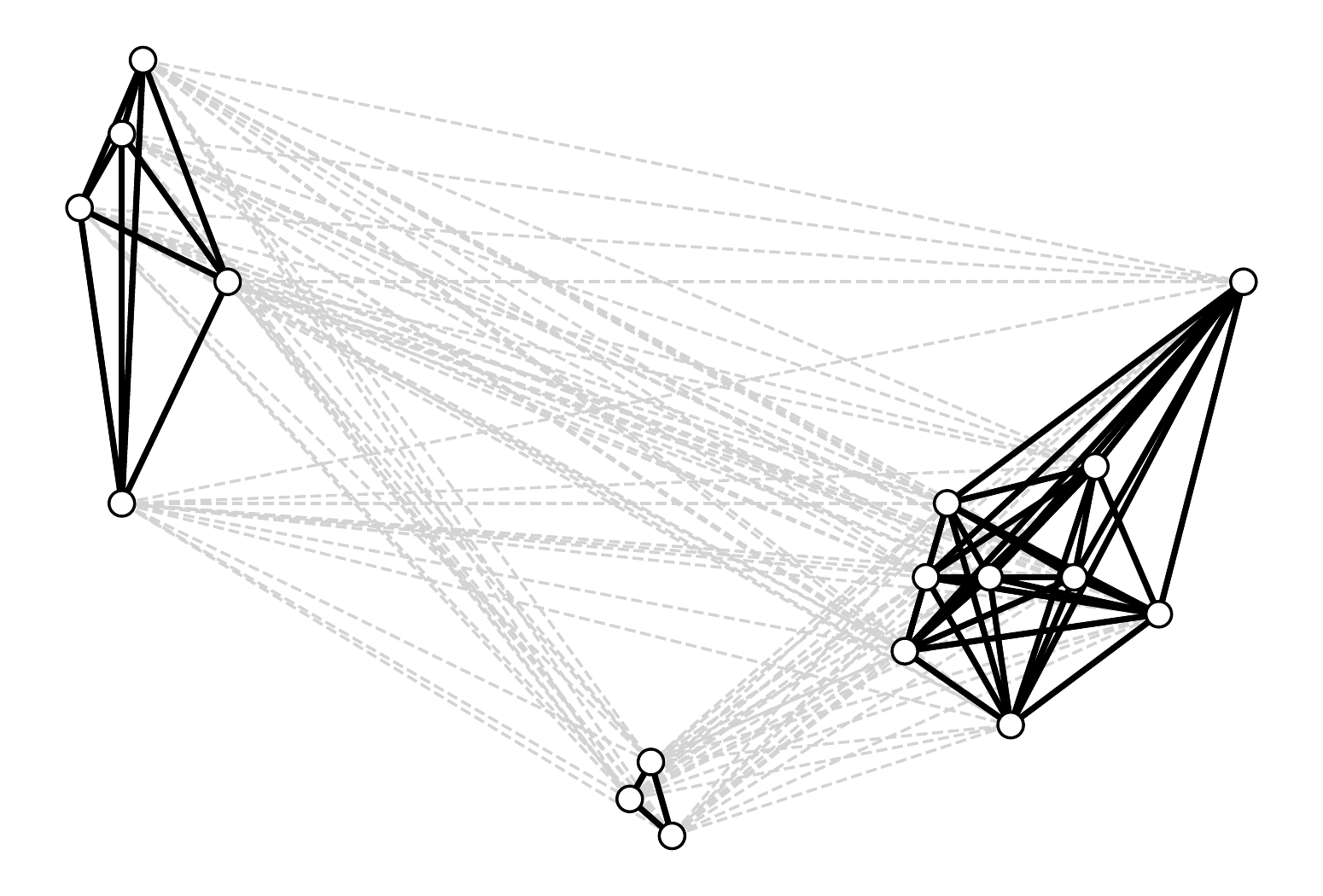} &
        \includegraphics[height=3.5cm]{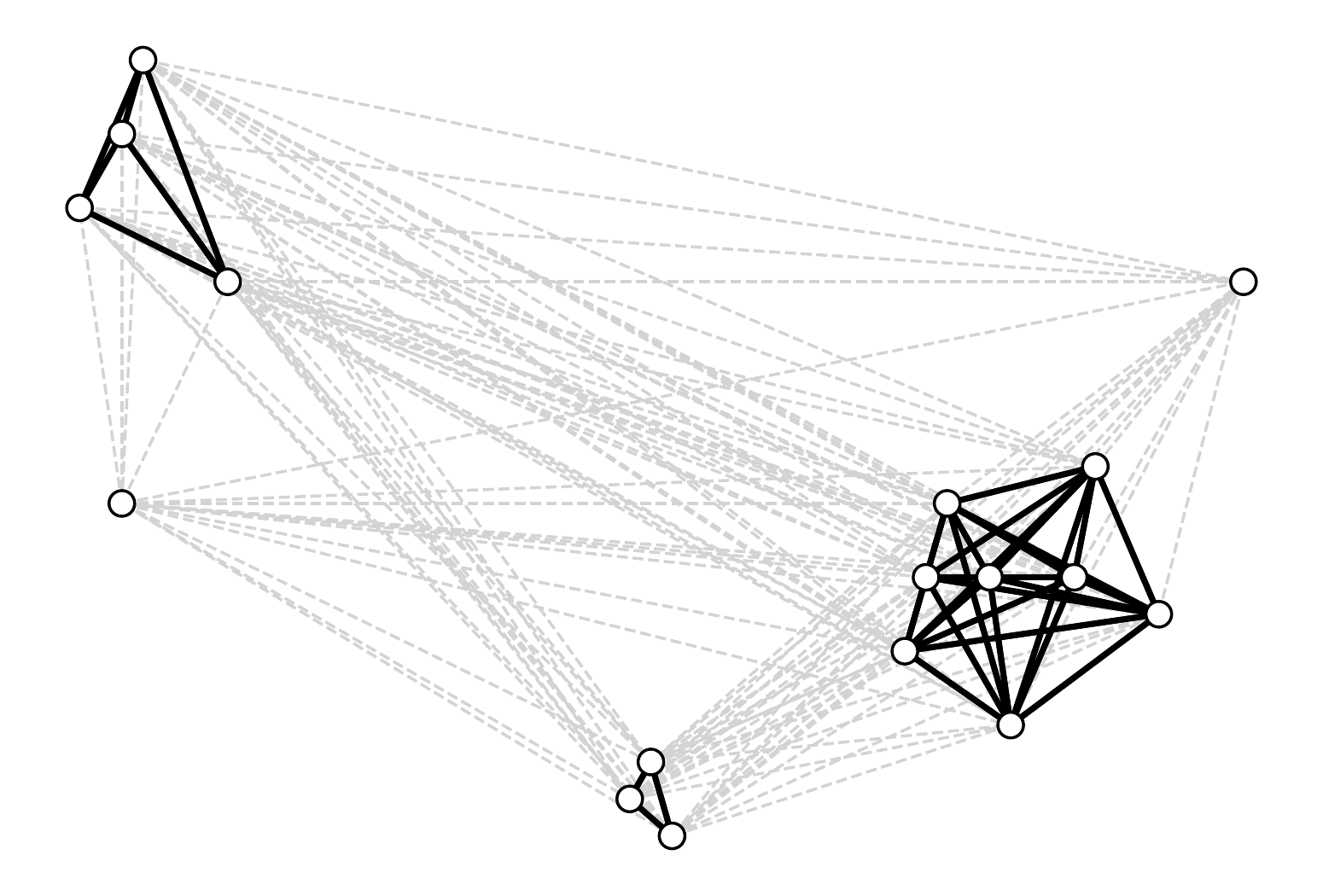} \\
        (a) Graph cut by model. &
        (b) Optimal solution. \\
        \includegraphics[height=4cm]{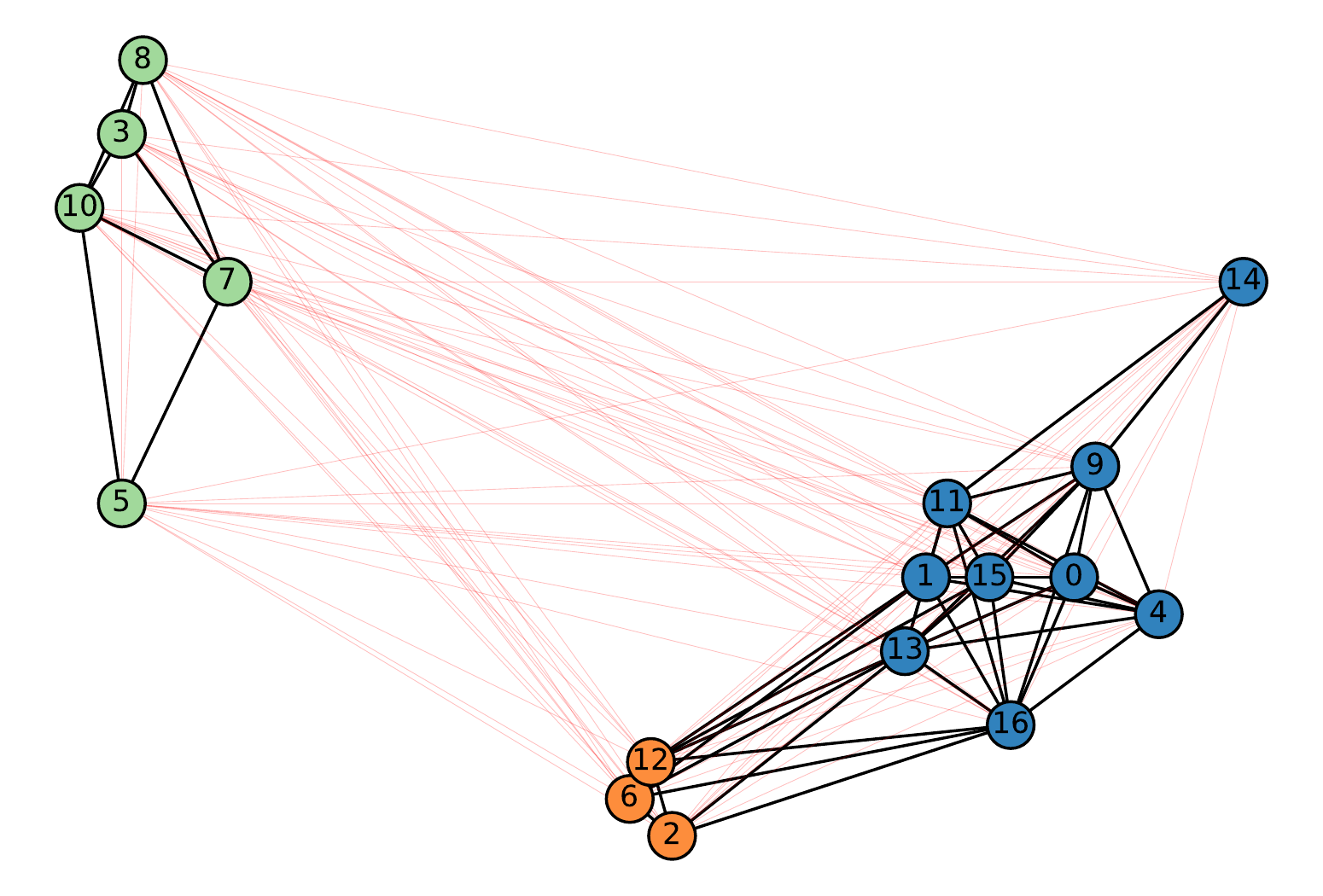} &
        \includegraphics[height=4cm]{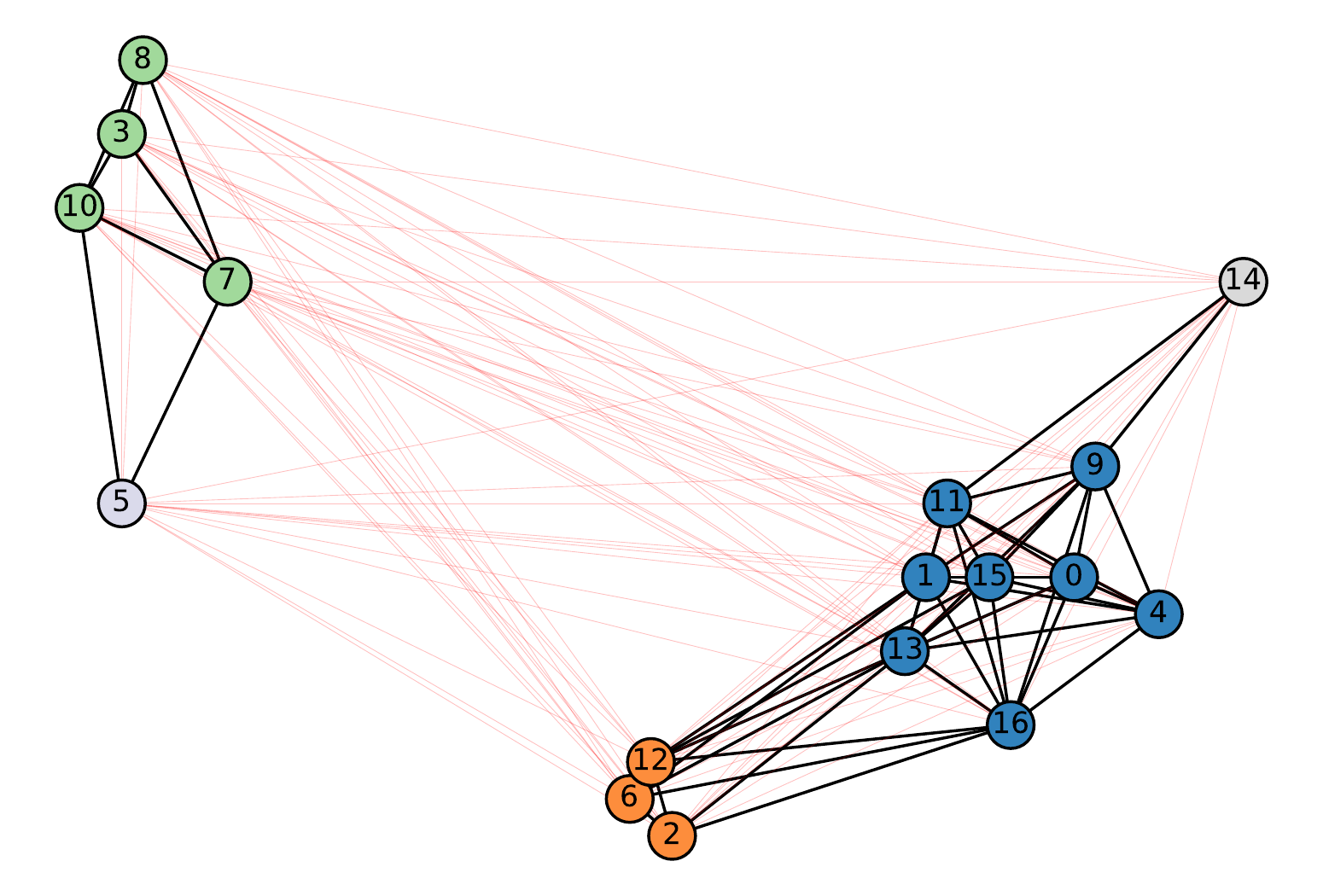} \\
        (c) Node clustering by model. &
        (d) Optimal solution. \\
        \includegraphics[height=4.6cm]{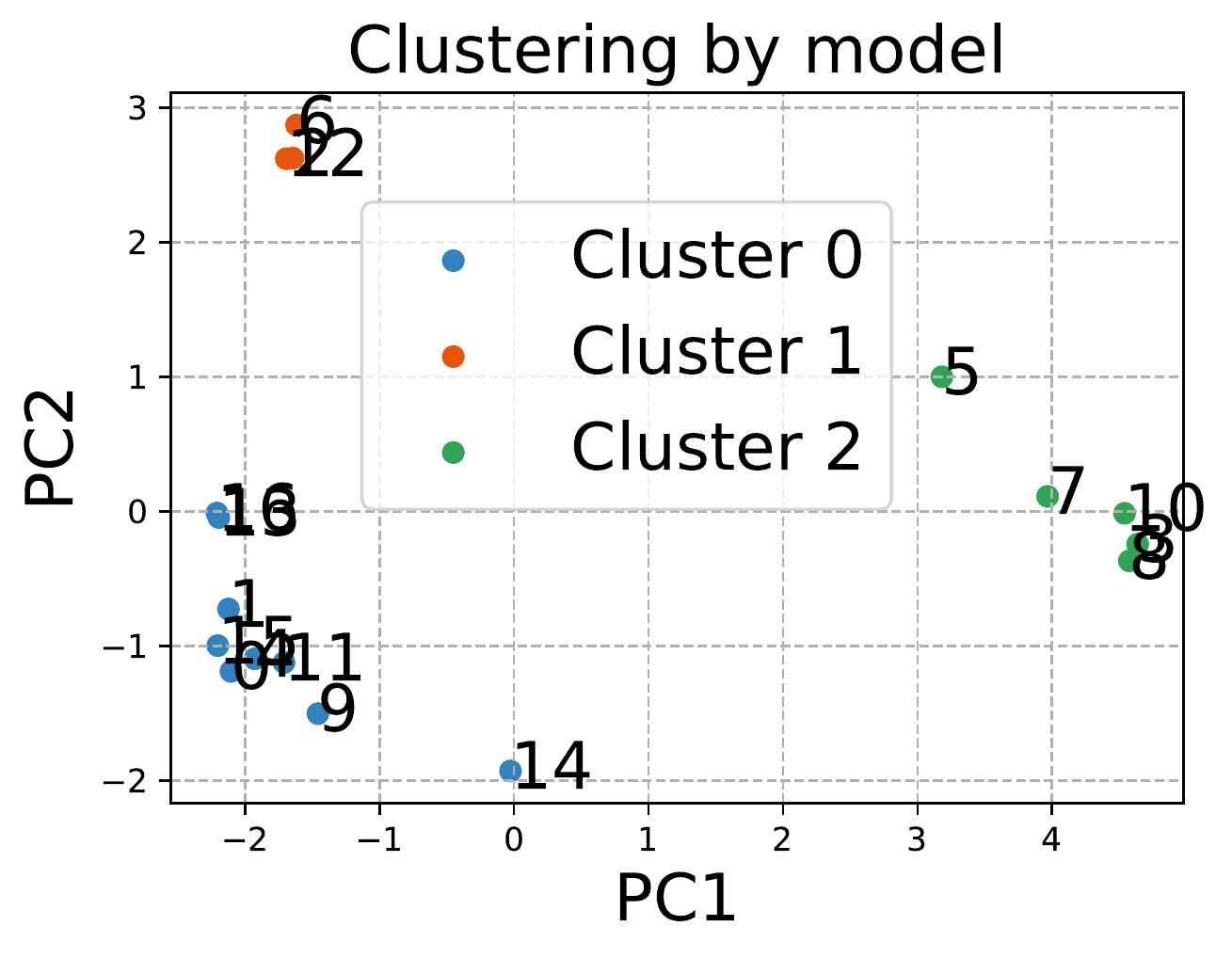} &
        \includegraphics[height=5cm]{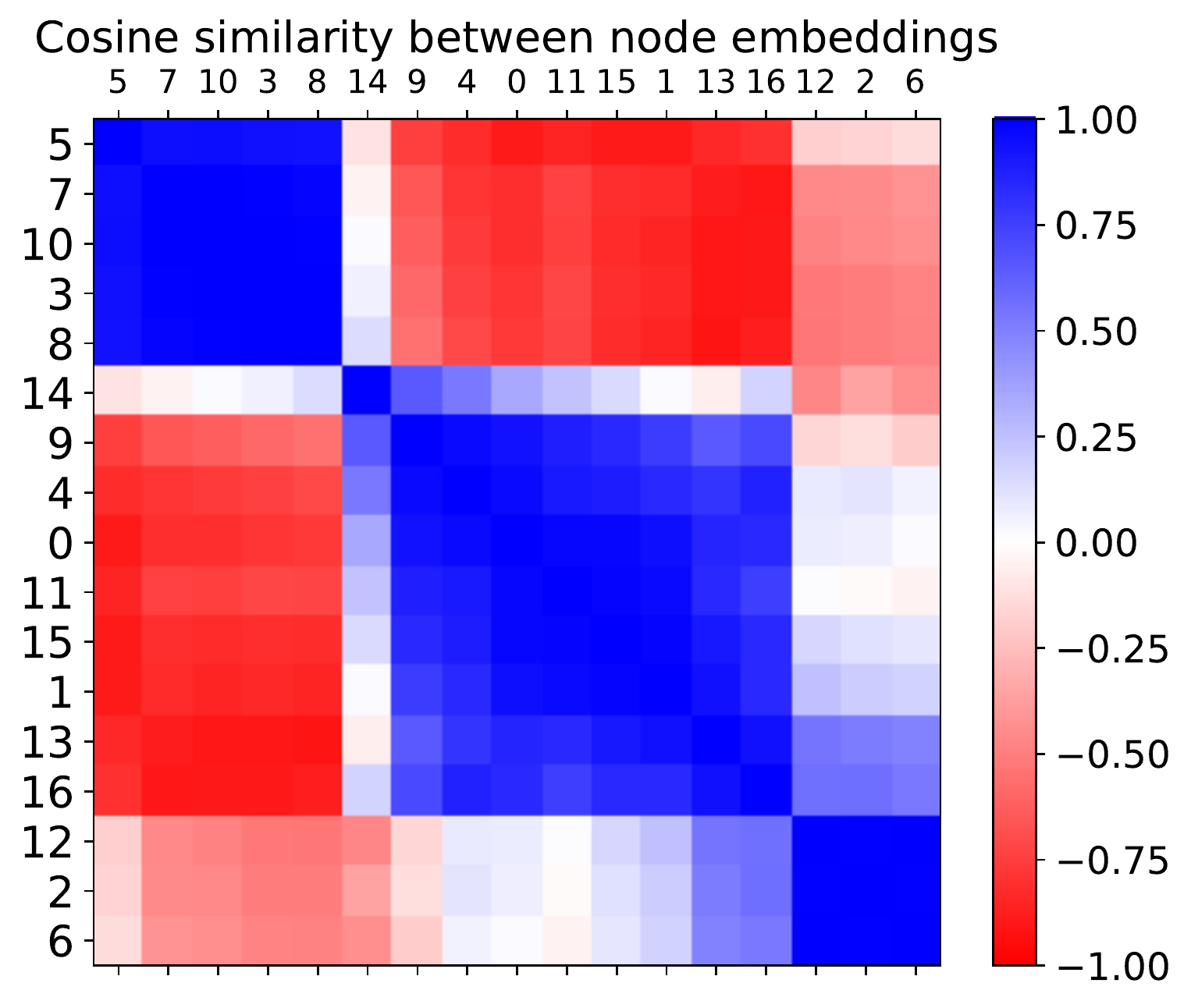} \\
        (e) Node clustering by model. &
        (f) Node embeddings. \\
    \end{tabular}
    \caption{
        (a) Graph cut solution computed by the model.
        (b) Graph cut of this graph according to the optimal solution.
        (c) Clustering of nodes according to the models' graph cut.
        (d) Clustering of nodes according to the optimal solution.
        (e) Node embeddings projected into a 2D space using PCA. Node colors according to the model prediction.
        (f) Cosine similarity between all node embeddings, ordered by similarity.
    }
    \label{fig:appendix-gcn-viz-16}
\end{figure}

\newpage
\noindent
\autoref{fig:appendix-gcn-viz-16} visualizes the results of our model (GCN\_W\_BN) on an IrisMP graph (\#16).

\begin{figure}[H]
    \centering
    \setlength\tabcolsep{0pt} 
    \begin{tabular}{cc}
        \includegraphics[height=3.5cm]{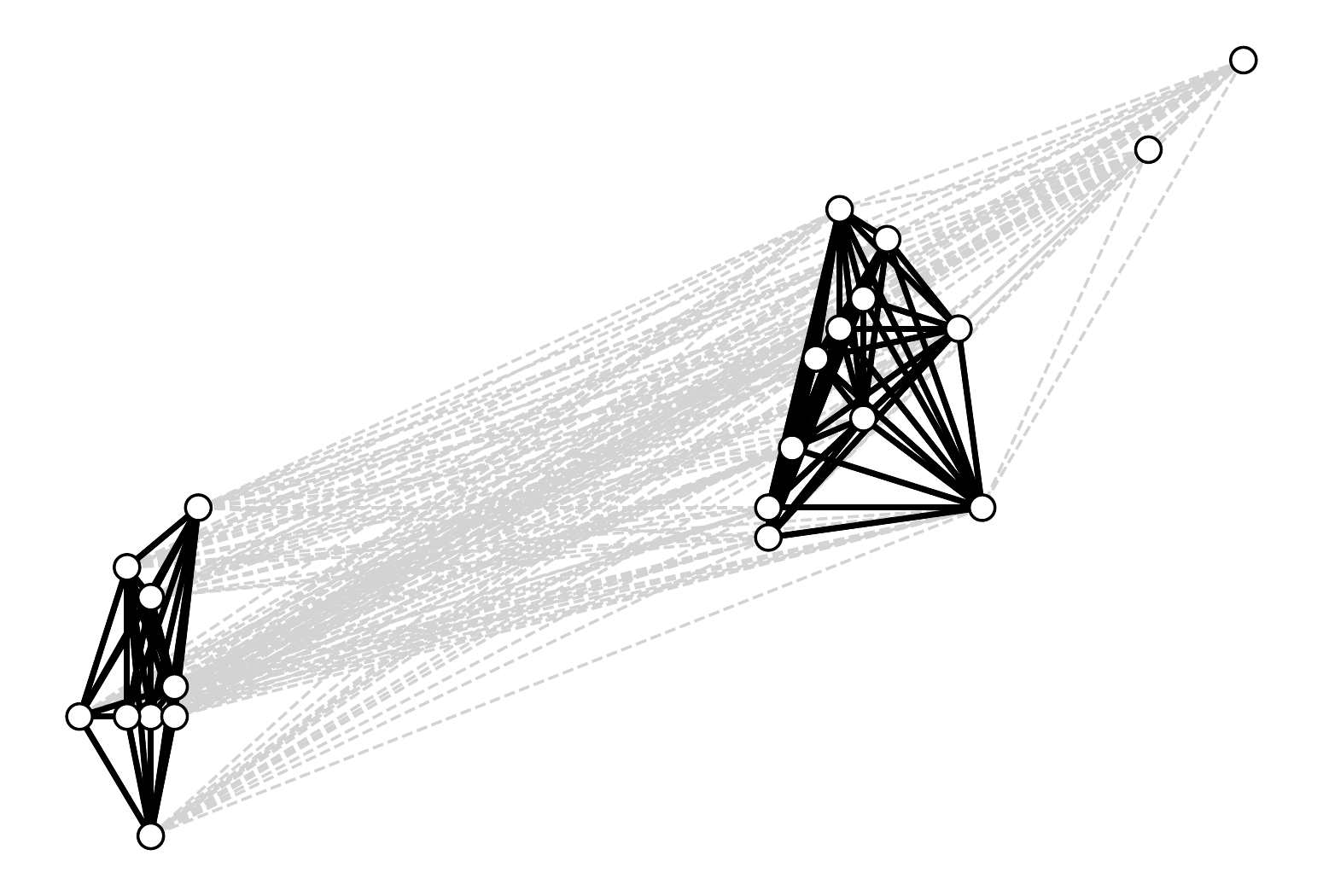} &
        \includegraphics[height=3.5cm]{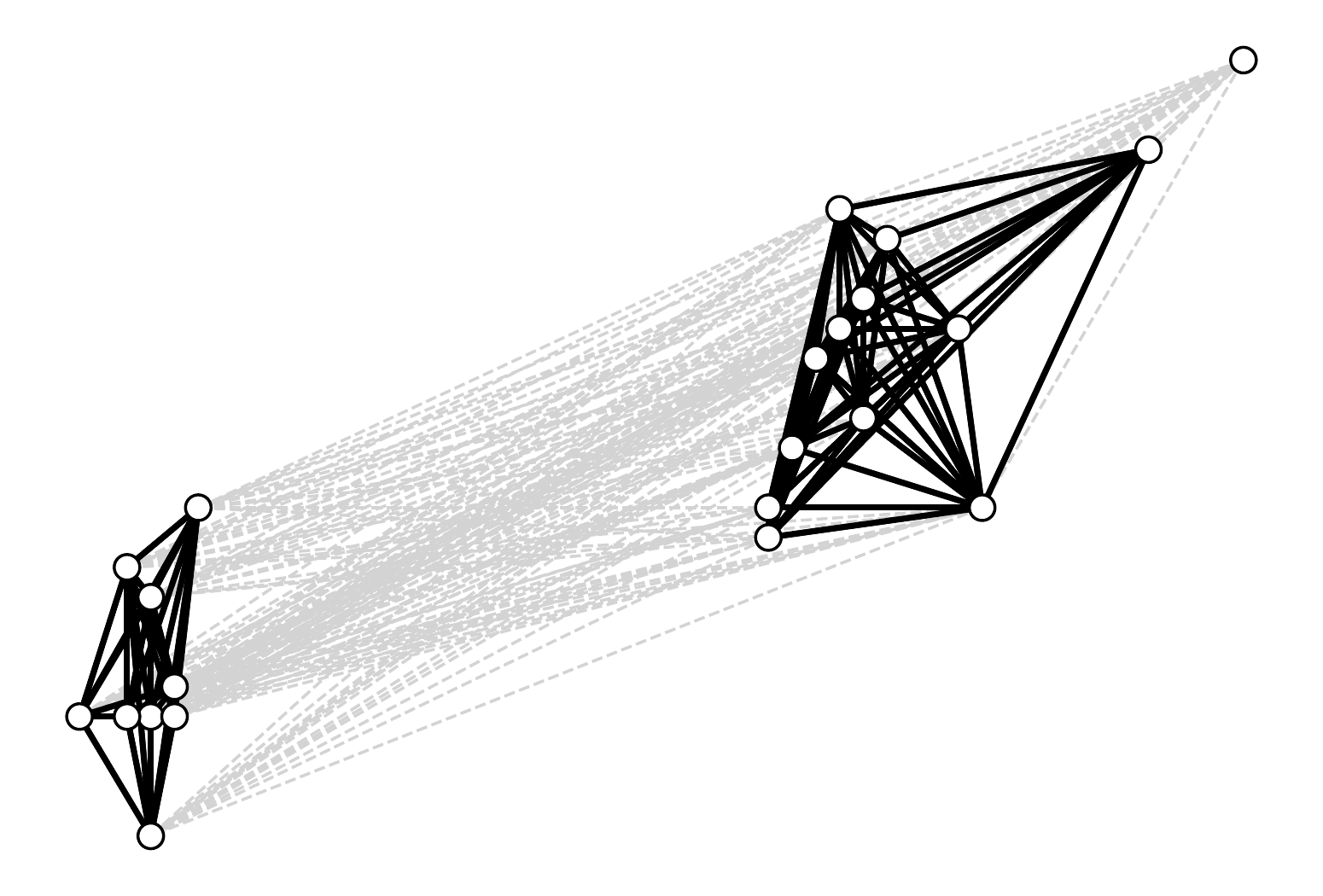} \\
        (a) Graph cut by model. &
        (b) Optimal solution. \\
        \includegraphics[height=4cm]{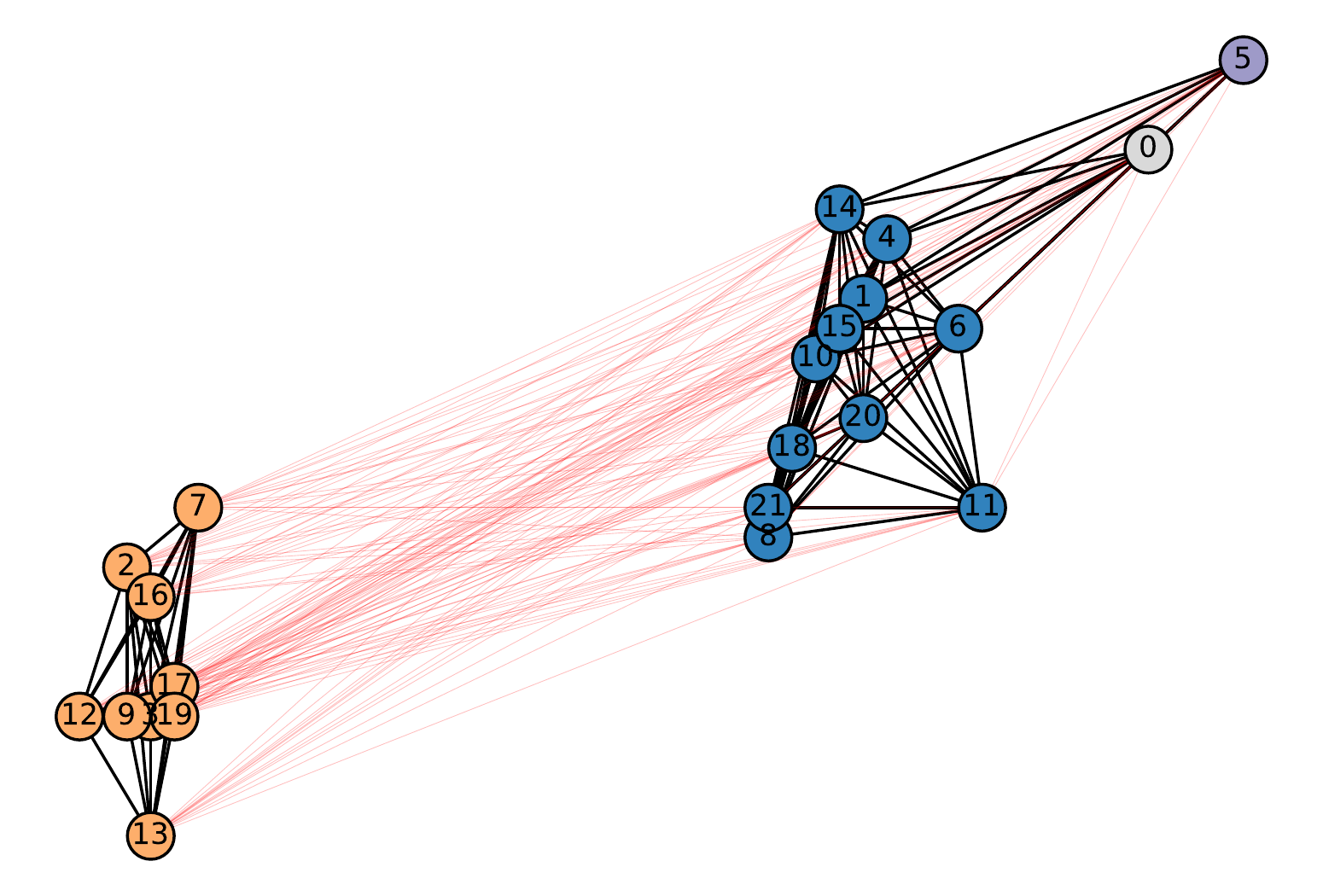} &
        \includegraphics[height=4cm]{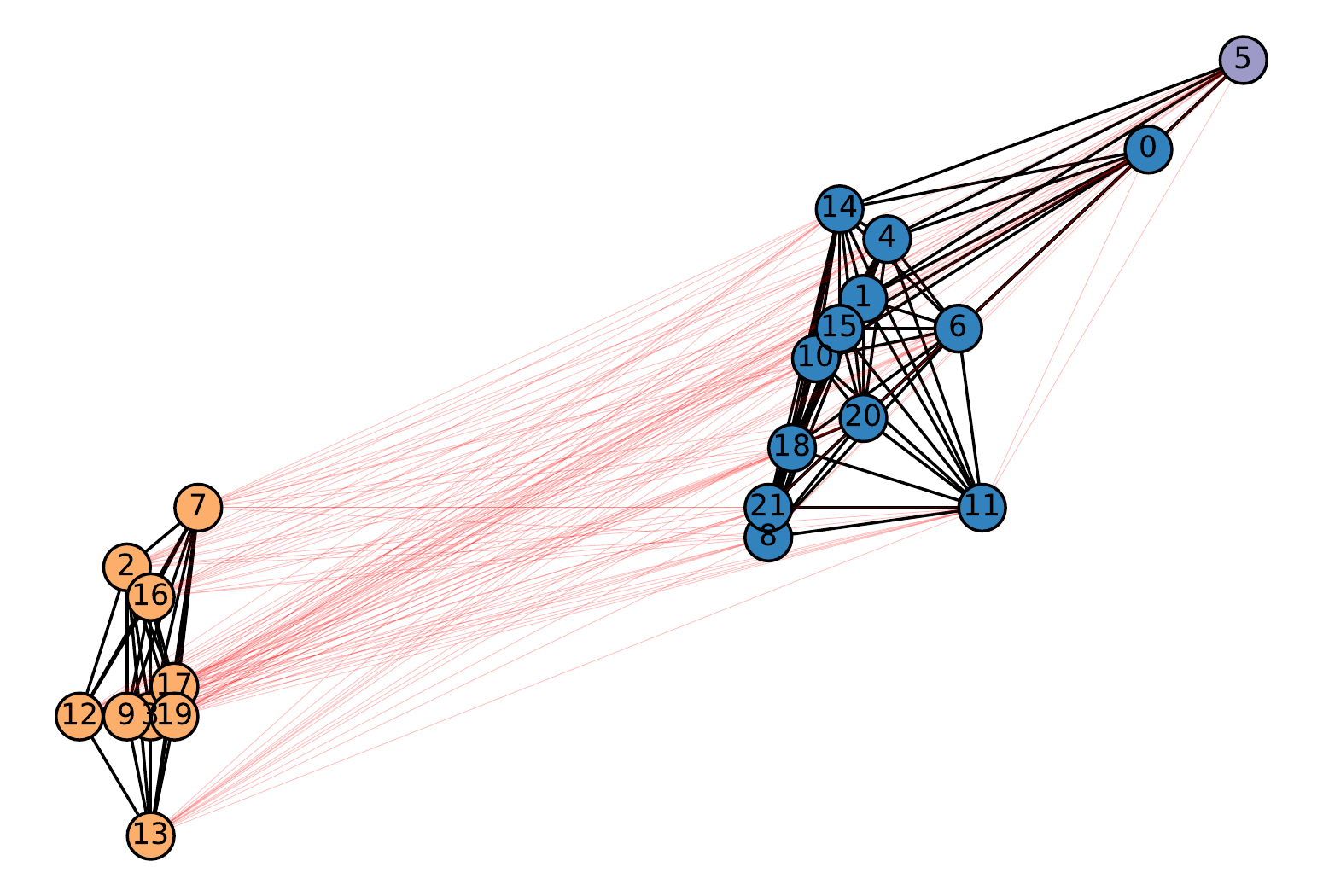} \\
        (c) Node clustering by model. &
        (d) Optimal solution. \\
        \includegraphics[height=4.6cm]{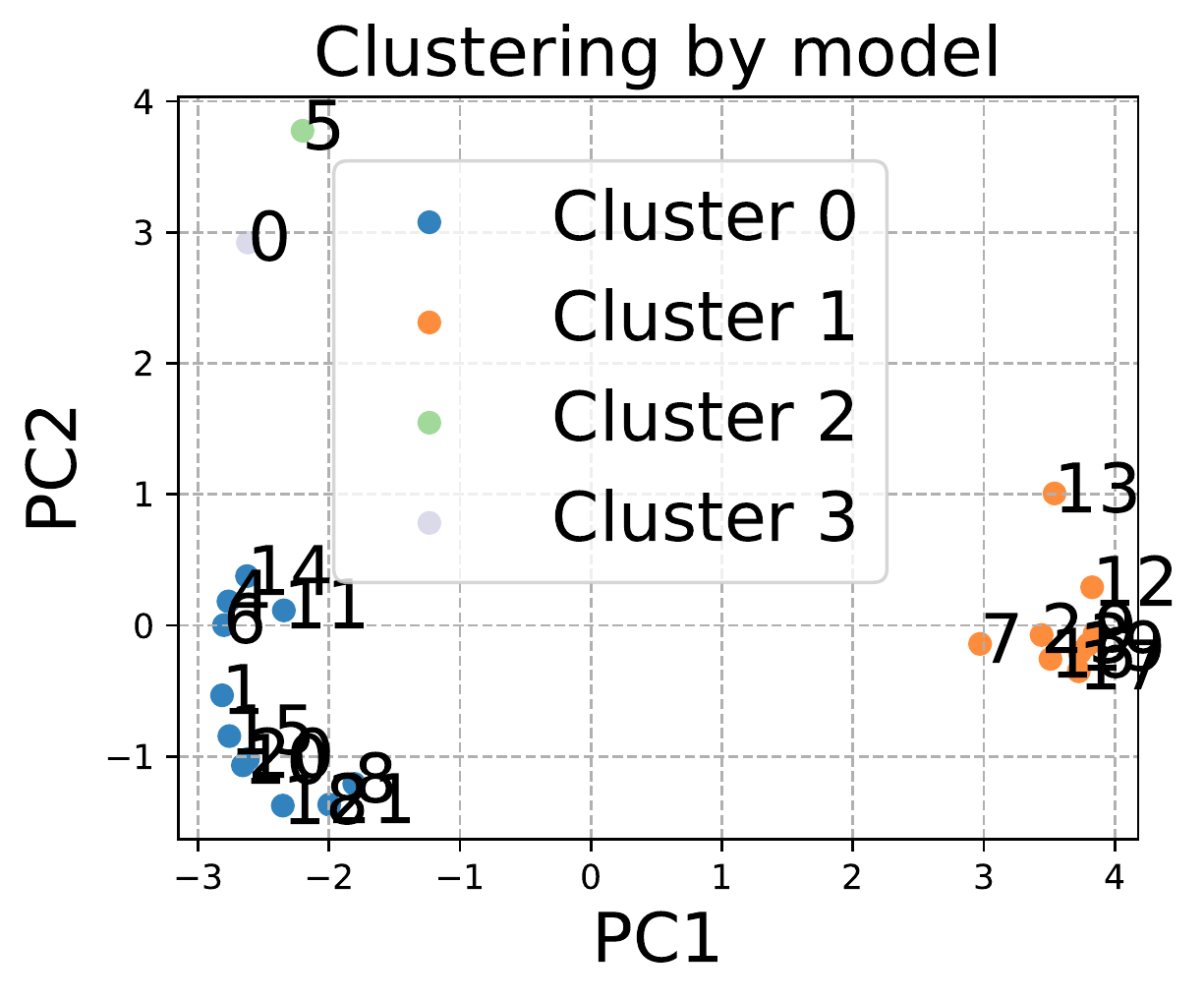} &
        \includegraphics[height=5cm]{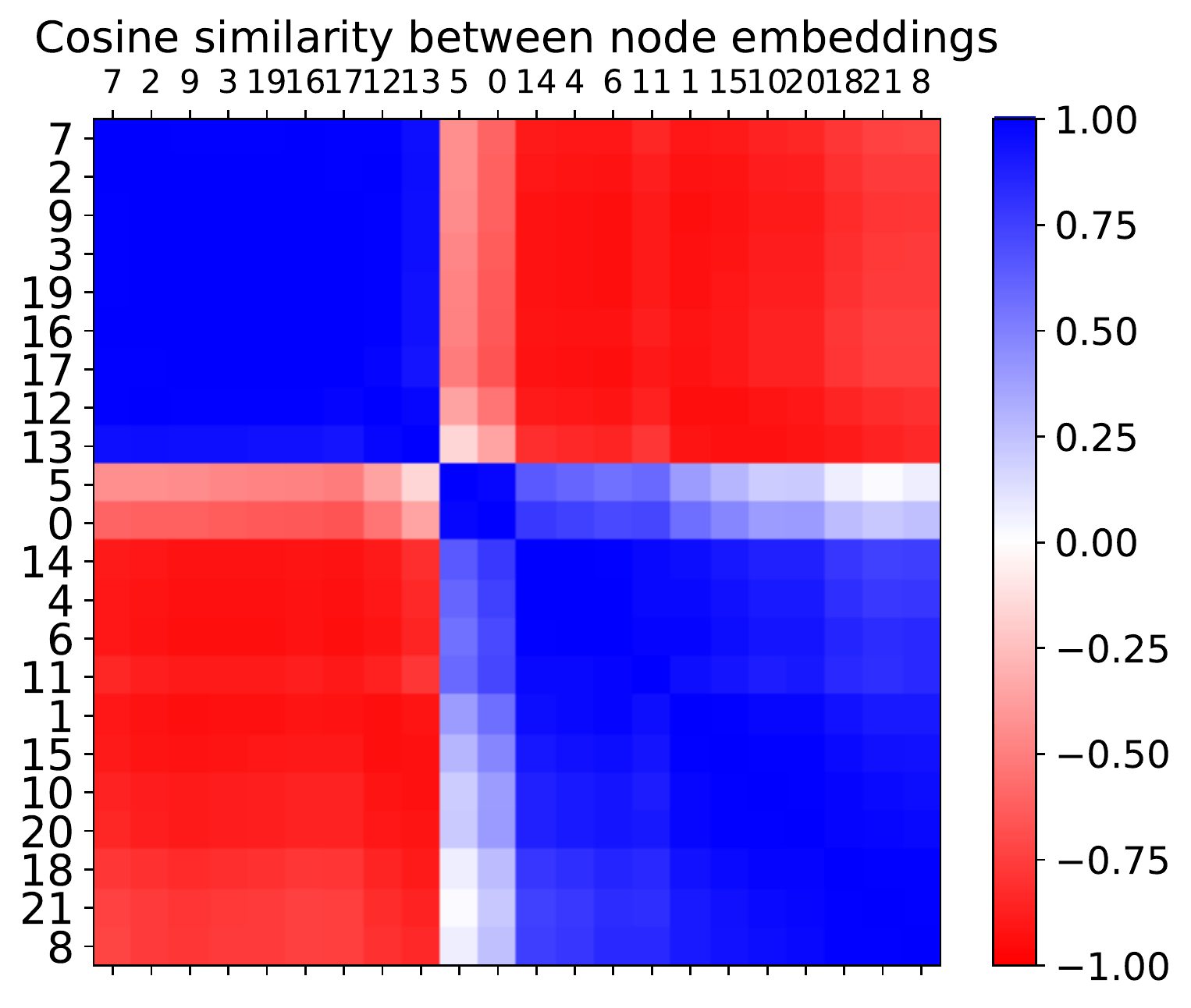} \\
        (e) Node clustering by model. &
        (f) Node embeddings. \\
    \end{tabular}
    \caption{
        (a) Graph cut solution computed by the model.
        (b) Graph cut of this graph according to the optimal solution.
        (c) Clustering of nodes according to the models' graph cut.
        (d) Clustering of nodes according to the optimal solution.
        (e) Node embeddings projected into a 2D space using PCA. Node colors according to the model prediction.
        (f) Cosine similarity between all node embeddings, ordered by similarity.
    }
    \label{fig:appendix-gcn-viz-17}
\end{figure}

\newpage
\noindent
\autoref{fig:appendix-gcn-viz-17} visualizes the results of our model (GCN\_W\_BN) on an IrisMP graph (\#17).

\newpage
\section{Training settings}
\label{sec:A-compute-time}

\begin{figure*}[h]
    \centering
    \setlength\tabcolsep{0pt} 
    \begin{tabular}{ccc}
        \includegraphics[height=3cm]{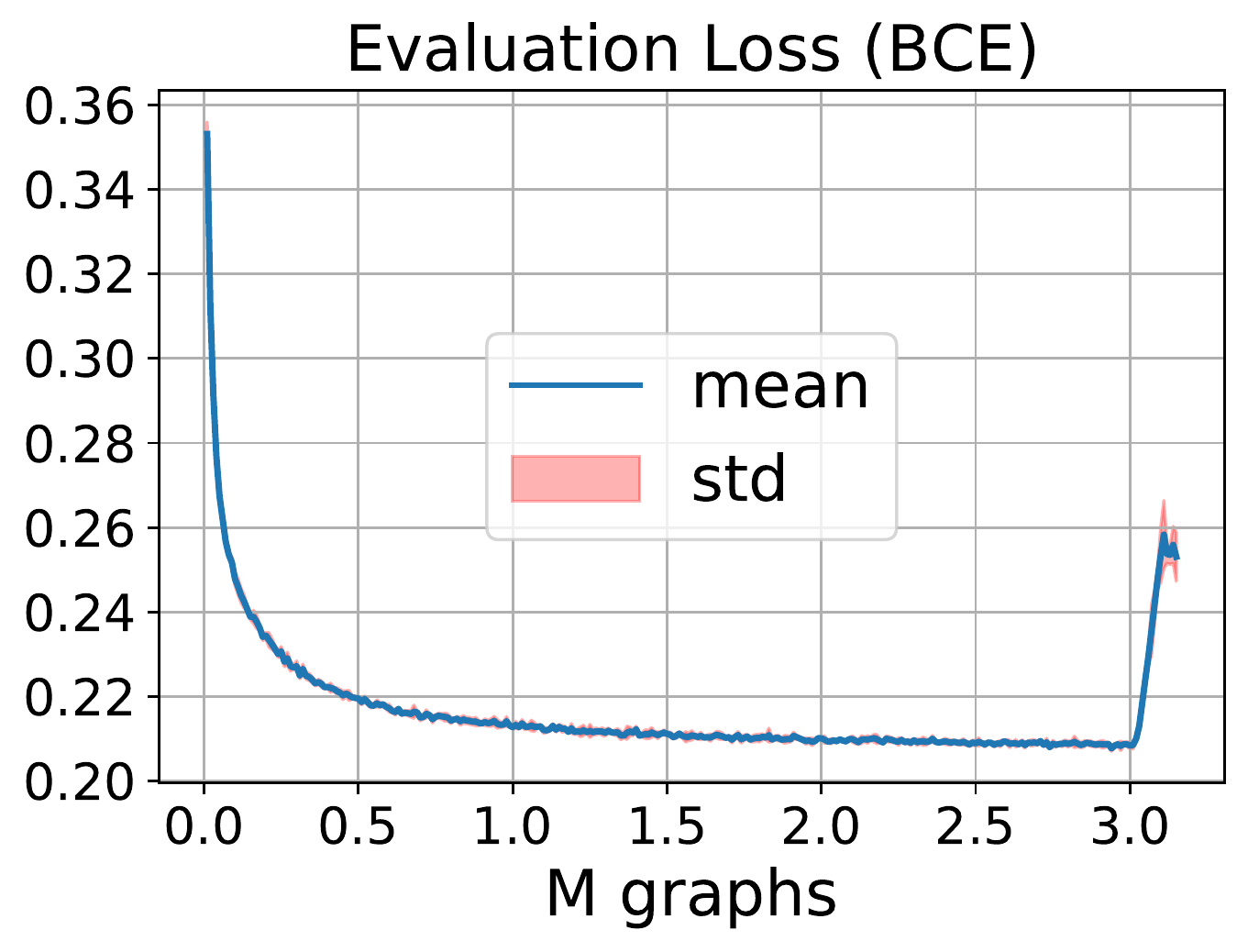} &
        \includegraphics[height=3cm]{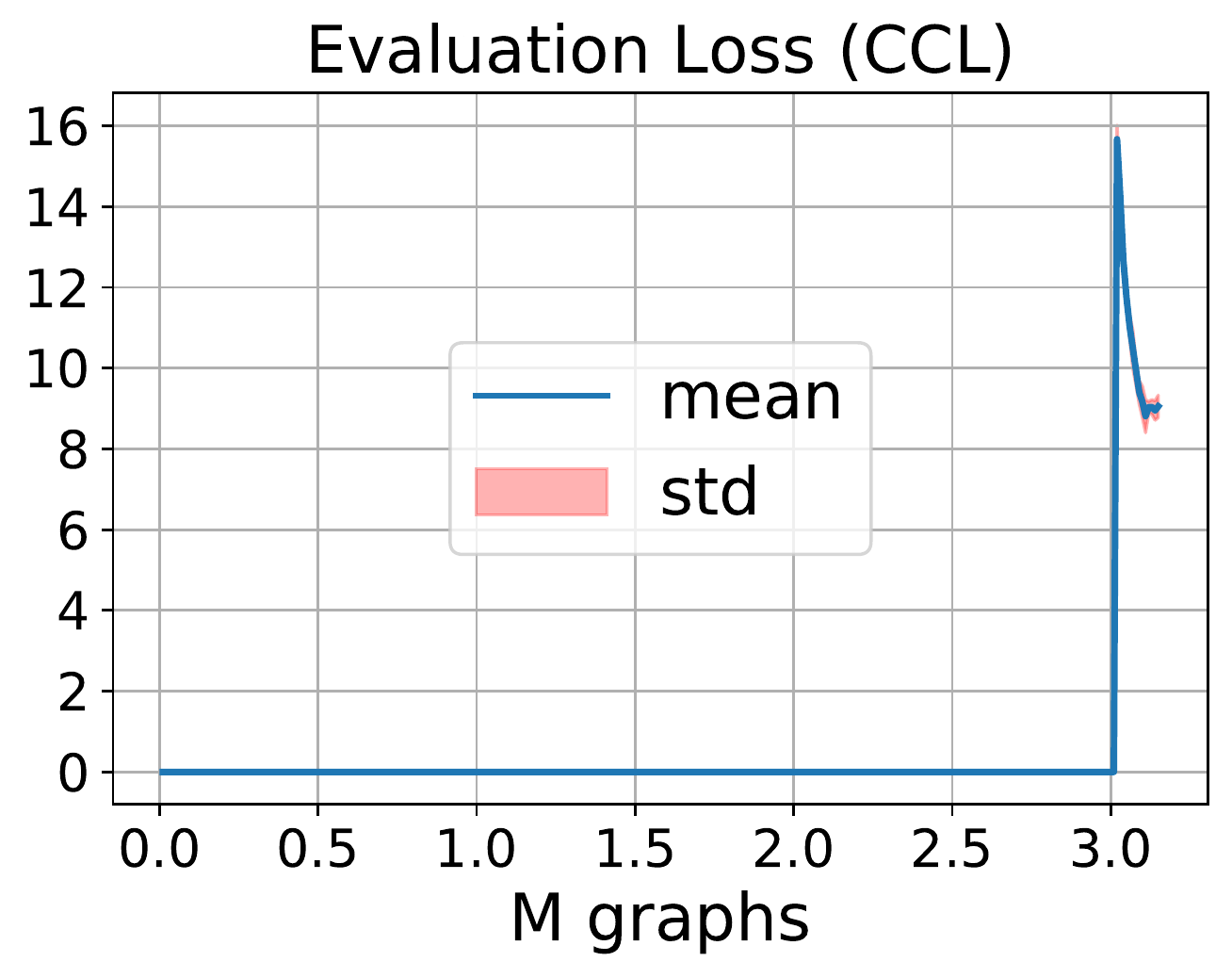} & 
        \includegraphics[height=3cm]{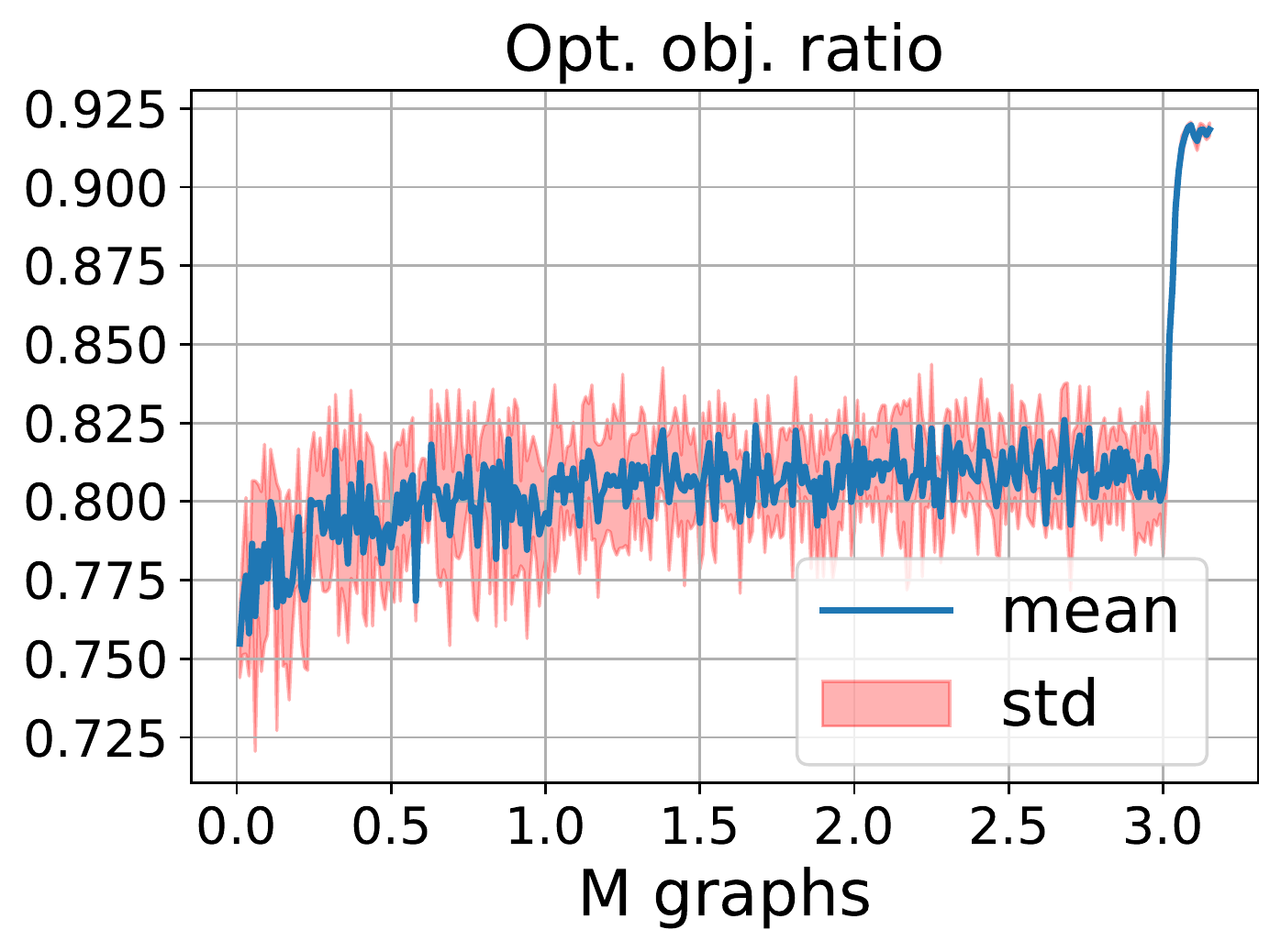} \\
        (a) BCE. &
        (b) CCL. &
        (c) Objective.
    \end{tabular}
    \caption{
        Mean and standard deviation (a) BCE evaluation loss, (b) CCL evaluation loss, and (c) evaluation optimal objective ratio of $5$ training runs of GCN\_W\_BN on RandomMP.
    }
    \label{fig:appendix-gcn-mean}
\end{figure*}

\autoref{fig:appendix-gcn-mean} shows mean evaluation plots of five training runs of GCN\_W\_BN on $3.15M$ instances of RandomMP.
No CCL was applied in the first $3M$ instances.
Then, $\alpha$ was linearly increased over $100k$ instances to $\alpha=0.01$.
Afterwards, the training continued for $50k$ instances with $\alpha=0.01$.
The node embedding dimensionality was set to $128$ and the number of GCN layers to $20$.
The MLP edge classifier consists of $2$ hidden layers with $256$ neurons.
Optimization was performed with Adam~\cite{2015Adam} ($0.001$ learning rate, $5\cdot10^{-4}$ weight decay, $(0.9, 0.999)$ betas) and a batch size of $200$.
Each training was performed on a MEGWARE Gigabyte G291-Z20 server on one NVIDIA Quadro RTX 8000 GPU and took $24$hrs on average, whereof the training time of the last $150k$ (CCL) instances took around $18$hrs.

\end{document}